\documentclass[11pt]{article}

\usepackage{amsmath, amsthm, amsfonts,amssymb}
\usepackage{bm}
\usepackage[left=1in,right=1in,top=1.1in,bottom=1.2in]{geometry}
\usepackage{fancyhdr}
\usepackage{natbib}
\usepackage{authblk}
\usepackage{graphicx, color}
\usepackage{subfigure}
\usepackage{epstopdf}
\usepackage{float}
\usepackage{hyperref}
\hypersetup{colorlinks,linkcolor={blue},citecolor={blue},urlcolor={red}} 
\usepackage{longtable}
\usepackage{float} 
\usepackage{amsmath, amssymb, amsfonts}
\usepackage[left=1in,right=1in,top=1.1in,bottom=1.2in]{geometry}
\usepackage{graphicx, float, booktabs, longtable, rotating} 
\usepackage{hyperref}
\hypersetup{colorlinks,linkcolor={blue},citecolor={blue},urlcolor={red}}  
\usepackage{todonotes}
\usepackage{ulem}
\usepackage{authblk}

\theoremstyle{plain}
\newtheorem{thm}{Theorem}

\newtheorem{lem}[thm]{Lemma}

\newtheorem{prop}[thm]{Proposition}

\theoremstyle{remark}

\newcommand{\beq}{\begin{equation}}
	\newcommand{\eeq}{\end{equation}}
\newcommand{\beas}{\begin{align*}}
	\newcommand{\eeas}{\end{align*}}
\newcommand{\bea}{\begin{align}}
	\newcommand{\eea}{\end{align}}
\newcommand{\bei}{\begin{itemize}}
	\newcommand{\eei}{\end{itemize}}
\newcommand{\ben}{\begin{enumerate}}
	\newcommand{\een}{\end{enumerate}}
\newcommand{\bet}{\begin{theorem}}
	\newcommand{\eet}{\end{theorem}}
\newcommand{\bel}{\begin{lemma}}
	\newcommand{\eel}{\end{lemma}}
\newcommand{\bep}{\begin{proposition}}
	\newcommand{\eep}{\end{proposition}}
\newcommand{\bed}{\begin{definition}}
	\newcommand{\eed}{\end{definition}}
\newcommand{\bec}{\begin{corollary}}
	\newcommand{\eec}{\end{corollary}}
\newcommand{\bex}{\begin{example}}
	\newcommand{\eex}{\end{example}}

\newcommand{\bK}{\bold{K}}

\newcommand{\R}{\mathbb{R}}

\newcommand{\vertiii}[1]{{\left\vert\kern-0.25ex\left\vert\kern-0.25ex\left\vert #1 
		\right\vert\kern-0.25ex\right\vert\kern-0.25ex\right\vert}}

\def\red{\textcolor{black}}

\date{}
\begin{document}
	\title{Uncovering smooth structures in single-cell data with PCS-guided neighbor embeddings}

\author[1,2,3]{Rong Ma\textsuperscript{*,\dag}}
\author[1]{Xi Li\textsuperscript{*}}
\author[1]{Jingyuan Hu}
\author[4,5,6]{Bin Yu\textsuperscript{\dag}}

\affil[1]{Department of Biostatistics, Harvard T.H. Chan School of Public Health}
\affil[2]{Department of Data Science, Dana-Farber Cancer Institute}
\affil[3]{Eric and Wendy Schmidt Center, Broad Institute of MIT and Harvard}
\affil[4]{Department of Statistics, University of California, Berkeley}
\affil[5]{Department of EECS, University of California, Berkeley}
\affil[6]{Center for Computational Biology, University of California, Berkeley}

\date{}

\maketitle

\begingroup
\renewcommand\thefootnote{\fnsymbol{footnote}}
\footnotetext[1]{These authors contributed equally}
\footnotetext[2]{Corresponding authors}
\endgroup

\newcommand{\bin}[1]{{\color{red} {\bf Bin:} #1}}
\newcommand{\rong}[1]{{\color{blue} {\bf Rong:} #1}}

  \begin{abstract}
    Single-cell sequencing is revolutionizing biology by enabling detailed investigations of cell-state transitions. Many biological processes unfold along continuous trajectories, yet it remains challenging to extract smooth, low-dimensional representations from inherently noisy, high-dimensional single-cell data. Neighbor embedding (NE) algorithms, such as t-SNE and UMAP, are widely used to embed high-dimensional single-cell data into low dimensions. But they often introduce undesirable distortions, resulting in misleading interpretations.  Existing evaluation methods for NE algorithms primarily focus on separating discrete cell types rather than capturing continuous cell-state transitions, while dynamic modeling approaches rely on strong assumptions about cellular processes and specialized data. To address these challenges, we build on the Predictability–Computability–Stability (PCS) framework for reliable and reproducible data-driven discoveries. First, we systematically evaluate popular NE algorithms through empirical analysis, simulation, and theory, and reveal their key shortcomings such as artifacts and instability. We then introduce NESS, a principled and interpretable machine learning approach to improve NE representations by leveraging algorithmic stability and to enable robust inference of smooth biological structures. NESS offers useful concepts,  
    quantitative stability metrics, and efficient computational workflows to uncover developmental trajectories  and cell-state transitions in single-cell data. Finally, we apply NESS to multiple single-cell datasets, including those about pluripotent stem cell differentiation, organoid development, and multiple tissue-specific lineage trajectories. Across these diverse contexts, NESS consistently yields useful and verifiable biological insights, such as identification of transitional and stable cell states and quantification of transcriptional dynamics during development. Notably, NESS resolves distinct neuronal subpopulations during embryoid formation and provides a deeper understanding of their cell-state dynamics.
 \end{abstract}

	
	\section{Introduction}

  Single-cell sequencing technologies generate high-dimensional molecular profiles at cellular resolution, enabling the study of cell-state transitions in processes such as differentiation, reprogramming, and disease progression \cite{yao2024single,verma2020robust,sritharan2021computing,clifton2023stalign,saelens2019comparison,wang2021current,ding2022temporal}. While these datasets are often sparse and noisy due to technical factors such as dropout events and measurement variability, the cellular states they capture typically arise from structured, continuous biological processes.
Consequently, large-scale single-cell datasets frequently exhibit lower-dimensional smooth manifold structures embedded within the high-dimensional feature space. These manifolds reflect the geometry of the underlying biological dynamics and can take the form of linear trajectories \cite{saelens2019comparison,heitz2024spatial}, cycles \cite{liang2020latent}, or branching tree-like structures \cite{jahn2016tree,walker2023nest}. Identifying and characterizing such geometric structures is essential for understanding lineage relationships, developmental pathways, and disease progression \cite{yao2024single,verma2020robust,sritharan2021computing,clifton2023stalign}. In particular, trajectory inference methods aim to organize cells along a continuous trajectory in a lower-dimensional space, capturing the progression of biological processes \cite{saelens2019comparison,wang2021current,ding2022temporal}. Moreover, analyzing geometric properties of the manifold, such as curvature, branching structure, and continuity, has provided valuable insight into the dynamics and regulation of cellular activity \cite{sritharan2021computing,cheng2025phlower}. 

A key challenge in analyzing single-cell omics data is to accurately represent and interpret biologically meaningful manifold structures in a low-dimensional space.\footnote{throughout, we use the words ``smooth structure" and ``manifold structure" interchangeably.} To address this, neighbor embedding (NE) methods, such as t-SNE \cite{van2008visualizing}, UMAP \cite{mcinnes2018umap}, PHATE \cite{moon2019visualizing} and their extensions \cite{tang2016visualizing, linderman2019fast,narayan2021assessing}, have been developed and become widely used tools for embedding high-dimensional single-cell data into more tractable low-dimensional representations. These methods typically begin by constructing an affinity graph among cells, followed by iterative updates to the low-dimensional embeddings that optimize a global objective function aimed at preserving local neighborhood relationships.
Beyond visualization, NE methods also play a central role in many state-of-the-art algorithms for downstream single-cell analysis, such as trajectory inference \cite{wolf2019paga,cao2019single,stassen2021generalized,setty2019characterization,lange2022cellrank,weiler2024cellrank} and RNA velocity analysis \cite{la2018rna,bergen2021rna,gorin2022rna,bergen2020generalizing}.
  
Despite their widespread applications, the reliability of NE algorithms, that is,  whether these algorithms faithfully\footnote{preserving both global and local structures.} capture underlying smooth structures and related biological signals, remains elusive. 
While simplifications are sometimes necessary to extract meaningful biological insights from complex data \cite{phillips2012physical}, a key limitation of NE methods is their tendency to introduce substantial artifacts that can misrepresent underlying biological structures. This problem is exacerbated by the high-dimensional, noisy nature of single-cell data and the intrinsic complexity of biological processes. Recent studies \cite{kobak2021initialization,kobak2019art,chari2023specious,ma2023spectral,ma2024principled,sun2023dynamic,liu2025assessing} have highlighted that NE methods can generate misleading structures, such as artificial separation or clustering of cell populations, which do not reflect true biological structures.
These artifacts can have significant downstream consequences, including misclassification of cell types \cite{ma2024principled,chari2023specious,liu2025assessing,kobak2019art}, spurious lineage relationships \cite{sun2023dynamic,ma2023spectral}, and incorrect inferences about differentiation trajectories \cite{gorin2022rna}.
Moreover, the results of NE algorithms can be  sensitive to parameter choices and distance metrics, with minor changes in them potentially leading to significant alterations in the inferred structures, affecting downstream biological conclusions \cite{cai2022theoretical,huang2022towards,xia2024statistical,kobak2021initialization}.

To address these limitations, several methods have been developed to assess the reliability of NE algorithms for a given dataset. For example,  EMBEDR \cite{johnson2022embedr} and scDEED \cite{xia2024statistical} generate random null data through permutation or re-sampling techniques to calculate  reliability scores for every cell embedding based on the similarity between the cell’s 2D-embedding neighbors and pre-embedding neighbors. Opt-SNE \cite{belkina2019automated} utilizes Kullback-Leibler divergence evaluation in real time to tailor the NE algorithm in a dataset-specific manner, enabling fast computation and hyperparameter selection to improve visualization of cluster structures.  DynamicViz \cite{sun2023dynamic} creates bootstrap samples from the original data to generate a sequence of dynamic visualizations, and LOO \cite{liu2025assessing} inspects the landscape of the objective function of specific NE optimization, each capturing the variability of the cell embedding against certain sample perturbations. While these methods are useful for detecting dubious cell embeddings\footnote{cell embeddings that do not preserve neighborhood structure as compared with the input data.} and optimize hyperparameters of NE algorithms,  they primarily focus on datasets with clear cluster structures, such as atlas-level single-cell datasets comprising distinct cell types, with the goal of better distinguishing these clusters. However, for applications involving smooth structures such as cell differentiation with continuously transitioning cell states, these methods typically do not account for the dynamic nature of the underlying biological processes. Consequently, {their evaluation metrics often do not adequately assess the biological relevance of the reconstructed low-dimensional cell-state manifolds} and provide only limited insight into the dynamics of cell-state transitions, as shown below. On the other hand, some recent works have explored the  theoretical foundations of NE algorithms such as their (in)consistency for visualizing clustered data \cite{cai2022theoretical,linderman2022dimensionality,arora2018analysis,bergam2025t} and the existence of global minimizer of the loss function \cite{auffinger2023equilibrium,jeong2024convergence}. However, there remains a lack of systematic evaluation and  understanding about the key factors underlying the performance of NE algorithms in capturing smooth structures. 

In this work, we introduce a novel machine learning approach to improve NE algorithms for reliably capturing low-dimensional smooth structures in single-cell data.
Our approach is based on the PCS framework for veridical data science \cite{yu2020data,yu2020veridical,yu2023uncertainty,yu2024veridical}, which employs core principles of Predictability, Computability, and Stability to provide formal guidelines for the entire data science life cycle, including method development, empirical evaluation and enhancement of trustworthiness in data analysis. Our contribution is two-fold. 
\red{First, we conduct a systematic PCS-guided assessment of popular NE algorithms such as t-SNE, UMAP, PHATE, and a few others. Our evaluation integrates benchmark biological datasets with biological labels, numerical simulations, and a rigorous theoretical analysis to provide a comprehensive reality check (or ``Predictability" assessment or Pred-check) for these NE algorithms. From this analysis, we uncover fundamental insights into their limitations in capturing low-dimensional smooth structures, {including algorithmic artifacts and  instability}.  Second, we propose NESS, a wrapper algorithm designed to work with \emph{NE} methods, which enhances their ability to capture \emph{Smooth} structures through \emph{Stability}-based measures. The proposed NESS approach leverages the intrinsic stochasticity of the NE algorithms, offered by the random initialization, to generate a series of low-dimensional embeddings and use the similarity between the embeddings' local neighborhood structure to define stability measures for each data point. The NESS approach aims at improving diverse NE algorithms for modeling smooth structures in single-cell data, as well as providing biological insights on cell state dynamics. In particular, it may help identify stable and transitional cell states along developmental trajectories and quantify cellular transcription activity dynamics across different cell states.  }


To demonstrate its efficacy, we use NESS {combined with t-SNE or UMAP} to analyze several real single-cell datasets (see Supplement Table \ref{table.1} for each dataset and their abbreviations). These datasets cover diverse biological processes including hematopoiesis \cite{paul2015transcriptional}, human induced pluripotent stem cells (iPSC) differentiation \cite{bargaje2017cell}, murine intestinal organoid development \cite{battich2020sequencing}, dentate gyrus neurogenesis \cite{hochgerner2018conserved}, spermatogenesis \cite{hermann2018mammalian}, and endoderm development in embryoid bodies \cite{moon2019visualizing}. They also encompass various sequencing technologies, such as scRNA-seq (10x Genomics Chromium), single-cell EU-seq, and single-cell RT-qPCR. Using biological labels from trusted external sources, we demonstrate that NESS (combined with t-SNE or UMAP) identifies both transitional and stable cell states across iPSC differentiation,  organoid development, and embryoid body formation, predicts key genes involved in cell-state transitions, and quantifies cell-specific transcriptional activity in  neurogenesis and spermatogenesis. Notably, our analysis resolves key neuronal subpopulations  during embryoid formation and reveals novel biological insights into their respective cell states. Overall, our analysis suggests that NESS improves on NE algorithms to obtain more reliable characterizations of smooth biological structures in single-cell data.

\section{Results}


\begin{figure}[htbp]
    \centering    \includegraphics[width=1\linewidth]{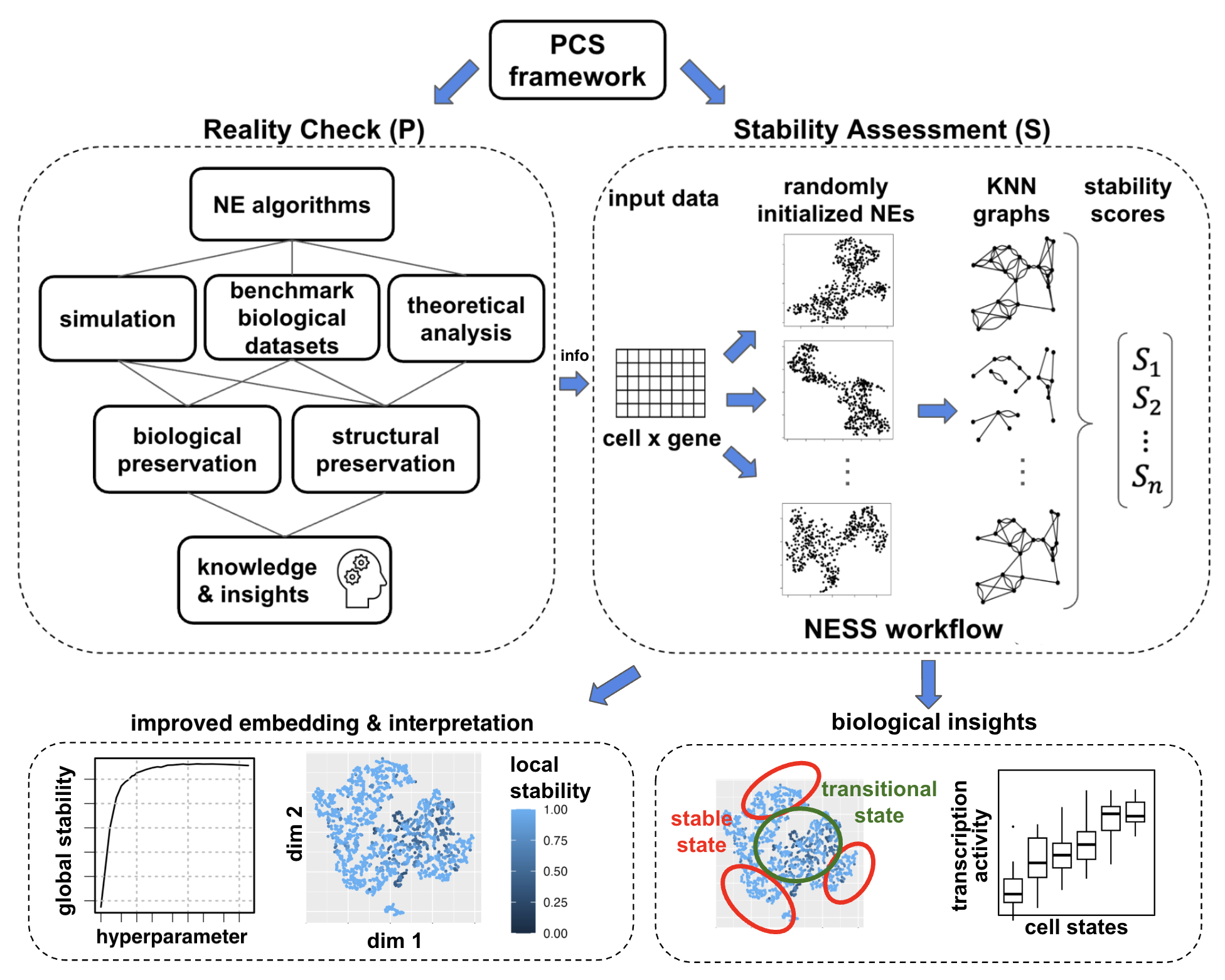}
    \caption{\small Overview and illustration of the proposed approach. Our approach is based on the PCS (Predictability, Computability, and Stability) framework for veridical data science. The ``P" in PCS requires a comprehensive reality check for popular NE algorithms. Using numerical simulations, benchmark biological datasets, and theoretical analysis, we evaluate their performance and identify key limitations in capturing smooth structures and retaining biological signals. Guided by the ``S" in PCS, we develop NESS, a novel algorithm for stability assessment and improvement of any NE algorithm on a given dataset without requiring ground-truth biological labels. NESS leverages the random initialization of NE algorithms to generate multiple low-dimensional embeddings. From each embedding, it constructs a KNN graph among the data points (cells) and compares these graphs to compute a local stability score for each data point. In addition, NESS also generates a refined low-dimensional embedding that better reflects the latent smooth structure and provides an intuitive line chart with an automated workflow for hyperparameter selection. Building on these outputs, NESS aims to offer biological insights into cell progression dynamics, such as identifying stable and transitional cell states along developmental trajectories and quantifying transcriptional dynamics across cell states. }
    \label{fig.1}
\end{figure}

\paragraph{PCS framework, overview of our approach, and NESS.} The PCS framework for veridical data science \cite{yu2020data,yu2020veridical,yu2023uncertainty,yu2024veridical} is a principled data science workflow designed to enhance the reliability and interpretability of data-driven scientific discoveries. It is built upon three key principles: Predictability (P), Computability (C), and Stability (S). Specifically, the predictability principle ensures that models fit real-world data well, providing a comprehensive reality check using benchmark datasets to validate their usefulness and generalizability. The computability principle emphasizes efficient, reproducible and scalable computational implementations, and real-world-data-inspired simulations, ensuring coding robustness in practical applications. The stability principle assesses the sensitivity of results to reasonable data perturbations, model choices, and algorithmic randomness, helping to distinguish genuine findings from artifacts.
\red{ The PCS framework integrates these principles throughout the entire data science pipeline, from problem formulation to method development, empirical evaluation, and interpretation. Our proposed approach, whose key idea and workflows are illustrated in Figure \ref{fig.1}, is developed under the conceptual lens of PCS framework.} 

\red{We first perform a comprehensive reality check for popular NE algorithms under ``P" in PCS, to evaluate their performance and identify key limitations in capturing smooth structures and retaining biological signals.} Specifically, we subject the low-dimensional embedding generated by each NE algorithm to several internal and external evaluation and validation checks. Internal validation uses benchmark single-cell datasets, numerical simulations, and/or rigorous theoretical analysis, to assess how well the embedding preserves local and global structure relative to the original high-dimensional data. External validation leverages biological labels (e.g., cell type or cell state annotations) associated with the benchmark datasets, available from trusted sources, to evaluate how well the NE algorithms retain biological signals. \red{As detailed below, our systematic assessment reveals  insights into the limitations of NE methods and the main factors determining their performance  in characterizing smooth structures.} These limitations include the artificial fragmentation of smooth structures within the embeddings and instability in the algorithm’s output with respect to random initializations.

\red{Next, informed by the above reality checks, we propose a new algorithm, NESS, which provides stability assessment (``S" in PCS) of any NE algorithm with respect to a given dataset, aiming to enhance the performance of the NE algorithm in capturing the underlying smooth structure without requiring ground-truth biological labels.} NESS exploits the inherent stochasticity of an NE algorithm, arising from random initialization of any iterative procedure, to introduce  a point-wise stability measure for the final low-dimensional embeddings. Specifically, for a given input dataset and an NE algorithm, NESS first generates multiple low-dimensional embeddings of the dataset under different random initializations. From each embedding, we construct a $k$-nearest neighbor (KNN) graph among the data points (cells), and then compare  KNN graphs across all embeddings to assess the stability of each data point's neighbors; see Methods for more details. As a result, NESS defines a local stability score for each data point, with a higher value indicating more consistent neighbors across multiple embeddings, providing a quantitative measure of the uncertainty in the obtained low-dimensional embeddings. Furthermore, NESS produces two additional outputs: a refined low-dimensional embedding that better reflects the underlying smooth structure, and a line chart with an automated procedure that guides hyperparameter selection. 

\red{Finally, building on these outputs, NESS may be used to obtain biological insights into cell progression dynamics. This includes using local stability scores to identify stable and transitional cell states along developmental trajectories together with the associated genes and pathways, and constructing a cell-specific transcription activity score to quantify the transcriptional dynamics across different cell states.}
NESS is computationally efficient and scalable to large single-cell datasets with tens of thousands of cells (``C" in PCS). It has been implemented as an R package, available at our GitHub page \url{https://github.com/Cathylixi/NESS}, accompanied by PCS documentation and tutorials for broader community use. 

\paragraph{PCS-guided assessment reveals limitations of NE algorithms for smooth biological structures.} Our PCS-guided assessment and validation of NE algorithms identify common limitations of popular NE algorithms, and provide empirical evidence and theoretical account of their unfavorable fragmentation of latent smooth structures.

\begin{figure}[htbp]
    \centering    
    \includegraphics[width=0.95\linewidth]{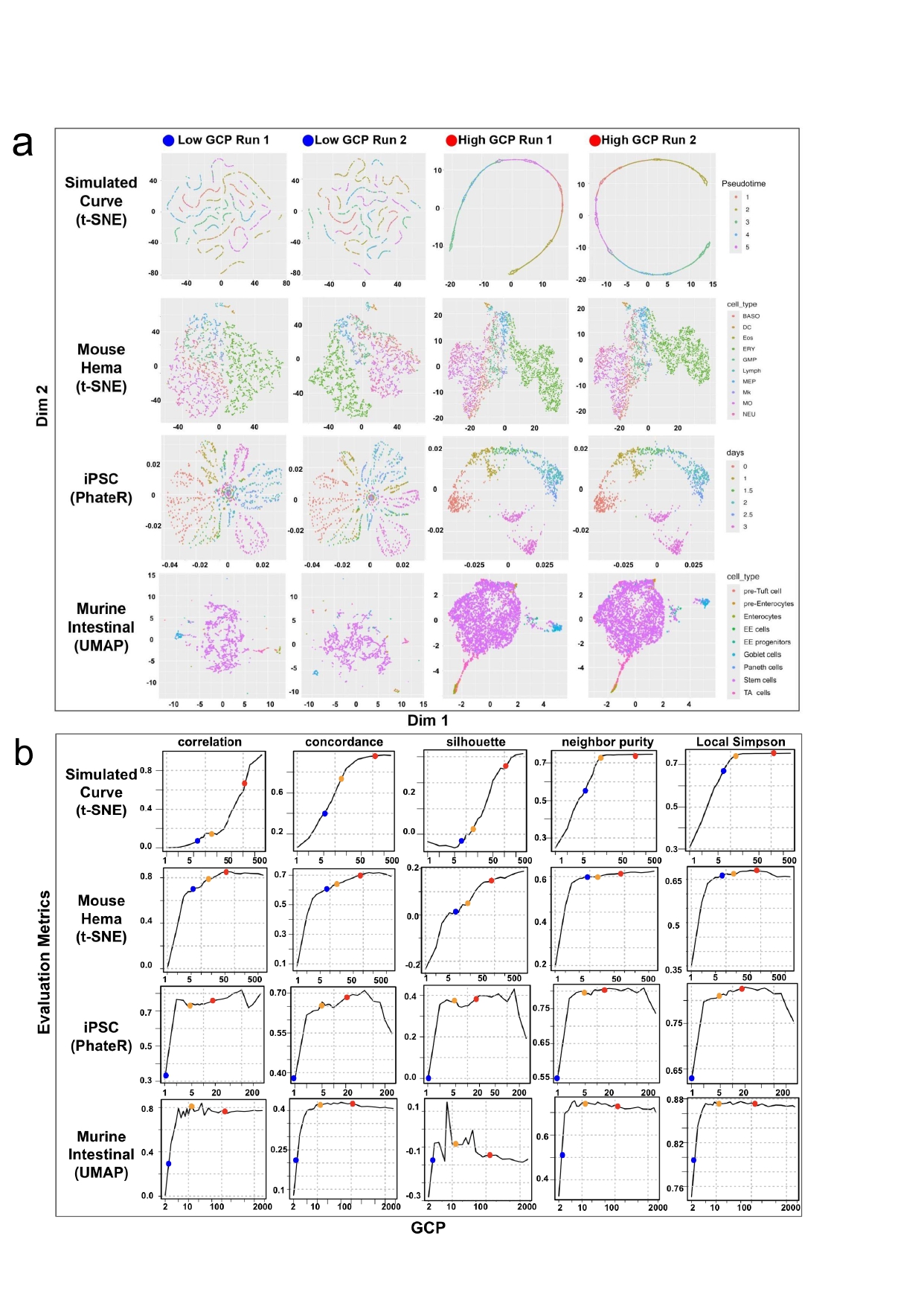}
    \vspace{-1cm}
    \caption{\small (Caption next page.)}
    \label{fig.2}
\end{figure}

\addtocounter{figure}{-1}
\begin{figure} [t!]
  \caption{\small (Previous page.) PCS-guided assessment identifies artifacts in NE algorithms. (a) Examples of low-dimensional embeddings for synthetic (first row) and benchmark single-cell datasets (last three rows) generated by popular NE algorithms, including t-SNE, UMAP, and PhateR, with cells colored by their labels. The left two columns correspond to two instances of RIs under relative low GCPs, whereas the right two columns show embeddings under relatively high GCP values. The specific low and high GCP values are indicated in (b). See Methods for the definition of GCP in each NE algorithm. All embeddings are generated using random initialization (RI). See Supplementary Figures \ref{sup.fig.2}--\ref{supp.fig.12} for more examples and similar results under alternative initialization schemes, and Supplementary Figure \ref{supp.fig.4} for the ground truth cell differentiation hierarchies. (b) Assessment of NE algorithm performance based on multiple metrics and the effect of GCP tuning. Line charts depict the relationship between GCP (x-axis, log scale) and five internal and external evaluation metrics (y-axis): Correlation, Concordance, Silhouette Score, Neighbor Purity, and Local Simpson Index, where higher values indicate more reliable embeddings. {The red dots and blue dots on the line charts indicate the respective high and low GCPs in (a).} \red{The yellow dots indicate the default GCPs.}
Our results suggest under-specifying GCP tends to introduce artificial fragmentation and disruption of the latent smooth structure, while increasing instability in the final embeddings. \red{See Supplementary Figure \ref{supp.fig.nonrandom} for similar results under alternative initialization schemes.} 
}
\end{figure}

\noindent {{\it\uline{Low graph connectivity is problematic for an NE algorithm to capture smooth structures.}}}
An important component in the existing NE algorithms is the construction of a neighborhood graph based on the input dataset.  In each NE algorithm, the connectivity of the resulting graph is either explicitly or implicitly determined by a hyperparameter, hereafter universally referred to as the ``graph connectivity parameter" (GCP), that controls the resolution of the neighborhood definition. For example, in UMAP, densMAP, and PHATE (hereafter referred to as PhateR), the GCP is defined as the number of neighbors $k$ in the $k$-nearest neighbor graph based on the input data \cite{mcinnes2018umap,moon2019visualizing,narayan2021assessing}; in t-SNE, it is defined as the ``perplexity," which determines edge weight assignment through entropy measures \cite{van2008visualizing}. In general, low GCP values produce sparser graphs, reducing computational cost, whereas high GCP values lead to denser graphs with increased computational demands. In practice, smaller GCPs are often preferred for efficiency, yet there is no principled approach for selecting GCPs that best preserve smooth structures. 

To  evaluate NE algorithms relative to their GCP hyperparameter settings, we carry out a systematic reality check of their performance on both synthetic data and {real-world single-cell datasets, each representing a distinct developmental process, including tissue-specific lineage trajectories (Mouse Hema), pluripotent stem cell differentiation (iPSC), and organoid development (Murine Intestinal)}. Every single-cell dataset contains appropriate biological labels such as cell types or time points (indicative of cell state). {We assessed four widely used NE algorithms, including t-SNE, UMAP, PhateR, and densMAP,} using multiple evaluation metrics. To quantify structure preservation, we measure the neighborhood concordance and the Pearson correlation of pairwise distances between the low-dimensional embedding and input data (Methods). To assess the preservation of biological signals, we use the Silhouette index, neighbor purity index, and local Simpson index, in each case comparing the embedding coordinates with biological labels (Methods). Our results highlight the critical role of GCP selection in ensuring reliable embeddings (Figures \ref{fig.2}, \ref{sup.fig.2}--\ref{supp.fig.12}). Specifically, we find that for all NE algorithms evaluated in this study, regardless of their initialization schemes, under-specifying GCPs tend to introduce substantial distortions, causing artificial fragmentation and disruptions in the latent smooth structures (Figures \ref{fig.2}a, \ref{sup.fig.2}--\ref{supp.fig.12}).  For example, for our synthetic data (Simulated Curve) randomly sampled from a one-dimensional curve representing an ideal, noiseless progression trajectory, t-SNE embeddings with low GCP values fragmented the continuous curve into multiple disconnected segments; in the Mouse Hema dataset, low-GCP t-SNE embeddings fragmented the two differentiation branches into many small, disconnected clusters, obscuring the true trajectory structure. \red{These observations are consistent with prior findings \cite{kobak2019art,wattenberg2016how}.} {Additional results across different initialization schemes and dataset–method combinations are provided in Supplementary Figures \ref{sup.fig.2}--\ref{supp.fig.12}.}  Conversely, except for PhateR, which exhibits a slight decline in performance across various metrics under very large GCPs, \red{most NE algorithms display robust performance with higher GCPs, and are relatively insensitive to GCP variations in that range (Figure \ref{fig.2}b and Supplement Figure \ref{supp.fig.nonrandom}). However, these benefits come at the cost of increased computational time (Supplementary Figure \ref{supp.fig.10}a), making it desirable to identify suitable GCP values without exhaustively exploring large ones. Moreover, in some cases (e.g., the iPSC (PhateR) example in Figure \ref{fig.2}b and a CD8+ T-cell (t-SNE) example in Supplementary Figure \ref{supp.fig.over}), excessively large GCP can lead to oversmoothing of the latent structure. Together, these observations motivate the need for a principled, data-driven approach to GCP selection.} 

\begin{figure}[htbp]
    \centering
    \includegraphics[width=1\linewidth]{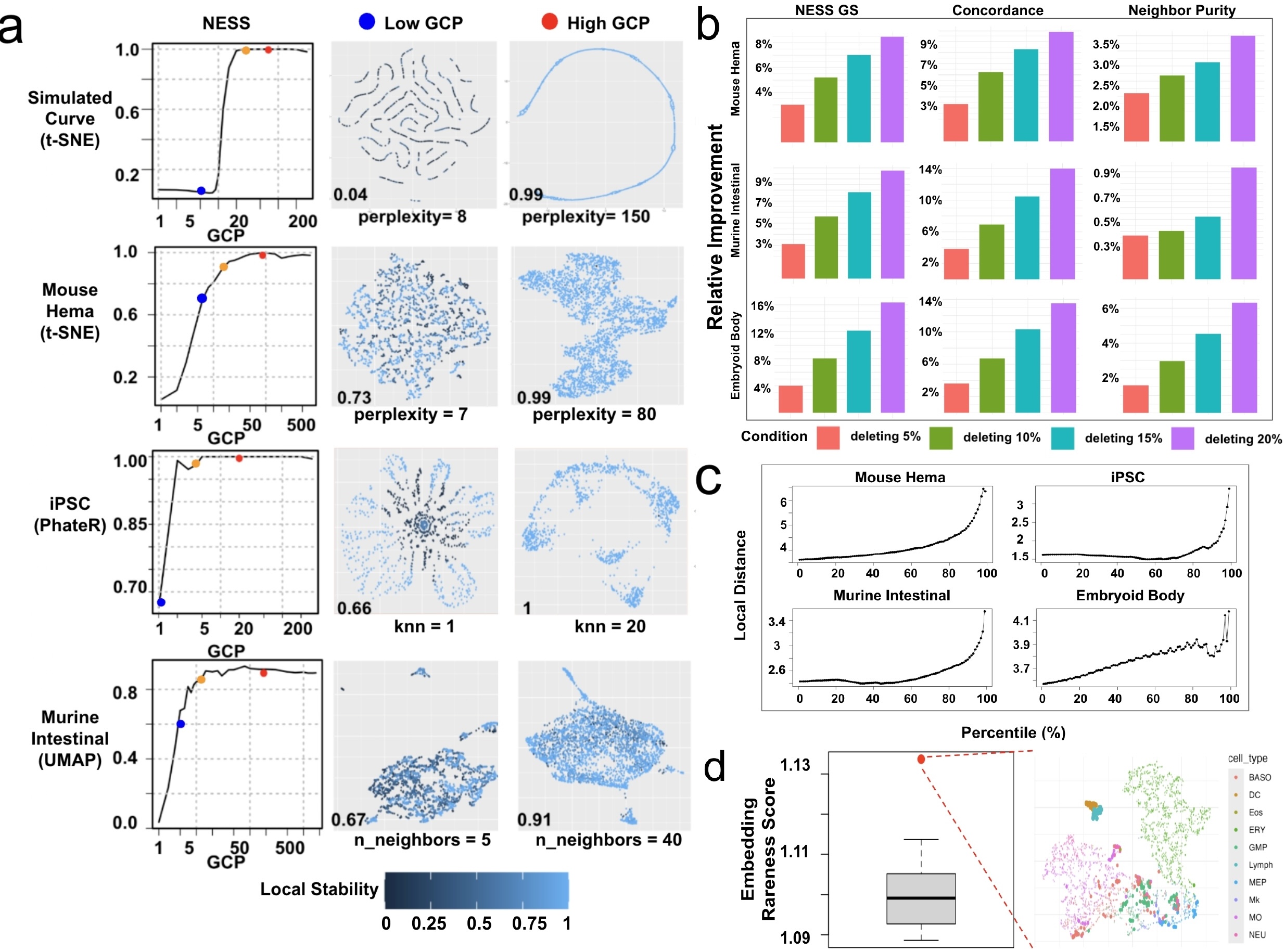}
    \caption{\small NESS enhances representation and uncertainty quantification of smooth structures. (a) Left column: line charts showing the relationship between GCP tuning and the NESS global stability (GS) score for various NE algorithms, evaluated on simulated and real single-cell datasets. Middle and right columns: low-dimensional embeddings corresponding to two GCP values, a lower GCP value (``Low GCP") and a higher GCP value (``High GCP"), indicated by {blue and red dots  respectively} in the line charts. The GCP and its value for each NE algorithm are indicated below the embeddings. Data points are colored by NESS local stability scores, with the corresponding GS scores shown in the bottom-right corner of each plot. Higher GCP values generally improve stability and structure preservation. \red{Supplementary Figures \ref{supp.fig.5} and \ref{supp.fig.6} illustrate how these stability measures can also guide the selection of NE methods and other hyperparameters. } (b) Barplots showing the relative improvement in NESS GS score, concordance, and neighbor purity across different datasets as increasing percentages (5\%, 10\%, 15\%, 20\%) of low-stability cells are removed. All GS scores are evaluated under the respective NESS recommended GCP values. The iPSC dataset is excluded since its global stability reaches 1. (c) Cells with low NESS local stability scores tend to have greater distances from their neighboring cells. Plots show the average distance between each cell and its 30-nearest neighbors, among cells whose local stability score falls in the $p\%$ upper percentile of the cells, all obtained under NESS recommended GCPs. Higher percentiles correspond to cells with lower NESS local stability and increased local distances to their neighbors, indicating their tendency to reside in low-density regions. \red{(d) Left: Boxplot of {embedding rareness scores} for 50 t-SNE embeddings of the Mouse Hema dataset under RI and the default GCP (perplexity=30), with the largest score marked in red. Right: t-SNE embedding corresponding to the largest embedding rareness score, with cells color-coded by cell type. Lymph and DC cells are highlighted along with their 50 nearest neighbors in the original gene expression space, which primarily include MEPs, GMPs, and a small number of BASO and Mono cells. In the embedding, these originally neighboring populations are dispersed across distinct regions, indicating structural distortions introduced by the NE algorithm.}}
    \label{fig.3}
\end{figure}

\noindent{\it\uline{Theoretical analysis of t-SNE connects low graph connectivity to fragmented embeddings.}}
In addition to the empirical assessments above, we also establish a mathematical connection between low GCP values and artifacts in low-dimensional embeddings produced by NE algorithms. In particular, through rigorous analysis of t-SNE, we find that for {synthetic data generated from a low-dimensional manifold model}, an under-specified GCP inevitably leads to a pathological deformation of the optimization landscape, causing severe distortions in the final embedding (Theorem \ref{tsne.thm} in Methods). Moreover, our analysis highlights that these artifacts arise primarily from discontinuities or ``gaps" in the low-dimensional representation, explaining the fragmented patterns observed in various examples (Figures \ref{fig.2}a and \ref{sup.fig.2}, left two columns, and Figure \ref{fig.3}a middle column). {Despite that our theoretical analysis only focuses on t-SNE,  the underlying similarity between the loss functions of t-SNE and other NE algorithms \cite{bohm2022attraction,kobak2021initialization,damrich2021umap} suggests that our findings likely extend to a broader class of NE methods.}

\noindent \red{ {\it\uline{Random initialization enables stability analysis through algorithmic perturbation.}}
Another crucial factor of the NE algorithms is the initialization scheme \cite{kobak2021initialization}. As iterative algorithms, NE methods commonly use random initialization (RI), PCA initialization, Laplacian initialization, and so on. RI is typically implemented by sampling initial embedding coordinates within a small region around the origin \cite{van2008visualizing}. While RI has been criticized for its limited interpretability and potential to distort global structure \cite{cai2022theoretical,kobak2021initialization}, we deliberately leverage RI as a principled source of algorithmic perturbation within the PCS framework. In contrast to structured initializations such as PCA, which often lead to nearly identical converged solutions and thus obscure instability, RI enables exploration of multiple local optima of the NE objective and reveals embedding configurations that are weakly constrained by the data.}
Consistent with this perspective, our findings reveal that low GCP values, combined with RI, lead to highly variable distortions across different runs (Figure \ref{fig.2}a and Supplement Figure \ref{sup.fig.2}). For example, in the synthetic data (Simulated Curve), the low-GCP t-SNE embeddings  contain {arbitrary fragmentation and an arbitrary layout} of the underlying curve; for the iPSC dataset, both low-GCP PhateR embeddings failed to capture the latent differentiation trajectory, yet displayed flower-shaped patterns, with notable structural discrepancies between {the resulting embeddings}. Similar fragmentation and instability are observed in the Mouse Hema dataset visualized with t-SNE and the Murine Intestinal dataset visualized with UMAP. In contrast, NE embeddings under relatively higher GCPs exhibit greater structural stability under RI. These findings suggest that careful selection of GCP not only enhances embedding quality but also improves the algorithm’s robustness to initialization variability. This key insight underpins the main idea behind the design of NESS, which is to develop stability measures to achieve more reliable performance.

\paragraph{Proposed NESS enhances NE representation and stability assessment of smooth biological structures.} 
To mitigate potential artifacts of NE algorithms in single-cell analysis, where ground-truth biological labels are typically unavailable, we propose the NESS approach to enhance the representation and stability assessment of smooth biological structures based on the stability principle. We demonstrate that NESS improves the performance of diverse NE algorithm in characterizing smooth biological structures in four key aspects.

\noindent {{\it \emph{NESS global stability score optimizes embedding of smooth structures and guides hyperparameter selection.}}} For a given dataset and an NE algorithm, NESS first defines a local stability score for each cell and uses its average across all cells to define a global stability (GS) score of this NE algorithm (Methods). It then generates a line chart where the x-axis is the GCP value and the y-axis is the global stability scores of the NE algorithm (Fig. \ref{fig.3}a, left column). \red{We find that the NESS GS score can consistently distinguish visualizations exhibiting structural distortions (Fig. \ref{fig.3}a, middle column) from those that better capture smooth biological structures across diverse NE algorithms and settings (Fig. \ref{fig.3}a, right column).} It thus may facilitate informed selection of  GCPs in an unsupervised manner (i.e., without biological labels).
\red{From the line chart, users can select a GCP associated with a higher GS value while avoiding unnecessary computational overhead that only marginally improves embedding quality.} To enhance computational efficiency, we develop a workflow (Methods) that automatically selects a suitable GCP for downstream analysis; \red{it recommends the smallest GCP that achieves a high GS score or shows no substantial improvement compared to a slightly smaller value. This strategy enables early stopping in the grid search and avoids unnecessary evaluation of larger candidate GCP values. See Section \ref{sec.time} for more discussions on computing time of NESS.}  

Similarly, the GS metric can be applied to compare and optimize other hyperparameters, including categorical ones such as the choice of distance metric in a given NE algorithm (Supplementary Figure \ref{supp.fig.5}). Additionally, it can aid in selecting among different NE algorithms, helping to identify the most suitable method for visualizing a specific dataset (Supplementary Figure \ref{supp.fig.6}).

\begin{figure}[htbp]
    \centering
    \includegraphics[width=1\linewidth]{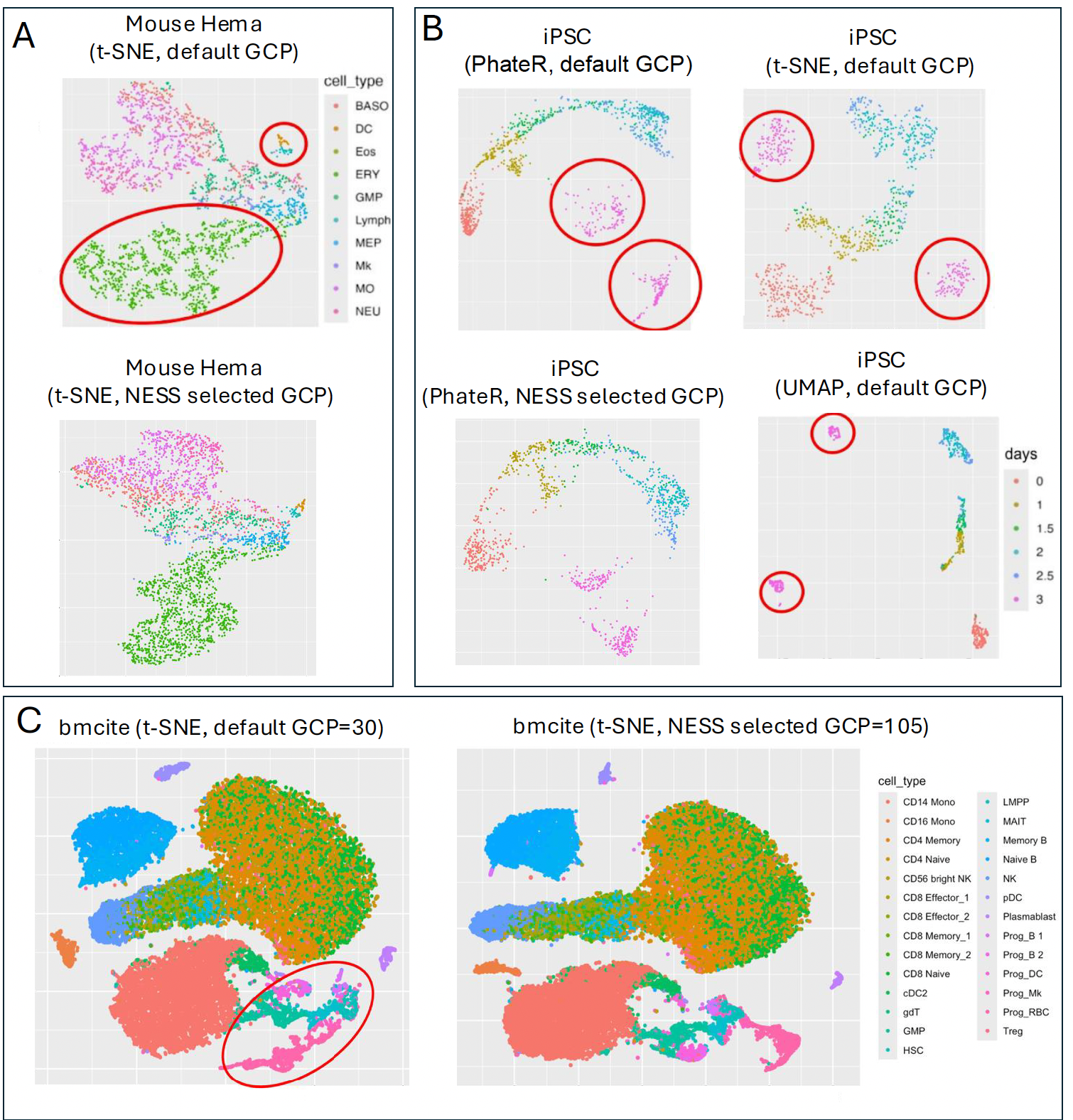}
    \caption{\small \red{Examples of low-dimensional embeddings obtained by NE algorithms under default and NESS-recommended GCPs. (a) t-SNE under default  GCP (GCP=30)  applied to the Mouse Hema dataset led to artificial fragmentation of erythrocytes (ERY) and separated dendritic cells (DC) from the main developmental trajectory, as highlighted in red circles. (b) default applications of PhateR (default GCP=5), t-SNE (default GCP=30), and UMAP (default GCP=15) to the iPSC dataset fail to capture the lineage relationship (Supplementary Figure \ref{supp.fig.4}b) between the primitive streak population at day 2-2.5 and the differentiated mesodermal and endodermal cells at day 3, as highlighted in red circles. (c) applying t-SNE under default GCP to the bmcite dataset lead to artificial fragmentation of regulatory T cells. These artifacts can be overcome by the embeddings under NESS recommended GCP. Supplementary Figure \ref{supp.bench} benchmarks NESS with existing methods in hyperparameter selection. }}
    \label{fig.4new}
\end{figure}

\noindent {{\it \emph{NESS local stability score quantifies pointwise embedding reliability.}}} For any NE algorithm with NESS recommended hyperparameters, the NESS pointwise local stability score quantifies the algorithm's technical stability on a given dataset. 
These scores help distinguish more stable regions of the embedding from those that are less stable, providing insight into where the algorithm consistently recovers structure. Importantly, regions with low stability scores may also correspond to areas of higher biological uncertainty, where the underlying structure is more ambiguous or less well-defined. By highlighting both algorithmic variability and potential ambiguity in the data, these measures support more informed interpretation of local structures in the embedding. For example, in the Murine Intestinal dataset, NESS(UMAP) under the recommended GCP value (GCP=40, or ``High GCP" in Figure \ref{fig.3}a), we identified a group of Enteroendocrine Progenitor (EE progenitors) cells whose low-dimensional embeddings exhibit greater local instability compared to other cells within the same embedding (Figure \ref{fig.3}a, bottom right panel; Supplement Figure \ref{supp.fig.7}), indicating higher variability or biological uncertainty  as compared with other cells.
Additionally, in the Mouse Hema, Murine Intestinal, and Embryoid Body datasets, we find that removing the top 5–20\% most unstable cells, identified by low {NESS(t-SNE/UMAP) local stability scores (see Methods)}, consistently improves embedding quality. These unstable cells likely contain higher levels of noise or subtle biological signals not aligned with the primary trajectory. Their removal thus enhances both the preservation of smooth structural patterns (measured by Concordance) and the biological signal (measured by Neighbor Purity) in the remaining embeddings, resulting in improved overall stability (Figure \ref{fig.3}b). The stratification of cells according to their local stability measures can thus refine the identification of global smooth structures. Moreover, across various biological datasets {and different NE algorithms}, we observe a strong association between NESS(t-SNE/UMAP) local stability and local cell density, where the cells with low stability tend to have greater distances from their nearest neighbors than expected (Figure \ref{fig.3}d, Methods). This suggests that highly unstable cells often reside in low-density regions of the feature space (Supplementary Figure \ref{supp.scatter}). If we assume continuity of the cell state changes, such low-density regions would therefore consist of cells undergoing fast state transitions. This connection allows us to interpret NESS local stability scores not only as a measure of the technical stability of NE algorithms but also as an indicator of the biological stability of cell states, providing valuable biological insights, as discussed below.

\noindent{{\it \emph{NESS embedding rareness score avoids technical artifacts from RI.}}} For any NE algorithm with NESS recommended hyperparameters, in order to choose a suitable low-dimensional embedding among the multiple (e.g., 50) embeddings obtained under different RIs, NESS also generates an {``embedding rareness" score} for each embedding, quantifying how similar each embedding is to the ensemble of embeddings generated from multiple RIs. It has been observed that even with a well-chosen GCP, a poor embedding may occasionally arise by mere chance {from RI} \cite{cai2022theoretical}.
\red{Here, under the default GCP, the low-dimensional embedding of the Mouse Hema dataset produced by t-SNE exhibits clear distortions (Figure \ref{fig.3}d, right). Specifically, Lymph and DC cells, which are originally closer in expression space to MEPs, GMPs, and a small subset of BASO and Mono cells than to other cell types, become dispersed across distinct regions in the embedding. The rareness score can help avoid such pathological cases caused by RI, thereby enabling the selection of more reliable embeddings for improved visualization (Figure \ref{fig.3}d, left).}

\noindent{{{\it \emph{NESS-assisted NE algorithms improves standard NE algorithms under default GCPs.}}}
We observe that while conventional application of NE algorithms under their respective default hyperparameters appeared to effectively capture the underlying biological structure in some datasets, such as Murine Intestinal dataset (Supplement Figure \ref{supp.fig.9}), it may introduce significant artifacts in other datasets. \red{For example, applying t-SNE with its default GCP (\texttt{perplexity}=30 in R) to the Mouse Hema dataset led to artificial fragmentation of erythrocytes (ERY) into  small clusters, and separated dendritic cells (DC) from the main developmental trajectory (Figure \ref{fig.4new}a, artifacts highlighted in red circles).  Default applications of PhateR (default GCP \texttt{knn}=5 in R), t-SNE (default GCP \texttt{perplexity}=30 in R), and UMAP (default GCP \texttt{n\_neighbors}=15 in R) to the iPSC dataset generated low-dimensional embeddings that fail to capture the lineage relationship (Supplementary Figure \ref{supp.fig.4}b) between the primitive streak population at day 2-2.5, and the downstream differentiated endodermal and mesodermal cells at day 3 (Figure \ref{fig.4new}b, artifacts highlighted in red circles). Similarly, applying t-SNE under default GCP to the bmcite dataset also led to artificial fragmentation of regulatory T cells (Figure \ref{fig.4new}c, artifacts highlighted in red circles).  In contrast, NESS-assisted embeddings do not exhibit similar distortions.}   \red{In addition, we benchmark NESS against EMBEDR, scDEED, and DynamicViz on both simulated and real datasets. Across these experiments, NESS performs on par with or outperforms existing methods in selecting appropriate GCP values (Supplementary Figure \ref{supp.bench}).} These results highlight the potential benefits  from  integrating NESS into routine single-cell analysis pipelines that utilize NE algorithms.}

\paragraph{NESS identifies transitional and stable cell states in diverse biological developments.} We show that NESS {combined with NE algorithms such as t-SNE or UMAP under the NESS recommended GCP} identifies transitional and stable cell populations through its local stability scores (Methods), providing biological insights into cellular differentiation and lineage progression. 

\noindent{\it \emph{Validating NESS local stability score using cell transition entropy and cell-state annotations.}}
To validate our method, we use the transition entropy score computed by MuTrans \cite{zhou2021dissecting}, an algorithm that directly models cell-fate transition dynamics based on biophysical principles. MuTrans computes a transition entropy score for each cell, quantifying its likelihood of being in a transition state, with  higher values indicating stronger evidence of intermediate cell states with mixed identities. We focus on the iPSC and Murine Intestinal datasets, on which MuTrans’ performance was certified in the original publication \cite{zhou2021dissecting}. We validate NESS’s performance by comparing the local stability score from {NESS(t-SNE)} with MuTrans' transition entropy scores. As another validation, we evaluate the local stability score from {NESS(UMAP)} on the Embryoid Body dataset where the detailed cell state annotations are available \cite{moon2019visualizing}.
To benchmark the performance and demonstrate the advantages of NESS, we compare its results with existing embedding assessment algorithms, including EMBEDR, DynamicViz, scDEED and CellTran \cite{wang2024transition}.

\begin{figure}
    \centering
    \includegraphics[width=1\linewidth]{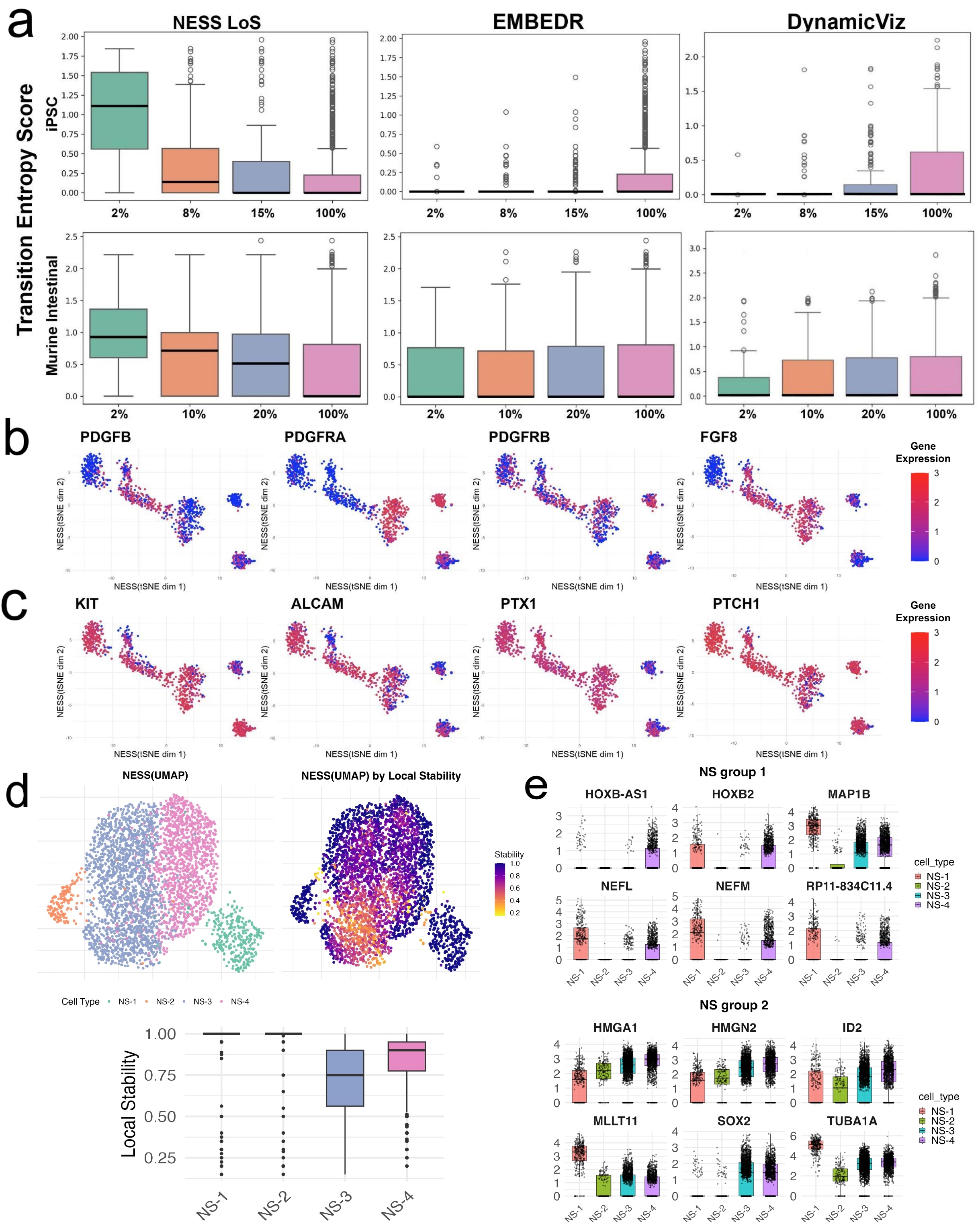}
    \caption{(Caption next page.)}
    \label{fig.4}
\end{figure}

We observe a strong association between the NESS {local stability (LoS) score} and the transition entropy score. Specifically, in both iPSC and Murine Intestinal datasets, cells in the bottom percentile with lower {local stability scores} tend to have higher transition entropy scores, suggesting that the cells demonstrating greater variability in the NE embeddings are more likely in transitional states. \red{In contrast, alternative embedding assessment methods do not exhibit the same level of associations (Figure \ref{fig.4}a and Supplementary Figure \ref{supp.fig.entropy})}. 

 {To further validate NESS, we apply it to the Embryoid Body dataset, comparing the  NESS(UMAP) local stability score across a spectrum of cell types and cell states, from pluripotent stem cells (ESCs) to lineage-committed progenitors (Supplement Figures \ref{supp.fig.4}d and \ref{supp.fig.8}b). The annotated cell states and the lineage relationships among them were previously characterized and experimentally validated in \cite{moon2019visualizing}.  Consistent with expected developmental hierarchies, intermediate cell states (e.g., NE-1/NS-5 and EN-1) exhibit lower local stability scores, while self-renewing ESCs and terminally differentiated populations (e.g., CPs, EPs, SMPs) display higher stability scores (Supplement Figures \ref{supp.fig.8}a).} 
As a comparison, alternative methods such as EMBEDR and scDEED do not appear to distinguish transitional and stable cell states; while DynamicViz is able to similarly identify ESC, EPs and SMPs as stable cell states, it does not capture the relative stability of CPs (Supplement Figure \ref{supp.fig.8}a).

\addtocounter{figure}{-1}
\begin{figure} [t!]
  \caption{\small (Previous page.) NESS identifies transitional and stable cell states in diverse biological developments. (a) Boxplots of MuTrans' transition entropy scores for cells whose NESS(t-SNE) local stability (LoS) score, and the pointwise stability measures from EMBEDR, and DynamicViz (see Methods) are at the bottom $2\%, 8\%, 15\%$ and $100\%$ percentile. Higher entropy score indicates greater likelihood of being in a transition cell state. Top row: iPSC dataset. Bottom row: Murine Intestinal dataset. Cells with low NESS local stability scores tend to have higher transition entropy score. 
  These results suggest the advantage of NESS local stability score in consistently identifying transitional cell states compared with EMBEDR and DynamicViz. \red{See Supplementary Figure \ref{supp.fig.entropy} for additional comparisons.} (b) NESS enhanced t-SNE embeddings of the iPSC dataset, colored by the expression level of \textit{PDGFB}, \textit{PDGFRA}, \textit{PDGFRB}, and \textit{FGF8}, whose expression is negatively correlated with NESS local stability. (c) NESS-assisted t-SNE embeddings of the iPSC dataset, colored by the expression level of \textit{KIT}, \textit{ALCAM}, \textit{PTX1}, and \textit{PTCH1}, whose expression is positively correlated with NESS local stability. (d) NESS-assisted UMAP visualization of neuronal subtypes NS-1 to NS-4 from the Embryoid Body dataset. The top left panel shows cell types, while the top right panel displays NESS local stability scores. Notably, NS-3 and NS-4 cells exhibit lower local stability.
(e) Boxplots of gene expression in NS-1 to NS-4 cells.
 Genes are grouped into two categories based on expression patterns: Group 1 includes genes for which NS-4 expression resembles NS-1, while Group 2 includes genes for which NS-4 expression resembles NS-3.
These results suggest that NS-4 and NS-3 represent progenitor-like or intermediate cell types, while NS-1 and NS-2 exhibit more stable, differentiated identities, consistent with their local stability scores, respectively.}
\end{figure}

\noindent{\it \emph{NESS helps identify key genes associated with iPSC differentiation.}} To demonstrate the implication of the NESS-identified transitional and stable cell states in downstream analysis, we evaluated associations between gene expression and {the NESS(t-SNE) local stability} across all cells in the iPSC dataset. Our analysis detected 48 key genes whose expressions are significantly associated with NESS local stability score (see Methods, Table \ref{supp.table.2}). Among them,  \textit{PDGFB} (adjusted p-value = $3.17\times 10^{-3}$), \textit{PDGFRA} (adjusted p-value = $1.4\times 10^{-4}$), \textit{PDGFRB} (adjusted p-value = $3.07\times 10^{-3}$) and \textit{FGF8} (adjusted p-value = $3.97\times 10^{-9}$) were top genes negatively correlated with NESS local stability. As expected, these genes display relatively higher expression in the less stable, transitional cell states (Figure \ref{fig.4}b and \ref{supp.fig.4}b),  highlighting both the transition from the initial epiblast to the  primitive streak population 
(as in \textit{PDGFB} and \textit{PDGHRB}), and the subsequent differentiation of the primitive streak  into endodermal and mesodermal cell types (as in \textit{PDGFRA} and \textit{FGF8}) \cite{bargaje2017cell}. In contrast, \textit{KIT} (adjusted p-value = $2.08\times 10^{-2}$), \textit{ALCAM} (adjusted p-value = $3.23\times 10^{-4}$), \textit{PTX1} (adjusted p-value = $6.2\times 10^{-4}$) and \textit{PCH1} (adjusted p-value = $3.45\times 10^{-3}$), among others, were positively correlated with NESS local stability score, showing higher expression in the more stable cell states (Figure \ref{fig.4}c), which mainly consist the complementary, non-transitional populations. In particular, among genes whose expression is localized in transitional cell states, \textit{PDGFB}, \textit{PDGFRA} and \textit{PDGFRB} are associated with the Platelet-Derived Growth Factor (PDGF) signaling pathway, which is essential for mesodermal differentiation, influencing vascular development, mesenchymal stem cell formation, and cardiac lineage commitment \cite{betsholtz2004insight,andrae2008role,hoch2003roles}; \textit{FGF8} is a key regulator of neurodevelopment, driving the differentiation of iPSCs into neural progenitors and contributing to midbrain formation \cite{crossley1996midbrain, toyoda2010fgf8,offen2023enrichment}.
Among genes whose expression is localized in more stable iPSC cell states, \textit{KIT} (c-KIT) is a cytokine receptor expressed on the surface of hematopoietic stem cells as well as other cell types, which plays a role in cell survival, proliferation, and differentiation \cite{lennartsson2012stem}; activated leukocyte cell adhesion molecule (\textit{ALCAM}) is known as a marker of fetal mouse and human iPSC-derived hepatic stellate cells \cite{asahina2009mesenchymal,koui2017vitro}.
Moreover, the enrichment analysis of the identified 48 genes highlighted important pathways such as MAPK signaling (p-value = $3.4\times 10^{-10}$), notch signaling (p-value = $4.9\times 10^{-7}$), cytokine-cytokine receptor interaction (p-value=7.1$\times 10^{-4}$), and Wnt signaling (p-value = 3.8$\times 10^{-3}$). 
These are known to be key pathways in regulating iPSC proliferation, differentiation and reprogramming \cite{garay2022dual,ichida2014notch,vethe2019effect,demine2020pro}. {These results demonstrate NESS’s ability to uncover biologically relevant transcriptional programs governing cell-state transitions.

\noindent{\it \emph{NESS resolves distinct neuronal subpopulations during embryoid formation.}} Finally, we show that the NESS(UMAP) local stability score helps to resolve distinct neuronal subpopulations during embryoid formation and provides new insights into their cell states.
In the Embryoid Body dataset, prior analyses identified four neuronal subtypes (NS-1 to NS-4) but were unable to resolve the finer structure of their respective cell states or the relationships among these subpopulations \cite{moon2019visualizing}. In our analysis, neuronal subtypes NS-3 and NS-4 display lower {local} stability scores compared to NS-1 and NS-2, suggesting the presence of subpopulations within NS-3 and NS-4 that demonstrate transcriptional mixing between them (Figure \ref{fig.4}d). To further investigate the molecular basis of this instability, we performed differential gene expression analysis between NS-1/2 and NS-3/4, identifying genes with adjusted p-values less than 0.001. This analysis revealed two distinct gene groups (Figure \ref{fig.4}e). In the first group, genes show higher expression in NS-1, moderate expression in NS-4, and minimal or no expression in NS-2 and NS-3; for example, we have \textit{NEFM} (Neurofilament Medium Chain), which encodes an intermediate filament essential for maintaining neuronal caliber, axonal transport, and structural integrity of neurons \cite{yuan2017neurofilaments}. In the second group, genes exhibit similar expression patterns between NS-3 and NS-4 but are distinctly expressed compared to NS-1 and NS-2. An example is \textit{SOX2}, a transcription factor essential for maintaining neural progenitor identity and regulating early neural differentiation \cite{wang2012distinct}, which is highly expressed in NS-4. This suggests that NS-4 may correspond to a progenitor-like population. NS-3 displays intermediate expression patterns, consistent with a transitional cell state, while NS-1 and NS-2 express neuronal maturation markers such as \textit{NEFM}, indicating more differentiated identities.
Together, these observations support a model in which NS-4 and NS-3 represent transcriptionally unstable or transitional populations, aligning with their lower local stability scores.



\paragraph{NESS reveals transcriptional dynamics during spermatogenesis and neurogenesis.} Building upon the above connection between the NESS local stability score and cellular transitivity, with the assumption that cells undergoing cell-state transitions likely exhibit a higher rate of change in the transcriptional activity, {we show that NESS combined with t-SNE or UMAP can also serve as a proxy for inferring transcriptional dynamics using only scRNA-seq data. To illustrate this utility, we analyze two additional scRNA-seq datasets, one on the sermatogenesis, and the other on the neurogenesis, with NESS(UMAP).} As a validation step, we compare the NESS local stability score with cell-specific RNA velocity estimates independently obtained using scVelo \cite{bergen2020generalizing}, which incorporate additional information from RNA splicing kinetics to quantify transcriptional activity. In particular, for each cell, we define total RNA velocity as the $L_2$-norm of the estimated RNA velocity vector across the top 2,000 most variable genes, quantifying the overall rate of change in transcription activity.

For the Spermatogenesis dataset, our analysis reveals a remarkable similarity between {the patterns of pointwise NESS instability score, defined as 1/(NESS local stability score),} and the total RNA velocity. Among the four major stages of spermatogenesis, cells in the spermatogonia and early spermatocyte stages exhibit greater NESS local instability and higher total RNA velocity, while cells in the late spermatocyte and spermatid stages display lower and moderate values, respectively (Figure \ref{fig.5}a, b). This pattern aligns with current biological understanding of germ cell development. Specifically, spermatogonia and early spermatocytes undergo rapid mitotic divisions and initiate meiosis, likely contributing to elevated transcriptional activity \cite{griswold2016spermatogenesis}. In contrast, as cells progress to late spermatocytes and spermatids, global transcription tend to decline and stabilize, as post-meiotic cells undergo extensive morphological changes, driven largely by pre-synthesized transcripts and post-transcriptional regulation \cite{braun1998post}.
Notably, within the spermatid subpopulation, we observe a gradual rise and subsequent decline in NESS  instability score along the progression trajectory (Figure \ref{fig.5}b, c). This dynamic likely reflects the transition from active gene expression in early spermatogenesis to transcriptional quiescence in later stages, facilitating efficient germ cell maturation through post-transcriptional mechanisms.

\begin{figure}[h]
    \centering
    \includegraphics[width=1\linewidth]{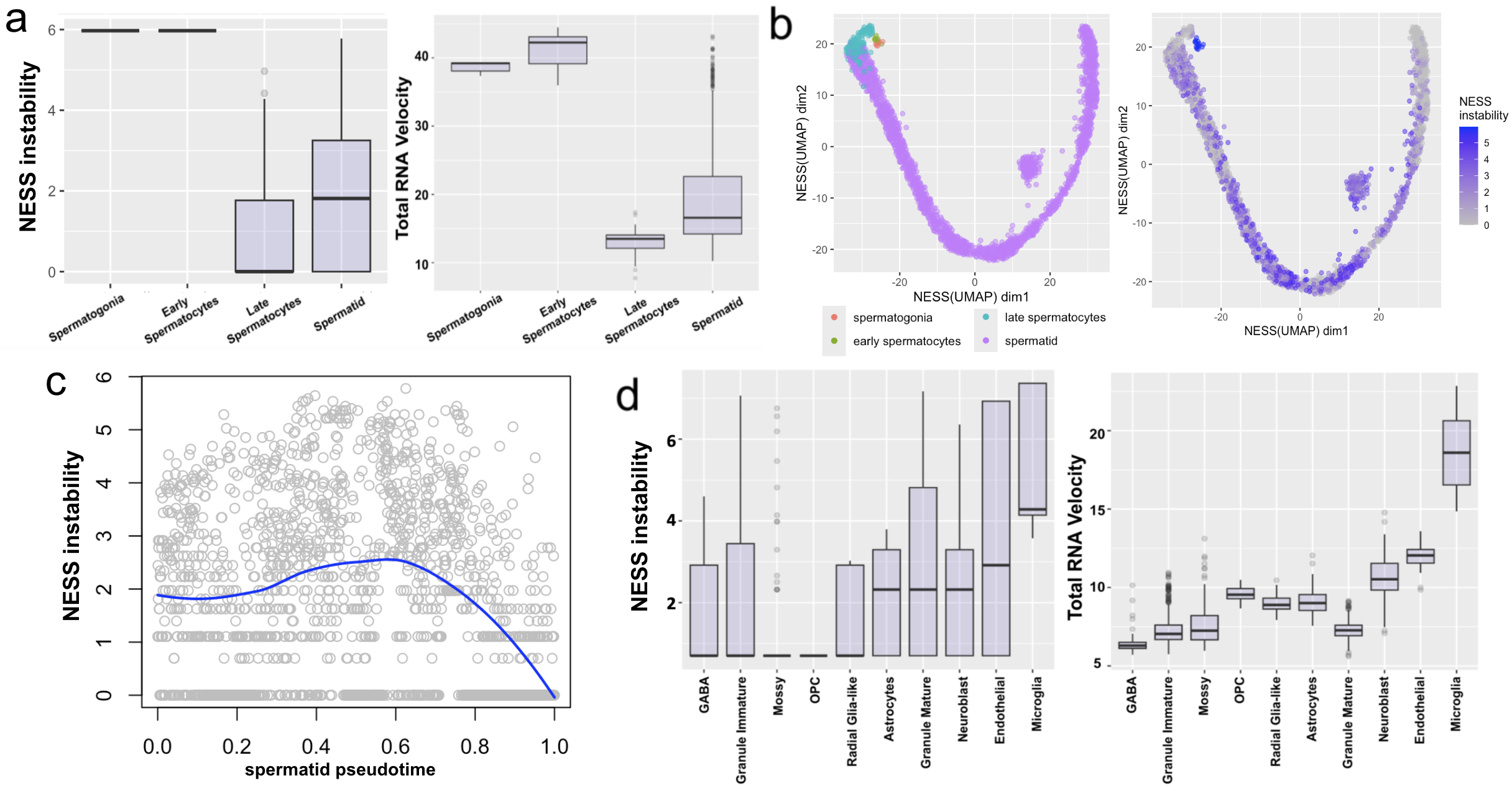}
    \caption{\small NESS reveals transcriptional dynamics during spermatogenesis and neurogenesis. (a) Comparison of pointwise NESS instability score, defined as 1/(NESS local stability score), with total RNA velocity for the Spermatogenesis dataset. Each boxplot contains the cells within a specific cell type. The total RNA velocity is defined as the $L_2$-norm of the RNA velocity vector across the top 2000 most variable genes, independently obtained by scVelo, which incorporate additional information from RNA splicing dynamics. The plots demonstrate remarkable similarity  between the patterns of NESS-TAG scores and the total RNA velocity across different cell tyles. (b) NESS-assisted UMAP visualization of the Spermatogenesis dataset, colored by cell types (left) and NESS instability score (right). (c) NESS instability score of the spermatid cells ordered by pseudotime along the cell progression trajectory in (b). The blue line indicates the loess fitted regression curve. (d) Comparison of NESS instability score with total RNA velocity for the Dentate gyrus neurogenesis  dataset. The boxplots are ordered by the median of NESS instability within each cell type. The results indicate notable concordance between the patterns of the two measures across cell types.}
    \label{fig.5}
\end{figure}

For the Dentate gyrus neurogenesis dataset, we also observe notable concordance between the patterns of NESS(UMAP) instability score and the total RNA velocity across different cell states (Figure \ref{fig.5}d). The greater variability in the NESS instability score within each cell type compared with the total RNA velocity is likely due to the higher noise level in scRNA-seq data when inferring transcription dynamics, relative to the spliced RNA data used to estimate RNA velocity.  In particular, our analysis highlights the relatively higher transcriptional activity of microglia and endothelial cells, while GABAergic neurons and immature granule cells exhibit lower transcriptional activity. These patterns align with known cellular functions and developmental dynamics of neurogenesis. Specifically, microglia and endothelial cells play crucial roles in immune response, neuroinflammation, and vascular remodeling, necessitating continuous gene expression and rapid transcriptional adaptation \cite{masuda2020microglia,crouch2023disentangling}. In contrast, immature granule cells are known to develop very slowly \cite{zhou2022molecular}, allowing for greater plasticity and adaptability during their maturation phase \cite{pardi2015differential}. Similarly, GABAergic neurons, which are largely post-mitotic, experience transcriptional stabilization following differentiation \cite{lim2018development}. These findings underscore the biological relevance of NESS in capturing transcriptional dynamics across different cell states in neurogenesis.


\section{Discussion}

Overall, we develop a unified approach NESS for assessing and improving NE algorithms based on PCS principles, enabling more reliable visualization and interpretation of smooth structures in single-cell data while offering valuable biological insights into cell development. A key strength of the NESS local stability score lies in its design principle—leveraging random initialization to introduce algorithmic perturbations, allowing a systematic evaluation of embedding stability. {Compared with existing embedding evaluation algorithms, NESS shows three key  advantages.} {First, the NESS evaluation is not restricted to any particular instance of embedding.} In contrast, a baseline method scDEED assesses neighborhood structures by comparing a single realization of a (potentially random) low-dimensional embedding with the original dataset. Consequently, the inferential insights from scDEED are limited to that particular embedding instance and may not readily generate additional insights into the input dataset, such as local density variations and cell transition stability. {Second, NESS’s algorithmic perturbation strategy is more effective at handling rare cell types and computationally more efficient than sample-perturbation-based methods.} For example, approaches like DynamicViz and EMBEDR generate uncertainty estimates by resampling the entire input data. \red{While this enables assessment of embedding variability, such methods tend to emphasize well-represented cell states with sufficient sample sizes, often underrepresenting rare populations due to low sampling rate. RI-based perturbations instead probe the sensitivity of learned neighborhood structure to optimization trajectories under fixed data, allowing unstable or weakly supported structures to be identified even when few cells are present.} 
{Finally, NESS is broadly applicable and can be integrated with any NE algorithm.} In contrast, techniques like LOO require detailed knowledge of specific embedding algorithms, such as their objective functions and explicit Hessian expressions, which restricts their utility to a narrow subset of methods such as t-SNE and UMAP. 

\red{Our work  extends the PCS framework in two ways. First, although stability principles have previously been used to improve modeling in unsupervised learning \cite{yu2024veridical,wu2016stability}, the present work focuses specifically on their application to nonlinear embedding. Second, while Predictability in PCS is most naturally defined in supervised settings, in unsupervised contexts such as NE it is interpreted as a ``reality check," assessing whether learned representations recover known biological structure, align with established trajectories, and preserve meaningful relationships. More broadly, we view PCS in unsupervised learning as a framework for constraining the space of plausible representations through stability under perturbations and consistency with domain knowledge, enabling the identification of structures that are both reproducible and scientifically meaningful, while acknowledging that no single optimal solution can be defined without task-specific criteria.}


\red{Although our theoretical and empirical results indicate that underspecifying the GCP can substantially degrade embedding quality, and that moderate overspecification is often less harmful, our goal is not to advocate replacing one default value (e.g., perplexity = 30) with another larger fixed value (e.g., perplexity = 100). The optimal GCP depends on dataset-specific trade-offs among embedding fidelity, resolution of local structure, and computational cost; no single value performs uniformly well across diverse biological systems. In some datasets, such as Mouse Hema, a larger-than-default GCP improves stability and better recovers the underlying biological structure. In others, however, excessive GCP can oversmooth the representation and artificially connect distinct clusters, obscuring meaningful heterogeneity. For example, in a CD8+ T-cell scRNA-seq dataset \cite{10x2019new}, the t-SNE embedding with perplexity 100 merged visually distinct clusters, whereas the NESS-selected perplexity (=10) preserved separation while maintaining stability (Supplementary Figure \ref{supp.fig.over}). This highlights the limitations of a simple “increase GCP” heuristic. Importantly,  NESS is not merely a hyperparameter selection tool: it also provides pointwise diagnostics of reliability for the final chosen embedding, and inform downstream biological inferences.}

NESS has some limitations that {we plan to address in future work}. First, one assumption behind NESS is that the underlying smooth structure of interest should be intrinsically low-dimensional, so that the low-dimensional embeddings do not face the fundamental impossibility of embedding a higher-dimensional manifold into a lower-dimensional space. This assumption may be violated in complex biological processes where cell development cannot be adequately represented by a low-dimensional cell-state manifold. {To address this challenge, we aim to develop methods for estimating the intrinsic dimensionality of high-dimensional datasets, providing a principled way to assess the suitability of NESS in a given context.} Second, NESS relies on existing NE algorithms to generate low-dimensional embeddings and is thus inherently constrained by their individual limitations, even though it can select the most suitable algorithm among them (Supplement Figure \ref{supp.fig.6}). A promising next step is to develop an ensemble approach \cite{ma2023spectral} that, after reality-check through ``P" in PCS, combines the strengths of different Pred-checked NE algorithms as well as other embedding algorithms, such as Laplacian eigenmap \cite{belkin2003laplacian} and diffusion map \cite{lafon2006data}, to improve the overall quality of the low-dimensional embeddings.


 \section{Methods}


\subsection{PCS-Guided Assessment and Validation of NE Algorithms}

Our PCS-guided reality check of NE algorithms involves two parts. On the one hand, we validate the preservation of biological information in the low-dimensional embedding using  known biological labels of the cells, such as cell type and cell state annotations, associated with the benchmark single-cell datasets. On the other hand, we evaluate the preservation of local and global structures of the original datasets in the low-dimensional embeddings. In each case, we consider multiple evaluation metrics.

\paragraph{Three biology-preservation metrics.}
For a given low-dimensional embedding and a set of cell-specific biological labels, we consider the following three evaluation metrics.
The Silhouette index measures the cohesion and separation of clusters \cite{rousseeuw1987}. It is formally defined as
$\frac{1}{n} \sum_{i=1}^n \frac{b_i - a_i}{\max(a_i, b_i)},$
where \( a_i \) is the average distance to points in the same cluster and \( b_i \) is the average distance to points in the nearest cluster. A higher Silhouette Score indicates better alignment between the data points and the cluster labels, with points being closer to their own clusters than to other clusters. The function \texttt{silhouette()} in the R package \texttt{cluster} is used.
The neighbor purity score measures the proportion of a cell's $k$-nearest neighbors belonging to the same group as that cell. A higher neighbor purity score indicates that the groups are more homogeneous. The function \texttt{neighborPurity()} in the R package \texttt{bluster} is used, with $k=50$.  
The local Simpson score is defined as the reciprocal of the Local Inverse Simpson index (LISI), defined in \cite{korsunsky2019fast} and implemented as the function \texttt{compute$\_$lisi()} in the R package \texttt{lisi}. LISI estimates the effective number of clusters contained in a cell's neighborhood. As such, a higher local Simpson score indicates better separation of the clusters.

\paragraph{Two structure-preservation metrics.} For a given pair of input high-dimensional dataset and its low-dimensional embedding, we consider the following two evaluation metrics.
The correlation score measures the similarity between pairwise distances in the original and reduced embedding spaces, defined as $ \text{corr}\left(\text{dist}(\mathbf{Y}), \text{dist}(\mathbf{X})\right)$,
where \(\text{dist}(\mathbf{X}) \) is the pairwise Euclidean distance among cells computed from the high-dimensional dataset and  \(\text{dist}(\mathbf{Y}) \) is the pairwise Euclidean distance among cells in the low-dimensional embedding. A higher correlation score indicates that the low-dimensional embedding preserves pairwise distances more faithfully.
The (neighbor) concordance score quantifies how well the neighborhood structure is preserved between the high-dimensional data (\( \mathbf{X} \)) and the low-dimensional embedding (\( \mathbf{Y} \)). Formally, it is defined as
\[
\text{Concordance} = \frac{1}{n} \sum_{i=1}^n \frac{\left| \mathcal{N}_i^{\text{high}} \cap \mathcal{N}_i^{\text{low}} \right|}{k},
\]
where \( n \) is the total number of data points in the dataset, \( \mathcal{N}_i^{\text{high}} \) is the set of \( k \)-nearest neighbors of point \( i \) in the high-dimensional space, \( \mathcal{N}_i^{\text{low}} \) is the set of \( k \)-nearest neighbors of point \( i \) in the reduced-dimensional space, and \( k \) is the number of nearest neighbors considered. A higher concordance score indicates better preservation of neighborhood relationships between the original and reduced-dimensional spaces. We used $k=100$ in all our analyses.

\red{\paragraph{Evaluation at both local and global scales.} We note that some evaluation metrics rely on nearest neighbors (e.g., concordance with $k=100$). These metrics are intended to measure how well an embedding preserves local relationships at a specified scale. Importantly, the purpose of this analysis is to examine the potential sensitivity of embedding performance to different GCP choices rather than to argue that default GCP values are universally inadequate. To provide a more balanced and scale-aware evaluation, we assess embeddings using both local metrics (e.g., concordance and neighbor purity) and global metrics (e.g., correlation and silhouette score). The global metrics do not depend on nearest-neighbor parameters and therefore offer scale-independent assessments of overall structure. By combining local and global measures, we aim to  comprehensively illustrate how varying GCP values can influence the resulting embeddings, particularly highlighting potential issues associated with overly small GCP values. }

\paragraph{Simulated datasets.} We generate simulated datasets containing smooth structures to evaluate the performance of NE algorithms using the metrics introduced above. In the first case, we randomly sampled $n = 2000$ data points from a one-dimensional curve embedded in high-dimensional space. The curve is divided into five equal-length segments, and each sample is assigned a ``pseudotime" label (from 1 to 5) based on the segment to which it belongs. In the second case, $n=2000$ data points were randomly sampled from the unit circle. In both examples, the simulated datasets contain a low-dimensional (smooth) manifold structure.

\subsection{Theoretical Insights on NE Algorithm Artifacts}

{To better understand the nature of the empirically observed artifacts in NE algorithms with default parameters, we conduct a rigorous theoretical analysis of t-SNE, the most basic NE algorithm. We formally establish the connection between low GCP values and these artifacts, particularly the fragmentation of the underlying manifold structure, in the final low-dimensional embeddings. We expect our findings about t-SNE to generalize to other NE algorithms, given the similarities in their loss functions and iterative optimization schemes \cite{bohm2022attraction,kobak2021initialization,damrich2021umap}.}

Recall that the t-SNE algorithm starts by constructing affinities $p_{ij}$ for each pair of data points  by
$ p_{ij} = \frac{p_{i|j} + p_{j|i}}{2n},$
where
$ p_{i|j} = \frac{\exp\left( -\|x_i - x_j\|^2/(2\sigma_i^2)\right)}{\sum_{k \neq i}  \exp\left( -\|x_i - x_k\|^2/(2\sigma_i^2)\right)},$
and $\sigma_i$'s are bandwidth parameters determined through computational procedures such as perplexity quantification detailed in \cite{van2008visualizing}. On the other hand, t-SNE also defines embedding affinities
$ q_{ij} = \frac{(1+\|y_i - y_j\|^2)^{-1}}{\sum_{k \neq \ell} (1+ \|y_k - y_{\ell}\|^2)^{-1}},$
where $y_i$'s are the desirable low-dimensional embeddings.
The objective of t-SNE is to minimize the KL-divergence between affinities $\{p_{ij}\}$ and $\{q_{ij}\}$, given by $
C^{**}(y_1,y_2,...,y_n) = -\sum_{i \neq j} p_{ij} \log\left( \frac{p_{ij}}{q_{ij}}\right).$
Since the affinities $\{p_{ij}\}$ are fixed for given data set $\{x_i\}_{1\le i\le n}$, the above optimization problem is equivalent to solving for $\{y_i\}$ that maximize
\beq \label{obj}
C^{*}(y_1,y_2,...,y_n) = \sum_{i \neq j} p_{ij} \log(q_{ij}).
\eeq
As we argued earlier, an important parameter determining the performance of t-SNE is the graph connectivity characterizing the smoothness of the neighborhood affinity around each point. In the case of t-SNE, the graph connectivity is implicitly determined by bandwith parameter $\sigma_i$, or  ``perplexity" in most software implementations. 

To provide deeper insights on the reason behind the empirically observed distance-distortion of NE algorithms for embedding manifold-like structures under lower GCPs, we consider the following \emph{prototypical case} where t-SNE is applied to embed data points uniformly distributed on a unit circle in $\R^2$.
Despite the simplicity of the example, we believe that our analysis can be generalized to all one-dimensional manifolds. Since this is beyond the scope of the current work, the fully generalized theoretical results will be developed elsewhere in a more systematic manner. 

Consider the discrete circle of $n$ points 
$ P_n = \left\{ \left( \cos\left( \frac{2\pi i}{n}\right),  \sin \left( \frac{2\pi i}{n}\right) \right): 1 \leq i \leq n \right\}$ in $\mathbb{R}^2$.
Our goal is to prove that, when the graph connectivity parameter is relatively small compared with the sample size $n$, the \emph{optimal} t-SNE embedding $\phi: P_n \rightarrow \mathbb{R}^2$ must distort distances, and the distortion becomes even bigger as $n$ increases. 

To facilitate our analysis of t-SNE, \red{we consider the following simplification of t-SNE for embedding the above discrete circle $P_n$, where $\sigma_i=\sigma$ is constant and therefore the affinities $p_{ij}$ are approximately given by the $k$ nearest neighbors
\beq\label{pij}
p_{ij} = \begin{cases} 1/(2 nk) \qquad \qquad &\mbox{if}~x_i, x_j~\mbox{are neighbors} \\ 0 \qquad &\mbox{otherwise.} \end{cases}
\eeq
}Here the parameter $k$ characterizes explicitly the graph connectivity in the affinity matrix $P=(p_{ij})$.

The natural mathematical framework that describes metric distortion is that of bilipschitz embeddings. Specifically, a mapping $\phi:X \rightarrow Y$ is bilipschitz with distortion parameter $L \geq 1$ and scaling factor $S>0$ if
\beq \label{bilip}
\forall~x_1, x_2 \in X \qquad S \leq \frac{\| \phi(x_1) - \phi(x_2) \|}{\| x_1 - x_2\|} \leq SL.
\eeq
Intuitively, a diverging distortion parameter $L$ suggests that the mapping $\phi$ is not Lipschitz, essentially containing  points of discontinuity, or ``gaps" in its image.
The following theorem explains the fragmentation patterns empirically observed in the final visualizations.

\begin{thm}[t-SNE artifact] \label{tsne.thm}
\red{Let $P_n$ be the discrete circle and let $\phi^*$ be the optimal embedding induced by the t-SNE objective (\ref{obj}) with the affinities (\ref{pij}) and the graph connectivity parameter $k$ satisfying $k=O(1)$ as $n\to\infty$. If $\phi^*$ is bilipschitz with distortion parameter $L(\phi^*)$ and scaling factor $S(\phi^*)$, such that $S(\phi^*)=n^\tau$ for some $0<\tau<1$, then, the distortion factor of $\phi^*$ is unbounded, that is,
$ L(\phi^*) \to \infty$ as $n \rightarrow \infty$.}
\end{thm}
\begin{proof}
See Section \ref{sec.proof} of the Supplement.
\end{proof}

\red{Theorem \ref{tsne.thm} relies on two key assumptions. First, we assume that the local graph connectivity $k$ is finite and does not grow with the sample size $n$. This assumption is motivated by our interest in understanding the behavior of t-SNE when the perplexity parameter is underspecified. Recall that perplexity is defined as $2^{H(P_i)}$, where $H(P_i)$ is the Shannon entropy of the conditional distribution $P_i = (p_{j \mid i})$. A smaller perplexity typically leads to smaller bandwidths $\{\sigma_i\}$ in the affinity matrix $P = (p_{ij})$, and consequently to a smaller effective neighborhood size $k$. The finite-$k$ assumption therefore reflects the regime in which perplexity does not scale with $n$. Second, the theorem requires that the scale factor of the final embedding diverges with $n$, but remains $o(n)$. This condition is needed for the technical derivation. Intuitively, it suggests that the diameter (or overall size) of the final t-SNE embedding depends on the sample size and grows gradually as $n$ increases. Although we do not yet have a rigorous proof of this behavior, our numerical experiments support this assumption. For example, when generating t-SNE embeddings of a discrete circle with fixed perplexity ($=30$) across varying $n$, we observe an approximately linear relationship between $\sqrt{n}$ and the range (or diameter) of the embedding coordinates in both the $x$- and $y$-directions (Supplementary Figure \ref{supp.fig.diameter}). Theorem \ref{tsne.thm} states that whenever the above two assumptions hold, the optimal t-SNE embedding given by the solution of (\ref{obj}) will have a diverging distortion parameter $L$, thus the discontinuity and the structural distortions.} 

\subsection{NESS Approach}

{In this section, we provide details of the proposed NESS approach. We begin by describing how to generate algorithmic perturbations for any NE algorithm using random initialization, followed by the construction of a KNN matrix that captures embedding stability. We then introduce the definitions of the NESS local stability score, global stability score, and embedding rareness score. Finally, we present a data-driven NESS workflow that automatically identifies the optimal GCP.}

NESS is designed to generate stability measures of an NE algorithm for better hyperparameter tuning,  interpretation, and downstream inferences. It evaluates the stability of neighborhood graphs derived from the low-dimensional embeddings across multiple NE runs based on different random initializations. First, for a given a dataset and NE algorithm under some hyperparameter configuration, we run \( N \) times under random initialization to generate \( N \) low-dimensional embeddings, denoted as \( Y^{(1)}, Y^{(2)}, \ldots, Y^{(N)} \). For each embedding, we create a \( k \)-Nearest Neighbors (KNN) graph, represented as \( M^{(i)} \), where \( M^{(i)}[j, 1:k] \) stores the indices of the \( k \)-nearest neighbors for point \( j \). To quantify how often two points are identified as neighbors across all \( N \) embeddings, we construct a k-nearest-neighbor matrix \( \mathbf{K}=(K[j,l]) \). This matrix is initialized as a zero matrix of size \( n \times n \), where \( n \) is the number of data points. As $i$ runs from 1 to $N$, the matrix is updated as $
K[j, l] \leftarrow K[j, l] + 1$ for all $l \in M^{(i)}[j, 1:k]$.
As a result, the final \( K[j, l] \) represents the matrix entry of $\bK$ at row \( j \) and column \( l \), which stores the number of times point \( j \) and point \( l \) appear as $k$-neighbors across the \( N \) embeddings.
To ensure symmetry in the neighbor relationships, we further symmetrize \( \mathbf{K} \) by $\mathbf{K} \leftarrow \max(\mathbf{K}, \mathbf{K}^\top).$ See Section \ref{implement.sec} for the recommended values of $k$ and $N$ in our implementation.

\paragraph{NESS local stability score.} 
The local stability score measures the consistency of neighbors for each data point across all \( N \) embeddings. 
For a given point \( j \), the local stability score \( S_j \) is calculated as:
\[
S_j = \text{Quantile}_{\lambda}\left( \bigg\{ \frac{K[j, l]}{N}:  K[j, l] > 0, 1\le l\le n \bigg\} \right), 
\]
where \( \text{Quantile}_{\lambda} \) represents the $\lambda$-th percentile of the normalized neighbor counts. Intuitively, if $S_j$ is large, it means that data point $j$ shares many of the same neighbors across the 
$N$ low-dimensional embeddings. In all our analyses we used the recommended value $\lambda=0.75$, which is motivated by the expected sparsity of the count matrix $\bK$. We also found the results to be stable across a wide range $\lambda\in(0.6,0.9)$. The collection of local stability scores is denoted as $\{ S_1, S_2, \ldots, S_n \}$.

\paragraph{NESS global stability score.}
The global stability score provides an overall measure of stability for the dataset by aggregating the local stability scores. It is computed using the mean of the local stability scores $\frac{1}{n} \sum_{j=1}^n S_j$ across all cells.
A higher Global Stability score indicates that the neighbors of a point are more consistently preserved across different embeddings. {To select an appropriate GCP threshold for a given NE algorithm, we recommend choosing the low-dimensional embedding corresponding to the smallest GCP value that falls within the top 5\% of all evaluated global stability scores.}

\red{\paragraph{Robustness of NESS to the choice of $k$.} NESS uses a neighborhood size $k$ in computing stability scores. We clarify that the resulting stability assessment is not sensitive to the specific choice of $k$. As shown in Supplementary Figure~\ref{supp.fig.krobust}, on the mouse hematopoiesis dataset, setting $k=30, 50, 100,$ or $200$ produces  similar global stability curves across different GCP values. 
This indicates that the role of $k$ in stability assessment is fundamentally different from that of GCP in constructing the embedding. Intuitively, the stability metric evaluates the consistency of local neighborhood relationships across repeated runs and perturbations, whereas GCP determines how neighborhood information is incorporated into the embedding itself. Because the stability score aggregates agreement patterns across neighborhoods, it remains robust to moderate changes in the evaluation scale $k$, provided $k$ lies within a reasonable range.}

\paragraph{NESS embedding rareness score.} The NESS embedding rareness score quantifies the dissimilarity of each NESS-generated low-dimensional embedding from other $N-1$ embeddings. {We start by counting how many times two cells share the same neighbor in the embeddings.} For each pair of embedding $i$ and $j$, for each data point \( l \), we evaluate the number $w_{ij,l}$ of shared $k$-nearest neighbors and define the similarity measure $W(i,j)=\text{median}(\{w_{ij,l}/k: 1\le l\le n\})$. Repeat this for all pairs of embeddings, we obtain an $N\times N$ pairwise similarity matrix $W$. For each embedding $i$, we evaluate the mean and variance of the $i$-th column of $W$ (excluding the diagonals), denoted as $m_i$ and $v_i$. Since when both quantities are small, there is bigger dissimilarity between the $i$-th embedding and the other embeddings. We define the embedding-specific rareness score $R_i=1/(m_iv_i)$. {By this definition, we hope to highlight (through a bigger embedding rareness score) embeddings which are (i) on average more dissimilar with  other embeddings (i.e., with smaller $m_i$), and (ii) more uniformly deviated from other embeddings (i.e., with smaller $v_i$).}  A higher embedding rareness score indicates that the embeddings differ significantly from the others, likely containing more severe distortions and artifacts.

\paragraph{An automated NESS workflow.} To improve computational efficiency and practicality of our proposed method, we propose an automated NESS workflow. For a given dataset and NE algorithm, we first consider a sequence of five equally spaced GCP parameters ranging from 10 to $10 \log n$, where $n$ is the sample size. Starting from the smallest GCP, we evaluate the NESS global stability (GS) score of the corresponding low-dimensional embedding, and stop if either the GS under the current GCP exceeds 0.9, or the GS under the current GCP is not greater than that under the previous, smaller GCP by at least 5\%. The final GCP is chosen as the recommended GCP. Under this GCP, we record and output the associated local stability scores, the NESS embedding rareness score (optional), and the final low-dimensional embedding.


\subsection{Biological Case Studies to Demonstrate the Effectiveness of NESS}

{In this section, we first define a local density metric and connect it to the  NESS local stability scores obtained in four single-cell datasets, to provide an explanation of the observed local instability of NE algorithms. Then we demonstrate the effectiveness of NESS in three different biological processes in (i) identifying transitional and stable cell states, (ii) quantifying cell-specific transcription activity dynamics, and (iii) differential expression analysis and pathway enrichment analysis.}

\paragraph{Quantifying local density.} 
For each $p$ from 0 to 100, we focus on the cells whose NESS local stability score lies in the upper $p$-th percentile across all cells. Denote the indices of these cells as the set $I_p$. Then for each cell in $I_p$, we compute its mean Euclidean distance to its 30-nearest neighbors in the PCA dimension-reduced feature space. Then we normalize the mean distances by dividing them by the $50\%$ percentile of all the pairwise distances between the 30-nearest neighbors. The normalized distance characterizes the average local density of the cells whose local stability score is in the $p$-th percentile. {For each of the single-cell dataset, by evaluating the average local density of the cells corresponding to different percentiles, we found a strong association: cells with lower local stability scores tend to have lower local density, as reflected by greater average distances from their neighboring cells (Figure \ref{fig.3}c).}

\paragraph{Identifying transitional and stable cell states.} \red{The idea of using low local stability scores to identify transitional cell states is based on the following considerations. On the one hand, Figure \ref{fig.3}c suggests that low local stability typically corresponds to low-density regions in the original space. Under the assumption that cells move continuously through the feature space during development, when they undergo cell-state transitions they tend to traverse these regions more rapidly, resulting in fewer cells being captured at transitional states and thus lower local density.
On the other hand, we emphasize that local stability is related to, but distinct from, local density. As shown in Supplementary Figure \ref{supp.scatter}, the two quantities exhibit moderate marginal correlation (Pearson correlation=0.23). Beyond connecting to low density, low local stability also reflects increased sensitivity of local neighborhood structure to perturbations, a hallmark of transitional states where cellular identities are not yet well established. In contrast, stable cell states typically form coherent low-dimensional manifolds with robust neighborhood structure and therefore exhibit higher local stability.}

{To identify transitional and stable cell states in the iPSC, Murine Intestinal, and Embryoid Body datasets, we first apply NESS in combination with an NE algorithm--t-SNE for the iPSC and Murine Intestinal datasets, and UMAP for the Embryoid Body dataset. Our choice of NE algorithm is guided by our empirical observations that NESS(t-SNE) generally performs well on moderate-sized datasets, while NESS(UMAP) offers greater computational efficiency for large datasets (e.g., those with more than 20K cells).} The low-dimensional embeddings are obtained under the NESS-recommended GCP in each case. For iPSC and Murine Instestinal datasets, under the NESS recommended GCP, we obtain the NESS local stability score and compare the cell-specific local stability scores with the transition entropy scores obtained from MuTrans to generate Figure \ref{fig.4}a. In particular, we compare the entropy values of the cells associated with the bottom $p\%$ ($p=2, 10, 20, 100$) of the NESS local stability scores (that is, across the $p\%$ most unstable cells). For the Embryoid Body dataset, we first apply our algorithm to obtain the NESS local stability score for each cell in the dataset, generating the boxplot in Figure \ref{supp.fig.8}a. We then focus on the four neuronal subtypes and rerun our algorithm to obtain the low-dimensional embeddings and the local stability scores (Fig \ref{fig.4}d). Then we perform standard differential expression analysis between NS-1/2 and NS-3/4 to identify key genes driving the cell state heterogeneity (Fig \ref{fig.4}e).

\paragraph{Quantifying cell-specific transcription activity dynamics.} For both Spermatogenesis and Dentate gyrus neurogenesis datasets, we first applied NESS to determine a suitable  hyperparameter configuration of the NE algorithm (UMAP in our analysis), to obtain the corresponding NESS local stability scores $\{S_1,...,S_n\}$.
Then, for each cell $i$, we define the pointwise NESS instability score as $\{1/S_1,...,1/S_n\}$ (Figure \ref{fig.5}). For better visualization and comparison with total RNA velocity, we apply the monotone transformation $\log(1+100\log(1/S_i))$ to the instability score. 
Once computed on these datasets, the NESS  instability scores are grouped by cell states and then compared with total RNA velocity, which is defined as the $L_2$-norm of the estimated RNA velocity vector for each cell. In the Spermatogensis dataset, the spermatid pseudotime is estimated by projecting cells onto the differentiation trajectory within the NESS-assisted UMAP embedding.

\paragraph{Differential expression analysis and pathway enrichment analysis.} For the iPSC dataset, we began by identifying genes whose expression are significantly associated with the NESS(t-SNE) local stability score using Pearson correlation test. As a result, 48 genes were found significant with a BH-adjusted p-value below 0.05 (Table \ref{supp.table.2}). To identify the key biological pathways associated with these significant genes, we use the DAVID Functional Annotation Bioinformatics Microarray Analysis tool (https://davidbioinformatics.nih.gov/). The DAVID platform identified 7 significant KEGG pathways including the Wnt signaling pathway, the Notch signaling pathway, the Hedgehog signaling pathway, the Cytokine-cytokine receptor interaction, the HIF-1 signaling pathway, the TGF-beta signaling pathway, and the MAPK signaling pathway. 

\subsection{Computing Time}\label{sec.time}


\red{The primary computational cost of NESS arises from two components: (i) running the NE algorithm multiple times (e.g., 30 repetitions) to estimate stability, and (ii) evaluating multiple GCP values to identify the optimal parameter.  Importantly, our benchmark is not the runtime of a single NE execution. In practice, tuning the GCP (e.g., perplexity) already requires scanning across multiple values to obtain a reasonable embedding; such parameter search is typically unavoidable. In this sense, NESS formalizes and systematizes what careful practice would already entail. Moreover, we mitigate excessive computational burden through two key mechanisms. First, our empirical results (Figures 2 and 3) reveal a structured relationship between stability scores and embedding quality, allowing early stopping once the global stability score plateaus or declines, thereby avoiding exhaustive exploration of large GCP values. Second, although each GCP requires multiple NE runs, the dominant cost in most NE methods lies in constructing the sample–sample affinity matrix, which can be reused across repeated runs for a fixed GCP. In Supplementary Figure \ref{supp.fig.10}b, we evaluate the runtime of full NESS under t-SNE. Across datasets of varying sizes, NESS(t-SNE) scales approximately linearly with sample size. In particular, for a dataset with 10,000 cells, NESS(t-SNE) with 30 repetitions required approximately 0.8 hours on a standard desktop; for 100,000 cells, it required approximately 3.4 hours.}
In Supplementary Figure \ref{supp.fig.10}v, we  evaluate  the runtime of NESS(UMAP) and compare to that of MuTrans, a popular method for modeling cell state transitions, across datasets with varying numbers of cells. While NESS(UMAP) is slightly slower than MuTrans for datasets with fewer than $1,000$ cells, it demonstrates significantly better scalability for larger datasets, particularly those with over 10,000 cells (Fig \ref{supp.fig.10}c). In particular, for datasets of approximately 10,000 cells, NESS(UMAP) completes in less than 17 minutes, which is the same amount of time MuTrans requires to analyze a dataset of only about 4,000 cells. Thess results demonstrate the scalability of NESS to potential large single-cell datasets.

\subsection{Implementation Details}\label{implement.sec}


\paragraph{GCP definition and implementation details of NESS.} For all the analyses presented in this study, we utilized the NESS approach, implemented in R, which supports t-SNE, UMAP, and PHATE for embedding high-dimensional data. The algorithm quantifies the stability of low-dimensional representations using k-nearest neighbor (k-NN) graphs. For t-SNE, the parameter \texttt{perplexity} is identified as the GCP parameter, and random initialization is used. For UMAP, the algorithm is randomly initialized with \texttt{lvrandom} and the nearest neighbor parameter \texttt{n\_neighbors} is identified as GCP. For PHATE, we apply \texttt{phate} with \texttt{ndim = 2}, identify \texttt{knn} as the GCP, and used a fixed seed for reproducibility. For DensMAP, we identify \texttt{n\_neighbors} as the GCP. All the other parameters are set as default values.

The function takes as input a set of GCP parameters, pre-processed (such as standard quality control and count normalization pipelines for single-cell data) high-dimensional data, an optional cell type annotation list, and the chosen dimensionality reduction method (\texttt{"tsne"}, \texttt{"umap"}, or \texttt{"phateR"}). 
By default and in all our analyses, we set \texttt{N = 30} independent runs and used \texttt{k = 50} neighbors for k-NN calculations. The stability of embeddings was quantified using a \texttt{knn.score} computed over multiple runs, with a stability threshold of \texttt{0.75} by default.

\paragraph{Data preprocessing.} For each dataset analyzed in this study, we first applied the singular value hard-thresholding algorithm (Algorithm 2 of the Supplementary Notes of Ma et al. \cite{ma2024principled}) to the normalized and scaled count matrices to denoise the data as a preprocessing step. The number of singular values is determined by $r_{max} = \max\{r : 1 \le r \le n, \lambda_r/\lambda_{r+1}>1+c\}$
for some small constant $c > 0$ such as 0.01. Theoretical justifications of such a method have been established based on standard concentration argument \cite{yao2015sample}.
For the Dentate gyrus dataset, we excluded the rare cell types/states with less than 50 cells. For the Spermatogenesis dataset, we excluded the cell types/states with less than 5 cells, and renamed ``A3-A4-In-B Differentiating spermatogonia" as ``spermatogonia", renamed ``Leptotene/Zygotene spermatocytes" as ``early spermatocytes", combined
``DIplotene/Secondary spermatocytes" and ``Pachytene spermatocytes" into ``late spermatocytes", and combined ``Early Round spermatids", ``Mid Round spermatids" and ``Late Round spermatids" into ``spermatid". The Murine Intestinal data, Dentate gyrus data, and Spermatogenesis data were pre-processed in previous studies, giving the 2,000 most variable genes.

\paragraph{Implementation of baseline methods.} Below we describe our implementation of other existing methods.
\begin{itemize}  
    \item \textbf{scDEED}: The R function \texttt{scDEED} from the scDEED package with \texttt{K = 50}, \texttt{reduction.method = "tsne"}, and \texttt{perplexity} set to the corresponding GCP value. Other parameters were set to default values.
    
    \item \textbf{pyMuTrans}: The Python function \texttt{pm.dynamical\_analysis} from the pyMuTrans library with \\
    \texttt{choice\_distance = "cosine"}, \texttt{perplex} set to the corresponding GCP value, \texttt{K\_cluster = 9.0}, \texttt{trials = 10}, and \texttt{reduction\_coord = "tsne"}. Other parameters were set to default values.

    \item \textbf{EMBEDR}: The Python function \texttt{calculate\_EES} from the embedr library, using \texttt{NearestNeighbors} (from \texttt{sklearn.neighbors}) with \texttt{n\_neighbors = 50} to construct a Gaussian kernel-weighted affinity matrix for computing embedding evaluation scores (EES). Other parameters were set to default values.

    \item \textbf{dynamicviz}: The Python function \texttt{boot.generate} from the dynamicviz library with \texttt{method = "tsne"}, \texttt{B = 4}, and \texttt{random\_seed = 42} for bootstrap sampling. Stability scores were derived using \texttt{score.stability\_from\_variance} with \texttt{alpha = 20}. Other parameters were set to default values.

     \item \textbf{scVelo}: The scVelo package was used to preprocess and analyze the RNA velocity dataset. Preprocessing was performed with default parameters. Unspliced and spliced gene expression matrices were smoothed using \texttt{scv.pp.moments} with \texttt{n\_pcs=30} and \texttt{n\_neighbors=30}. RNA velocity was subsequently inferred using \texttt{scv.tl.velocity} and default parameters.
     
\end{itemize}

 \subsection{Data Availability}

   The iPSC data can be downloaded from \url{https://github.com/cliffzhou92/MuTrans-release/blob/main/Data/ipsc.h5ad}. The pre-processed Murine Intestinal scEU-seq data can be downloaded from figshare \url{https://doi.org/10.6084/m9.figshare.23737116.v1}. The Embryoid Body data can be downloaded from figshare \url{https://doi.org/10.6084/m9.figshare.23737416.v1}. The Mouse Hema data can be downloaded from the PhateR GitHub repository \url{https://github.com/KrishnaswamyLab/PHATE/blob/main/data/BMMC_myeloid.csv.gz}. The pre-processed Spermatogenesis data can be accessed from the R package scRNAseq under the data name ``HermannSpermatogensisData." The pre-processed Dentate gyrus dataset can be accessed from the Python package scVelo under the data name ``scvelo.datasets.dentategyrus." 
 
 \subsection{Code Availability}

  The NESS R package and reproducible analysis scripts can be retrieved and downloaded from our online GitHub repository:  
\url{https://github.com/Cathylixi/NESS}.

\section*{Acknowledgments}

We would like to thank the three anonymous reviewers for their helpful and constructive comments and suggestions. RM would like to thank Jonas Fischer, Allon Klein, Dmitry Kobak, Zhexuan Liu, Tiffany Tang, Olivier Pourqui\'e, and Yiqiao Zhong for helpful discussions. RM and BY would like to thank Stefan Steinerberger for inspiring discussions on the theoretical analysis of t-SNE.

\bibliographystyle{abbrv}
\bibliography{references}

    \newpage
	\appendix

        \setcounter{table}{0}
        \renewcommand{\thetable}{S\arabic{table}}%
        \setcounter{figure}{0}
        \renewcommand{\thefigure}{S\arabic{figure}}%

\begin{centering}
	
	{\Huge Supplemental Notes\par}

	\end{centering}
	
	\setcounter{page}{1}
	\vspace{1cm}
	
Our supplemental notes contains additional tables and figures from our experiments and data analyses, as well as the proof of Theorem 1 in our main text.

\appendix

    \section{Supplementary Tables and Figures}

\begin{table}[h!]
\centering
\renewcommand{\arraystretch}{1.3} 
\resizebox{\textwidth}{!}{ 
\begin{tabular}{l|l|l|l|l|l}
\hline
\textbf{Data Name}                     & \textbf{Cell Number} & \textbf{Feature Number} & \textbf{Biological Process} & \textbf{Technology} & \textbf{Citation} \\ \hline
\textbf{Mouse Hema}                    & 2730                 & 3451                     & Blood cell differentiation   & scRNA-Seq           & \cite{paul2015transcriptional} \\ \hline
\textbf{iPSC}                                   & 1081                 & 96                       & iPSC differentiation         & scRT-qPCR           & \cite{bargaje2017cell} \\ \hline
\textbf{Murine Intestinal}       & 3452                 & 2000                     & Murine intestinal organoids development & scEU-Seq & \cite{battich2020sequencing} \\ \hline
\textbf{Embryoid Body}                  & 31029                & 19122                    & Embryonic stem cell differentiation & scRNA-Seq & \cite{moon2019visualizing} \\ \hline
\textbf{Dentate Gyrus}                  & 2930                 & 2000                     & Dentate gyrus neurogenesis  & scRNA-Seq           & \cite{hochgerner2018conserved} \\ \hline
\textbf{Spermatogenesis}            & 2325                 & 2000                     & Spermatogenesis             & scRNA-Seq           & \cite{hermann2018mammalian} \\ \hline
\textbf{CD8+ T cells}            & 500                 & 2000                     &  Human CD8+ T cells           & scRNA-Seq           & \cite{10x2019new} \\  \hline
\textbf{bmcite}            & 30672               & 2000                     &   Human bone marrow mononuclear cells     & scCITE-seq           & \cite{stuart2019comprehensive} \\ \hline
\end{tabular}
}
\caption{Summary of the analyzed single-cell datasets.}
\label{table.1}
\end{table}

\begin{table}[h!]
\footnotesize
    \centering
    \begin{tabular}{lccccl} 
        \toprule
        \textbf{Gene} & \textbf{Direction} & \textbf{Spearman Correlation} & \textbf{P-value} & \textbf{Adjusted P-value} \\
        \midrule
        ACVR2A  & Positive  & 0.0811  & $7.63 \times 10^{-3}$  & $1.79 \times 10^{-2}$  \\
        ALCAM   & Negative  & -0.1217  & $6.05 \times 10^{-5}$  & $3.23 \times 10^{-4}$  \\
        ANF     & Negative  & -0.0733  & $1.59 \times 10^{-2}$  & $3.32 \times 10^{-2}$  \\
        BAMBI   & Positive  & 0.1786  & $3.33 \times 10^{-9}$  & $1.60 \times 10^{-7}$  \\
        BMP2    & Positive  & 0.0793  & $9.12 \times 10^{-3}$  & $2.04 \times 10^{-2}$  \\
        DLL3    & Positive  & 0.0967  & $1.46 \times 10^{-3}$  & $4.16 \times 10^{-3}$  \\
        EMILIN2 & Positive  & 0.1312  & $1.50 \times 10^{-5}$  & $1.20 \times 10^{-4}$  \\
        EOMES   & Positive  & 0.0931  & $2.19 \times 10^{-3}$  & $5.84 \times 10^{-3}$  \\
        EPCAM   & Negative  & -0.1041  & $6.08 \times 10^{-4}$  & $2.34 \times 10^{-3}$  \\
        EVX1    & Positive  & 0.0906  & $2.86 \times 10^{-3}$  & $7.23 \times 10^{-3}$  \\
        FGF12   & Positive  & 0.0939  & $1.99 \times 10^{-3}$  & $5.47 \times 10^{-3}$  \\
        FGF8    & Positive  & 0.1989  & $4.14 \times 10^{-11}$  & $3.97 \times 10^{-9}$  \\
        FGFR1   & Negative  & -0.1731  & $1.02 \times 10^{-8}$  & $3.27 \times 10^{-7}$  \\
        FGFR2   & Negative  & -0.1106  & $2.70 \times 10^{-4}$  & $1.12 \times 10^{-3}$  \\
        FZD2    & Positive  & 0.1189  & $8.82 \times 10^{-5}$  & $4.45 \times 10^{-4}$  \\
        FZD4    & Positive  & 0.1178  & $1.03 \times 10^{-4}$  & $4.96 \times 10^{-4}$  \\
        GATA4   & Positive  & 0.1274  & $2.65 \times 10^{-5}$  & $1.59 \times 10^{-4}$  \\
        GATA6   & Positive  & 0.0892  & $3.34 \times 10^{-3}$  & $8.23 \times 10^{-3}$  \\
        HAND1   & Positive  & 0.1691  & $2.23 \times 10^{-8}$  & $5.36 \times 10^{-7}$  \\
        HAND2   & Positive  & 0.0845  & $5.43 \times 10^{-3}$  & $1.30 \times 10^{-2}$  \\
        HEY1    & Positive  & 0.1632  & $6.85 \times 10^{-8}$  & $1.31 \times 10^{-6}$  \\
        HHIP    & Negative  & -0.1031  & $6.89 \times 10^{-4}$  & $2.55 \times 10^{-3}$  \\
        INHBA   & Negative  & -0.1249  & $3.84 \times 10^{-5}$  & $2.17 \times 10^{-4}$  \\
        KDR     & Positive  & 0.1287  & $2.18 \times 10^{-5}$  & $1.40 \times 10^{-4}$  \\
        KIT     & Negative  & -0.0788  & $9.53 \times 10^{-3}$  & $2.08 \times 10^{-2}$  \\
        LEFTY1  & Positive  & 0.1614  & $9.43 \times 10^{-8}$  & $1.50 \times 10^{-6}$  \\
        MESP1   & Positive  & 0.0907  & $2.83 \times 10^{-3}$  & $7.23 \times 10^{-3}$  \\
        MESP2   & Positive  & 0.1483  & $9.77 \times 10^{-7}$  & $1.17 \times 10^{-5}$  \\
        MIXL1   & Positive  & 0.1305  & $1.69 \times 10^{-5}$  & $1.25 \times 10^{-4}$  \\
        MSX1    & Positive  & 0.0757  & $1.28 \times 10^{-2}$  & $2.73 \times 10^{-2}$  \\
        MSX2    & Positive  & 0.1430  & $2.35 \times 10^{-6}$  & $2.51 \times 10^{-5}$  \\
        MYL4    & Positive  & 0.1115  & $2.40 \times 10^{-4}$  & $1.05 \times 10^{-3}$  \\
        MYOCD   & Positive  & 0.0986  & $1.18 \times 10^{-3}$  & $3.53 \times 10^{-3}$  \\
        PDGFB   & Positive  & 0.1006  & $9.26 \times 10^{-4}$  & $3.17 \times 10^{-3}$  \\
        PDGFRA  & Positive  & 0.1287  & $2.18 \times 10^{-5}$  & $1.40 \times 10^{-4}$  \\
        PDGFRB  & Positive  & 0.1012  & $8.64 \times 10^{-4}$  & $3.07 \times 10^{-3}$  \\
        PTCH1   & Negative  & -0.0996  & $1.04 \times 10^{-3}$  & $3.45 \times 10^{-3}$  \\
        PTX1    & Negative  & -0.1158  & $1.36 \times 10^{-4}$  & $6.20 \times 10^{-4}$  \\
        SIRPA   & Positive  & 0.1055  & $5.09 \times 10^{-4}$  & $2.04 \times 10^{-3}$  \\
        TBX2    & Positive  & 0.0693  & $2.28 \times 10^{-2}$  & $4.65 \times 10^{-2}$  \\
        TBX20   & Positive  & 0.0990  & $1.11 \times 10^{-3}$  & $3.48 \times 10^{-3}$  \\
        TGFB2   & Positive  & 0.0794  & $9.01 \times 10^{-3}$  & $2.04 \times 10^{-2}$  \\
        TGFB1   & Positive  & 0.1497  & $7.67 \times 10^{-7}$  & $1.05 \times 10^{-5}$  \\
        TGFBR1  & Positive  & 0.1391  & $4.43 \times 10^{-6}$  & $3.87 \times 10^{-5}$  \\
        TGFBR2  & Positive  & 0.0989  & $1.12 \times 10^{-3}$  & $3.48 \times 10^{-3}$  \\
        VEGFA   & Negative  & -0.0689  & $2.36 \times 10^{-2}$  & $4.72 \times 10^{-2}$  \\
        WNT5A   & Positive  & 0.1403  & $3.63 \times 10^{-6}$  & $3.49 \times 10^{-5}$  \\
        WNT5B   & Positive  & 0.0966  & $1.47 \times 10^{-3}$  & $4.16 \times 10^{-3}$  \\
        \bottomrule
    \end{tabular}
    
    \caption{\small List of key genes in iPSCs with significant correlations (BH-adjusted $p$-value $<$ 0.05) between gene expression levels and NESS local stability from t-SNE embedding (GCP = 150, recommended by NESS). The table includes the correlation direction, Spearman correlation coefficients, $p$-values, and BH-adjusted $p$-values.} \label{supp.table.2}
\end{table}

\newpage



    \begin{figure}[h!] 
    \centering
    \includegraphics[width=1\textwidth]{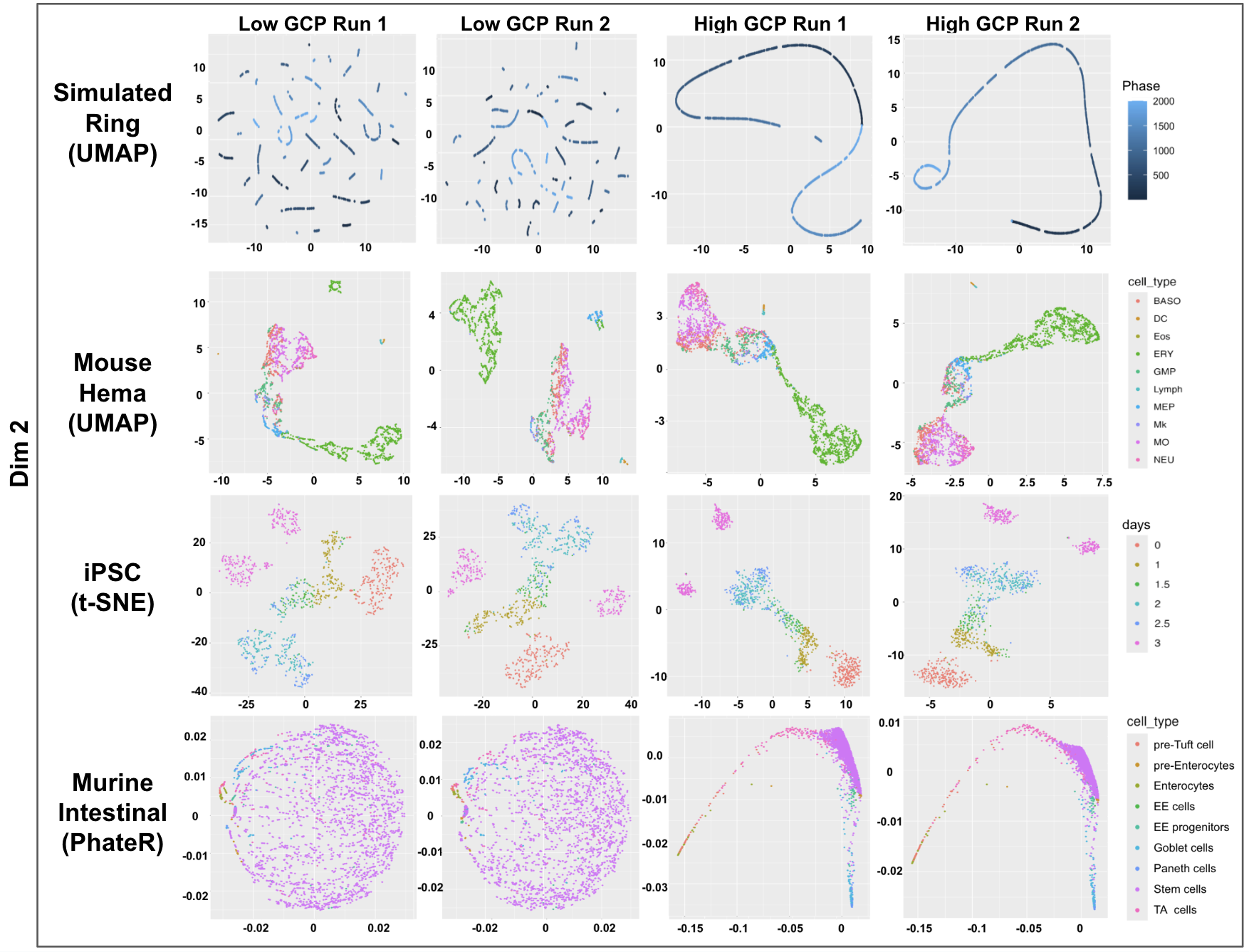} 
    \caption{Additional examples of NE embeddings for simulated (first row) and benchmark single-cell datasets (last three rows) generated by popular NE algorithms, with cells colored by their labels. All embeddings are generated using random initialization. The left two columns correspond to two instances of RIs under relative low GCPs, whereas the right two columns show embeddings under relatively high GCP values. The results suggest low GCP can introduce artifacts.}
    \label{sup.fig.2}
\end{figure}

\begin{figure}[h] 
    \centering
    \includegraphics[width=0.8\textwidth]{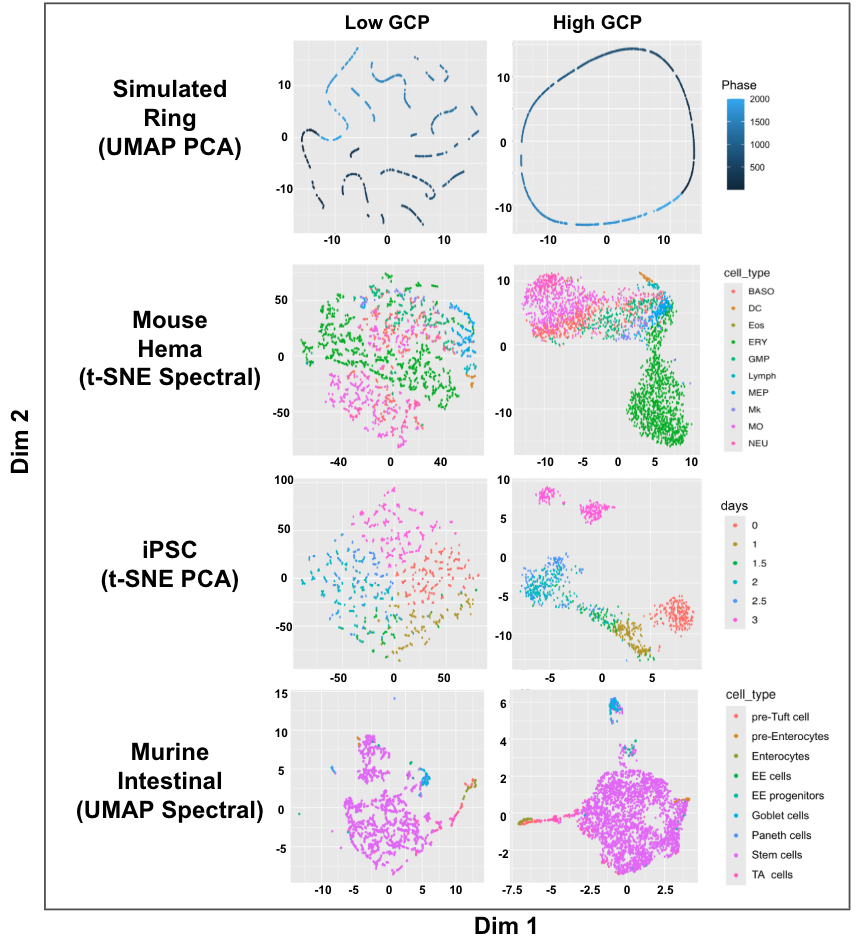} 
    \caption{Additional examples of NE embeddings for simulated (first row) and benchmark single-cell datasets (last three rows) generated by popular NE algorithms, with cells colored by their labels. Two embeddings were generated using PCA initialization (first and third rows), and two using spectral (i.e., Laplacian eigenmap) initialization (second and fourth rows). The left column displays representative embeddings under low GCP value, while the right column shows results under high GCP value. The results suggest similar artifacts under low GCP present under other initializations.}
    \label{supp.fig.11}
\end{figure}

\begin{figure}[h] 
    \centering
    \includegraphics[width=1\textwidth]{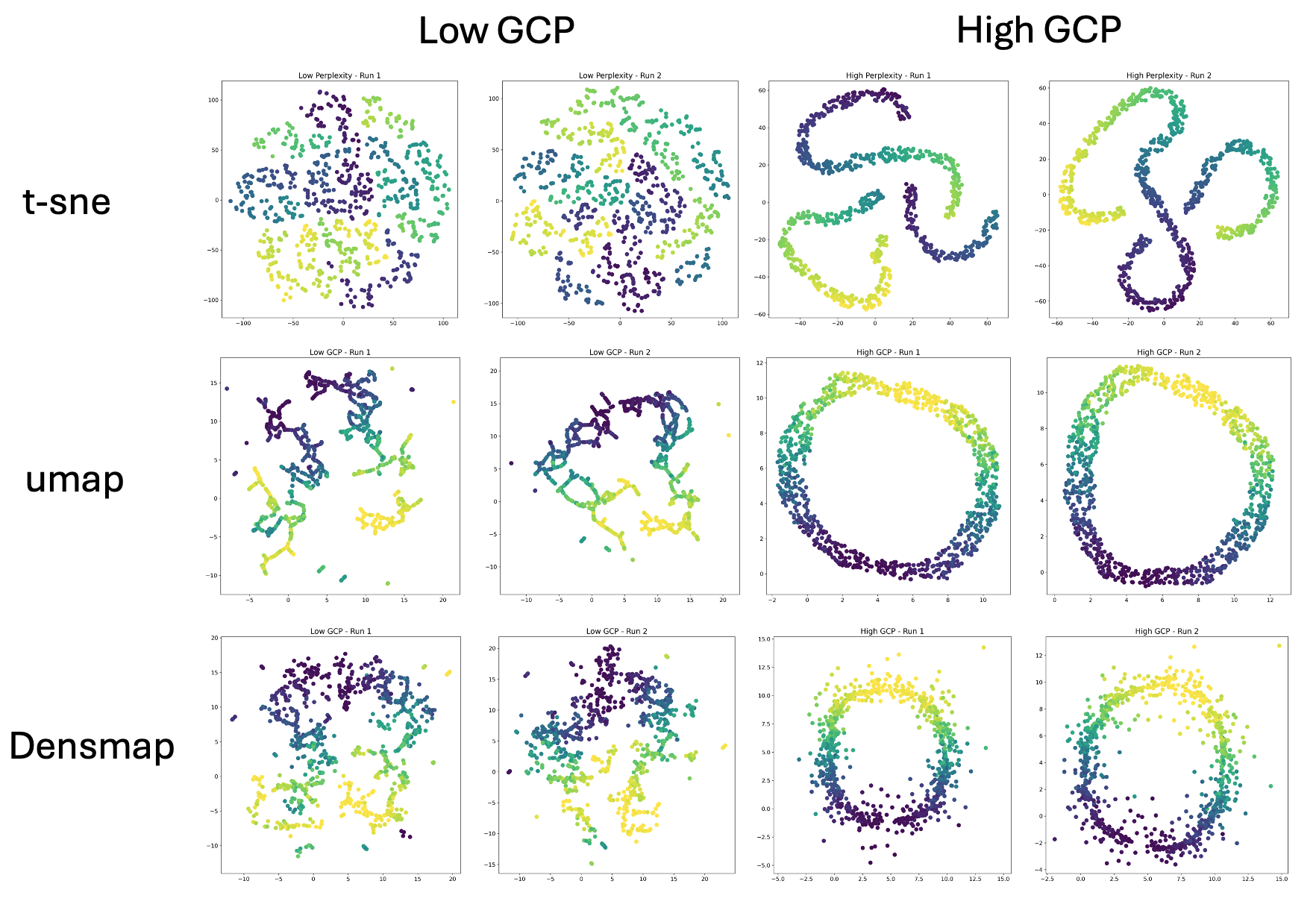} 
    \caption{Additional examples of NE embeddings for simulated data (a ring) generated by popular NE algorithms: t-SNE, UMAP, and DensMAP. Data points are colored by their simulated temporal order. Each method was run twice using different random initializations. The left two columns show embeddings generated with low GCP values, while the right two columns show those with high GCP values. The results suggest lower GCP can introduce greater artifacts.}
    \label{supp.fig.12}
\end{figure}

\begin{figure}[h] 
    \centering
    \includegraphics[width=1\textwidth]{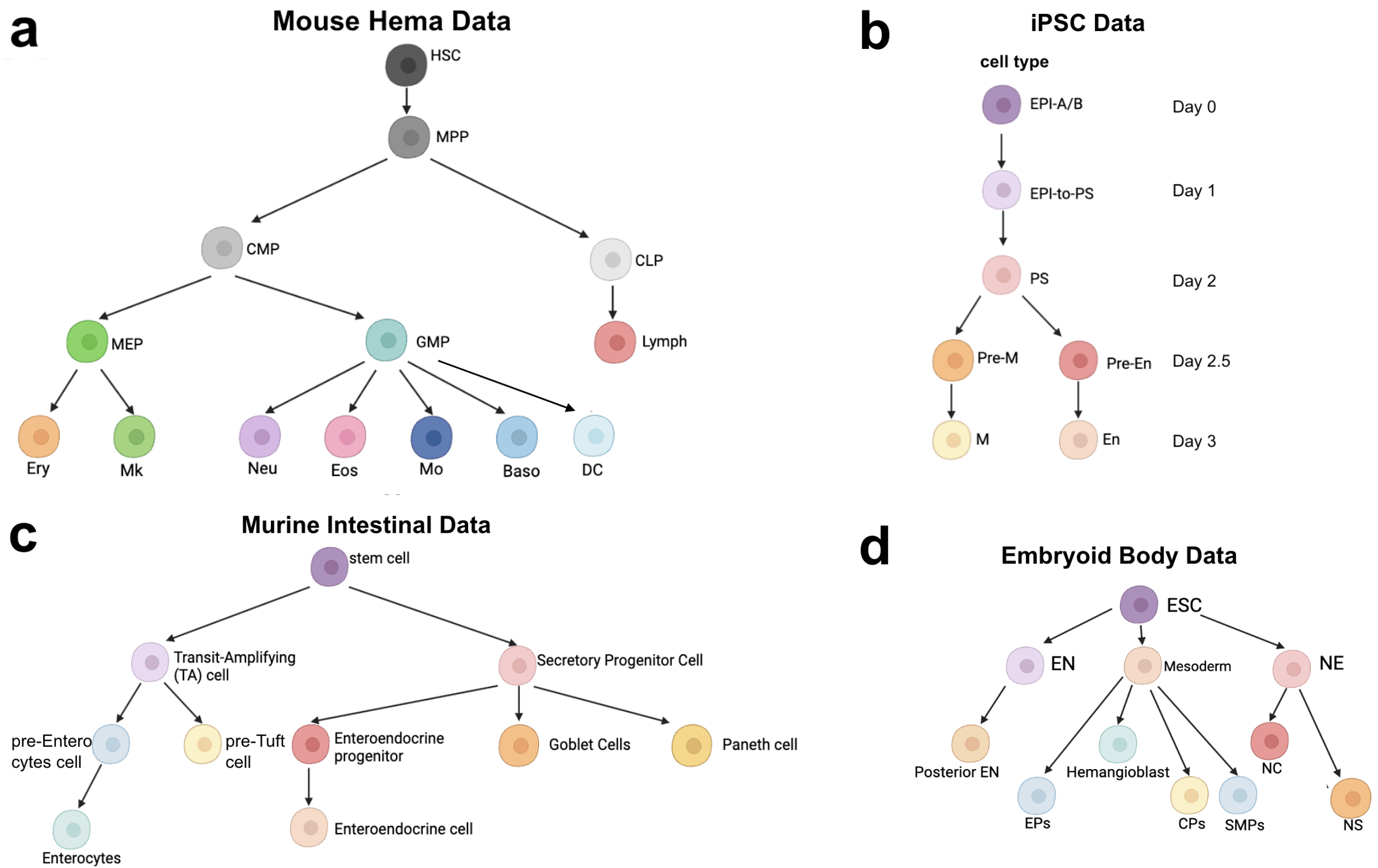} 
    \caption{Cell differentiation hierarchies of the benchmark single-cell datasets. (a) Mouse Hematopoiesis dataset \cite{paul2015transcriptional}. (b) iPSC dataset \cite{bargaje2017cell}. EPI: epiblast; PS: primitive streak; M: mesoderm; En: endoderm. (c) Murine Intestinal dataset \cite{battich2020sequencing}. (d) Embryoid Body dataset \cite{moon2019visualizing}.}
    \label{supp.fig.4}
\end{figure}

\begin{figure}[h] 
    \centering
    \includegraphics[width=1\textwidth]{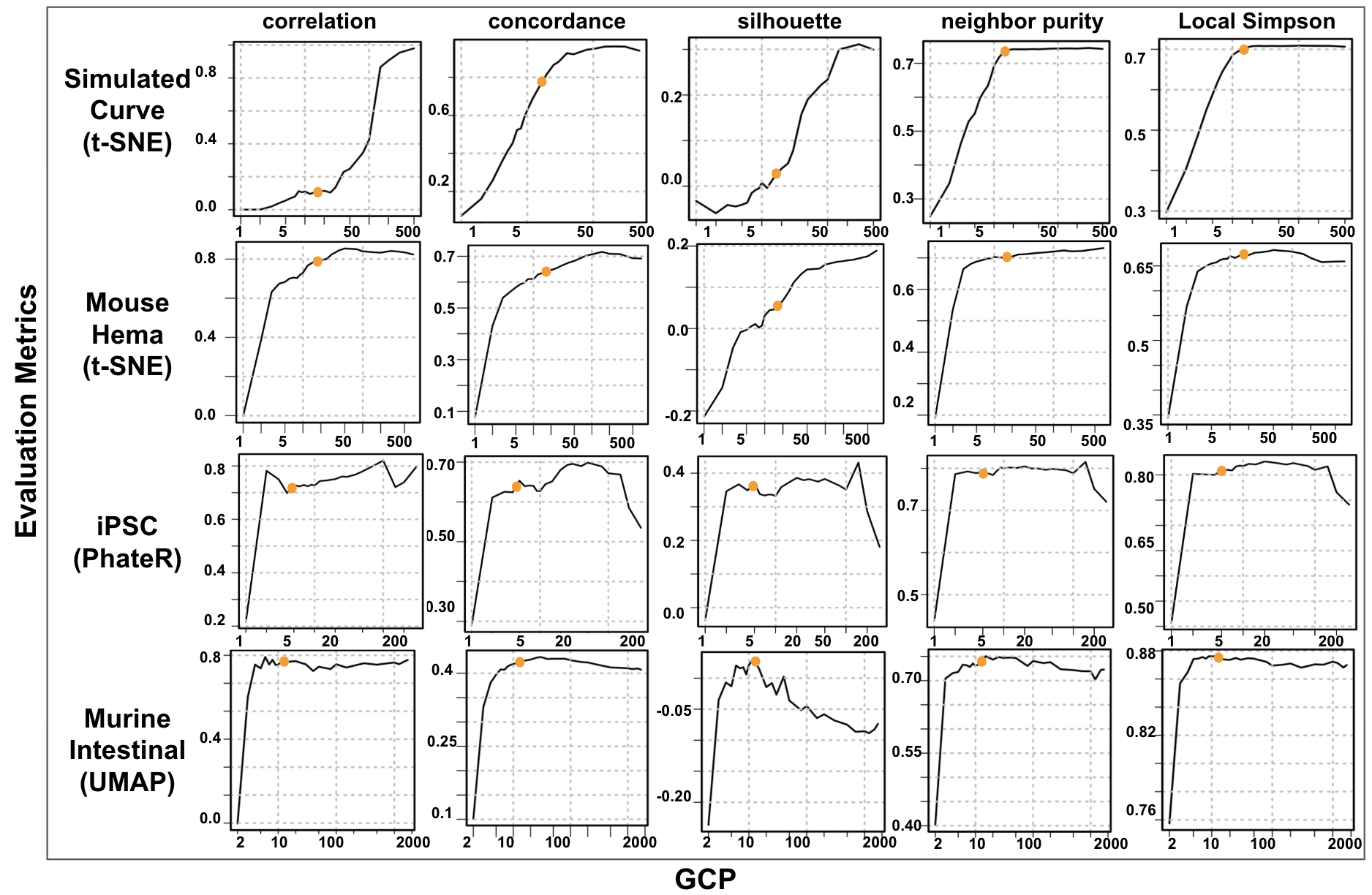} 
    \caption{{Effect of non-random initialization on embedding performance. Results are analogous to Figure 2 but using standard non-random initializations—PCA for t-SNE, Laplacian Eigenmaps for UMAP, and MDS for PhateR. No systematic shift in evaluation metrics is observed relative to random initialization, indicating that the observed patterns are primarily driven by GCP rather than initialization. Yellow dots are default GCP values.}}
    \label{supp.fig.nonrandom}
\end{figure}

\begin{figure}[h] 
    \centering
    \includegraphics[width=1\textwidth]{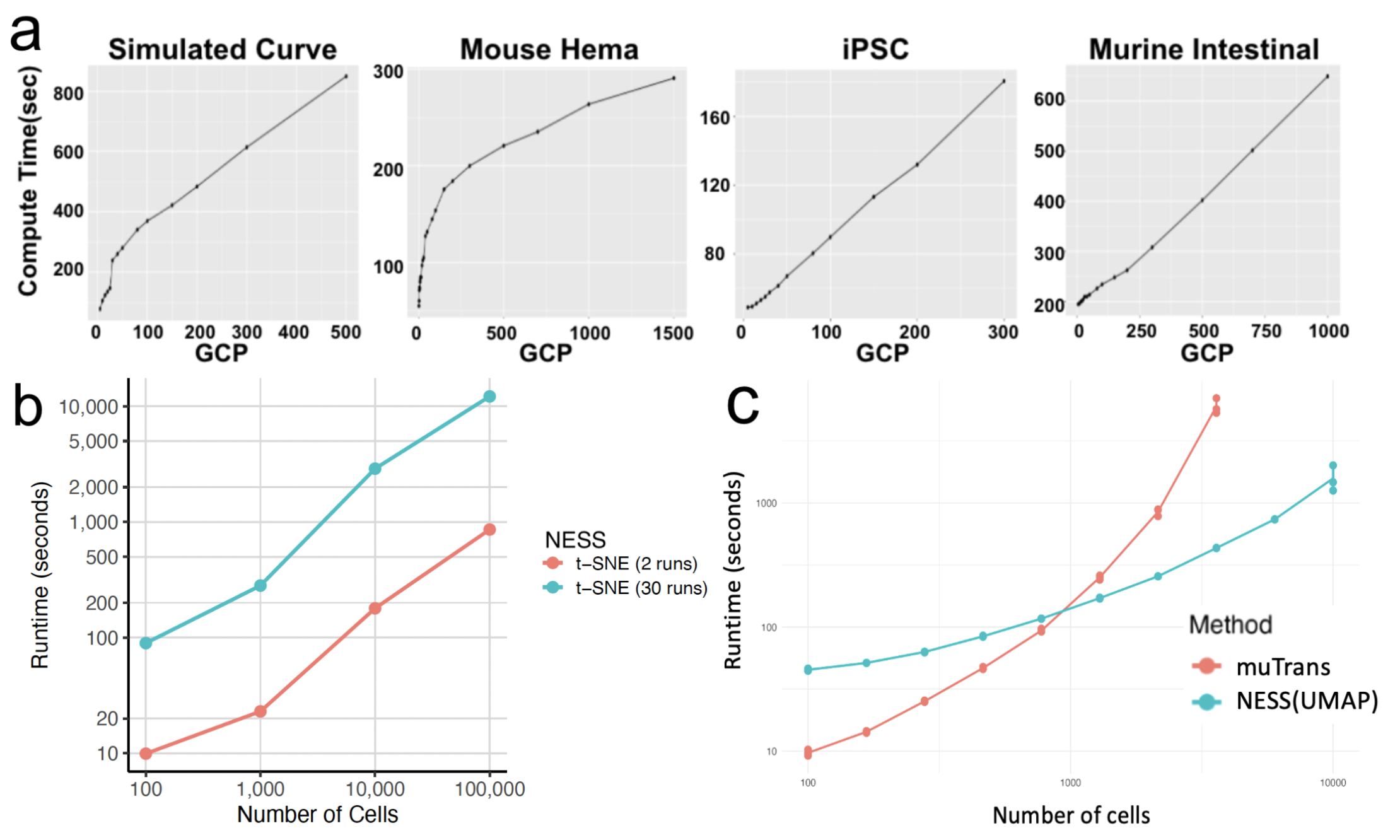} 
    \caption{Running time evaluation. (a) Computation time (in seconds) of the NESS method as a function of the GCP parameter for real and synthetic datasets: Simulated Curve (UMAP), Mouse Hematopoiesis (UMAP), iPSC (t-SNE), and Murine Intestinal (PhateR). A generally linear relationship between GCP and computation time is observed among these examples. All embeddings are initialized with random initialization. In general, a bigger GCP requires more computational time. {(b) Runtime scaling of NESS (t-SNE). Comparison of full NESS (t-SNE) under  30 repetitions per GCP with a reduced version using 2 repetitions per GCP across the same parameter grid. Runtime scales approximately linearly with sample size.} (c) Runtime comparison between NESS(UMAP) with NESS recommended GCP (=30) and the MuTrans method across various sample sizes. Subsamples ranging from 100 to 10,000 data points were randomly sampled from the Mouse Hematopoiesis dataset to evaluate scalability. This comparison suggests better scalability of NESS to large datasets as compared with MuTrans.}
    \label{supp.fig.10}
\end{figure}

\begin{figure}[h] 
    \centering
    \includegraphics[width=1\textwidth]{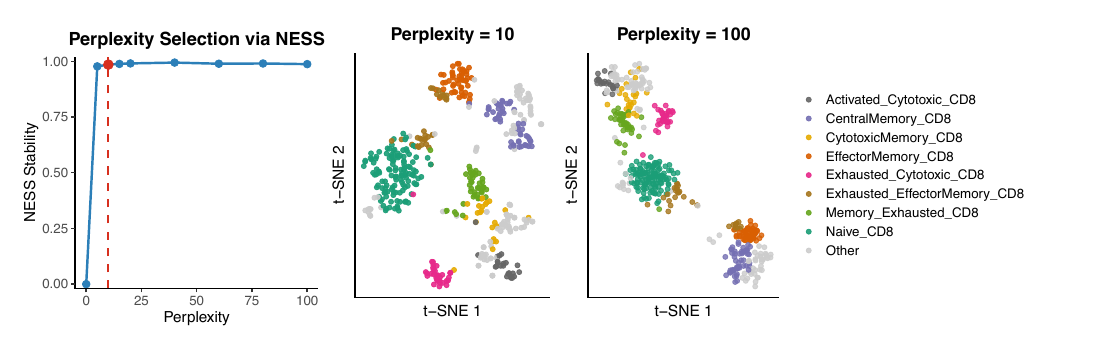} 
    \caption{{Overspecifying perplexity can obscure biologically meaningful heterogeneity. Left: full range NESS global stability line chart (without early stopping) and the NESS recommended GCP (in red). Middle and Right: Comparison of t-SNE embeddings for a CD8+ T-cell scRNA-seq dataset (500 cells) using perplexity = 100 and the NESS-selected perplexity (=10).}}
    \label{supp.fig.over}
\end{figure}

\begin{figure}[h] 
    \centering
    \includegraphics[width=1\textwidth]{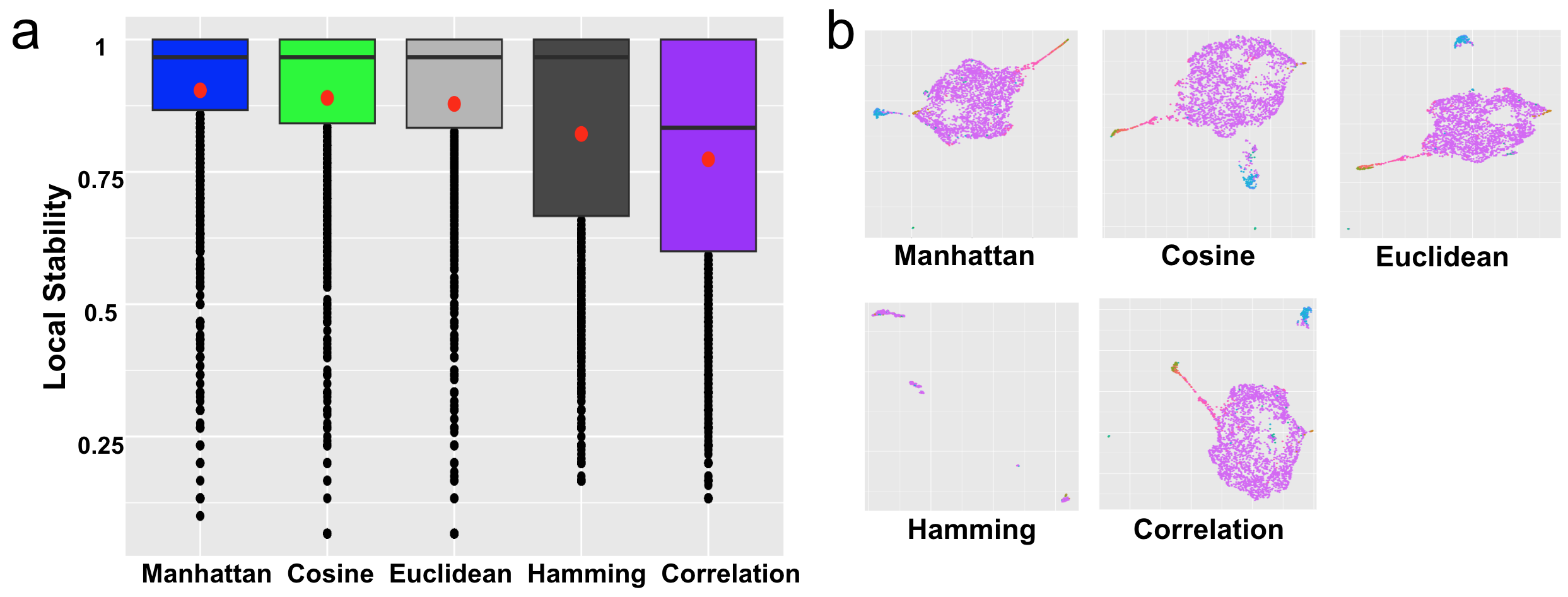} 
    \caption{Use of local stability score for distance selection. (a) Boxplots of NESS local stability scores for UMAP (GCP=30, recommended by NESS) applied to the Murine Intestinal dataset under different distance metrics. The red dots indicate the corresponding global stability scores. (b) UMAP embeddings  (GCP = 30) generated using different distance metrics, colored by cell type. UMAP under Manhattan distance generated the most stable embedding compared with other distance metrics. This result suggests that the NESS global and local stability scores can be used to guide the selection of an appropriate distance metric.}
    \label{supp.fig.5}
\end{figure}

\begin{figure}[h] 
    \centering
    \includegraphics[width=1\textwidth]{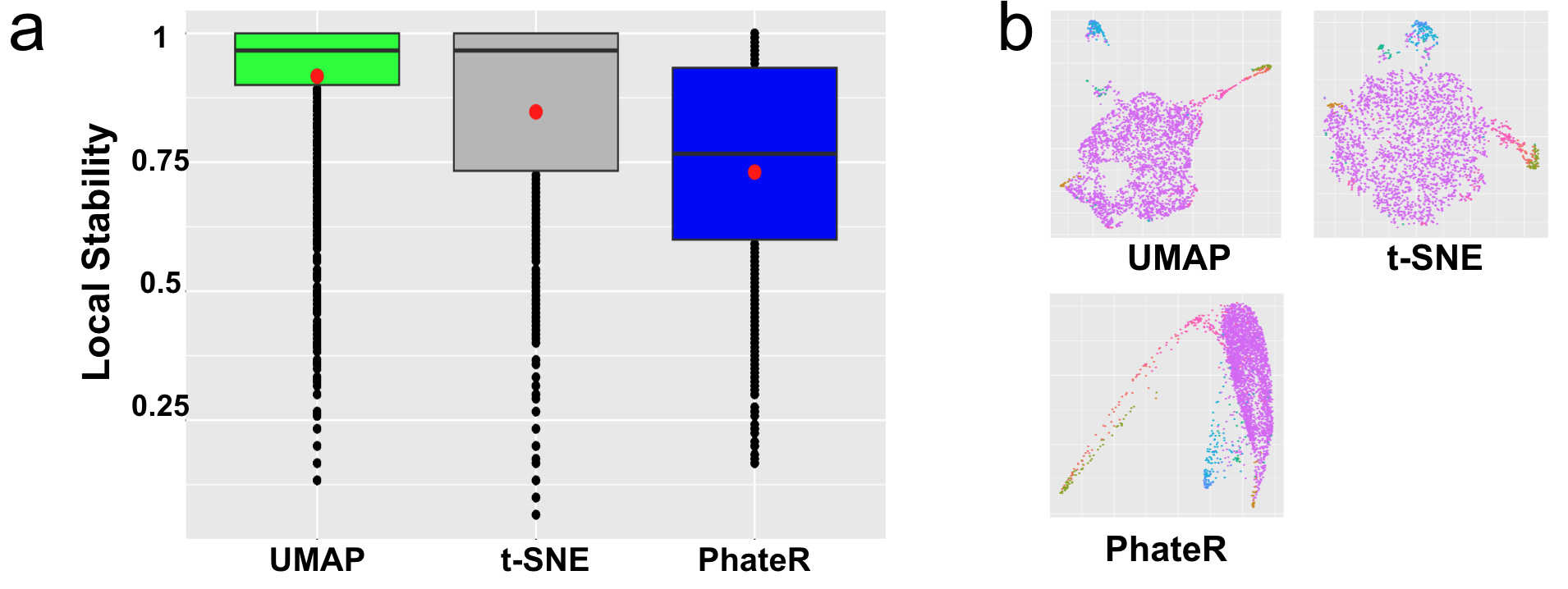} 
    \caption{Use of local stability score for algorithm selection.  (a) Boxplots of NESS local stability scores for different NE algorithms with GCPs recommended by NESS and other hyperparameters under default setting, applied to the Murine Intestinal dataset. The red dots indicate the corresponding global stability scores. (b) Low-dimensional embeddings generated by different NE algorithms with NESS recommended GCPs, with the cells colored by cell types.  UMAP generated the most stable embedding among the three method. This result suggests that the NESS global and local stability scores can be used to guide the selection of an appropriate NE algorithm.}
    \label{supp.fig.6}
\end{figure}

\begin{figure}[h] 
    \centering
    \includegraphics[width=0.4\textwidth]{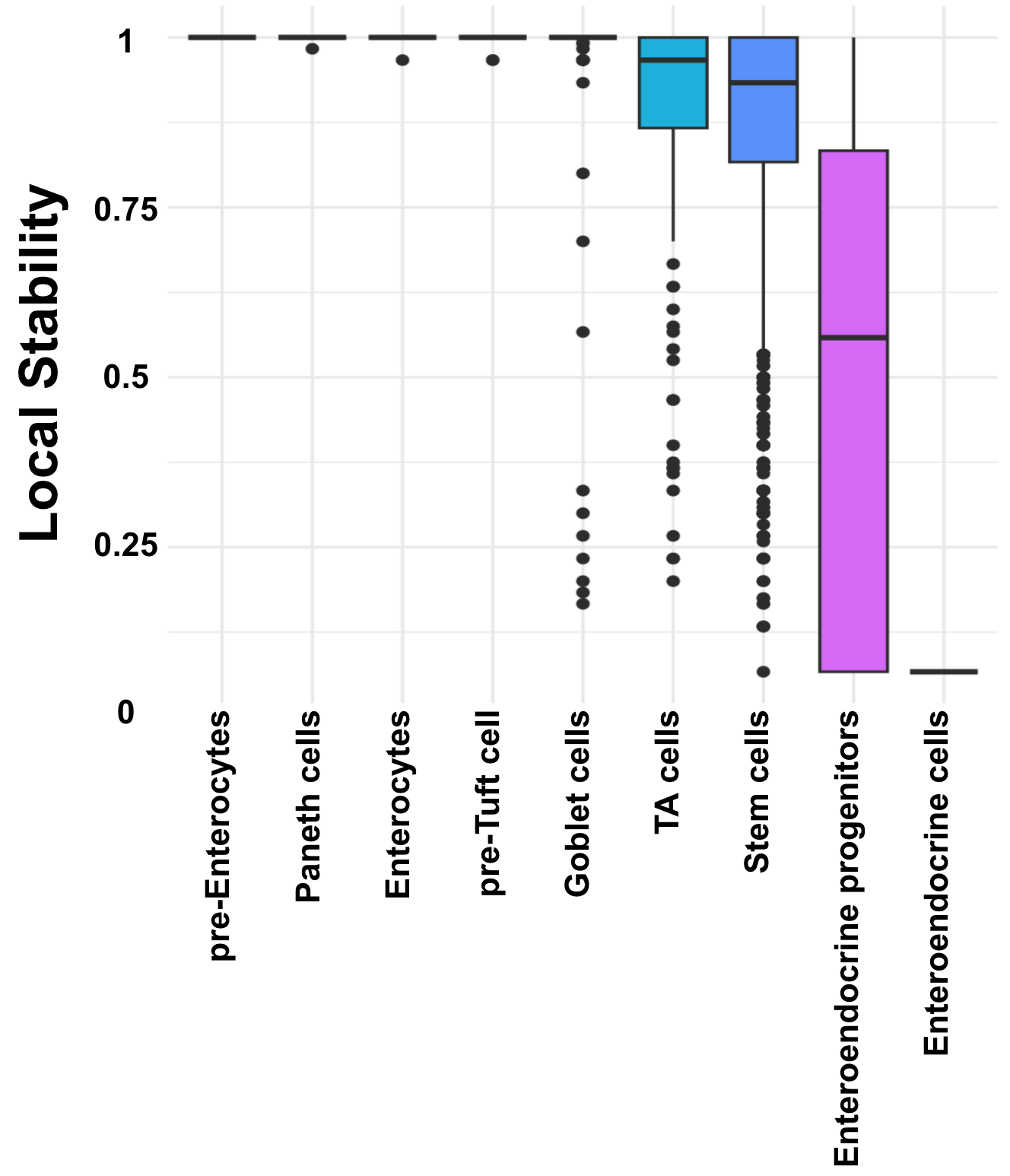} 
    \caption{ Boxplots of NESS(UMAP) local stability scores for the Murine Intestinal dataset,  grouped by different cell types. The stability scores were obtained under  the NESS recommended GCP. This result suggests that the NESS local stability score can vary significantly between cell types.}
    \label{supp.fig.7}
\end{figure}

\begin{figure}[h] 
    \centering
    \includegraphics[width=0.5\textwidth]{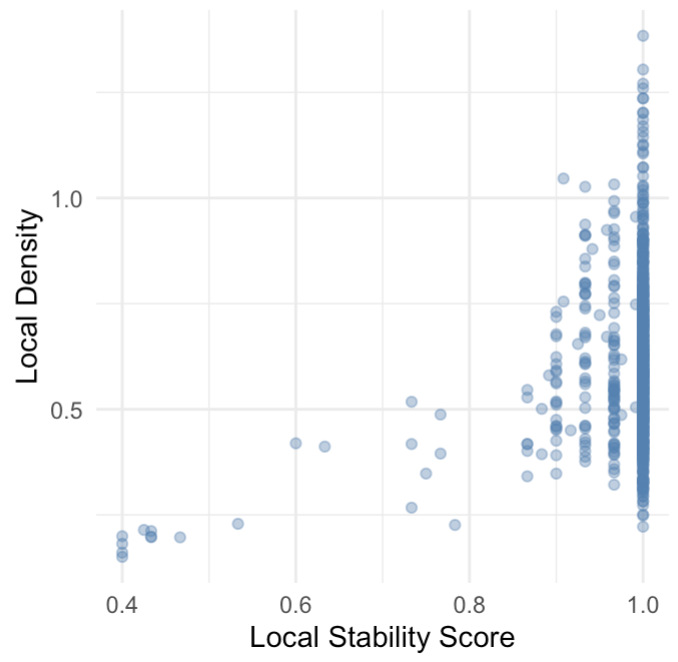} 
    \caption{{Scatter plot between the NESS local stability score and the local density measure on the iPSC data. Local stability shows weak correlation with local density  (Pearson correlation=0.23), indicating that the two quantities are related, but distinct. }}
    \label{supp.scatter}
\end{figure}

\begin{figure}[h] 
    \centering
    \includegraphics[width=0.8\textwidth]{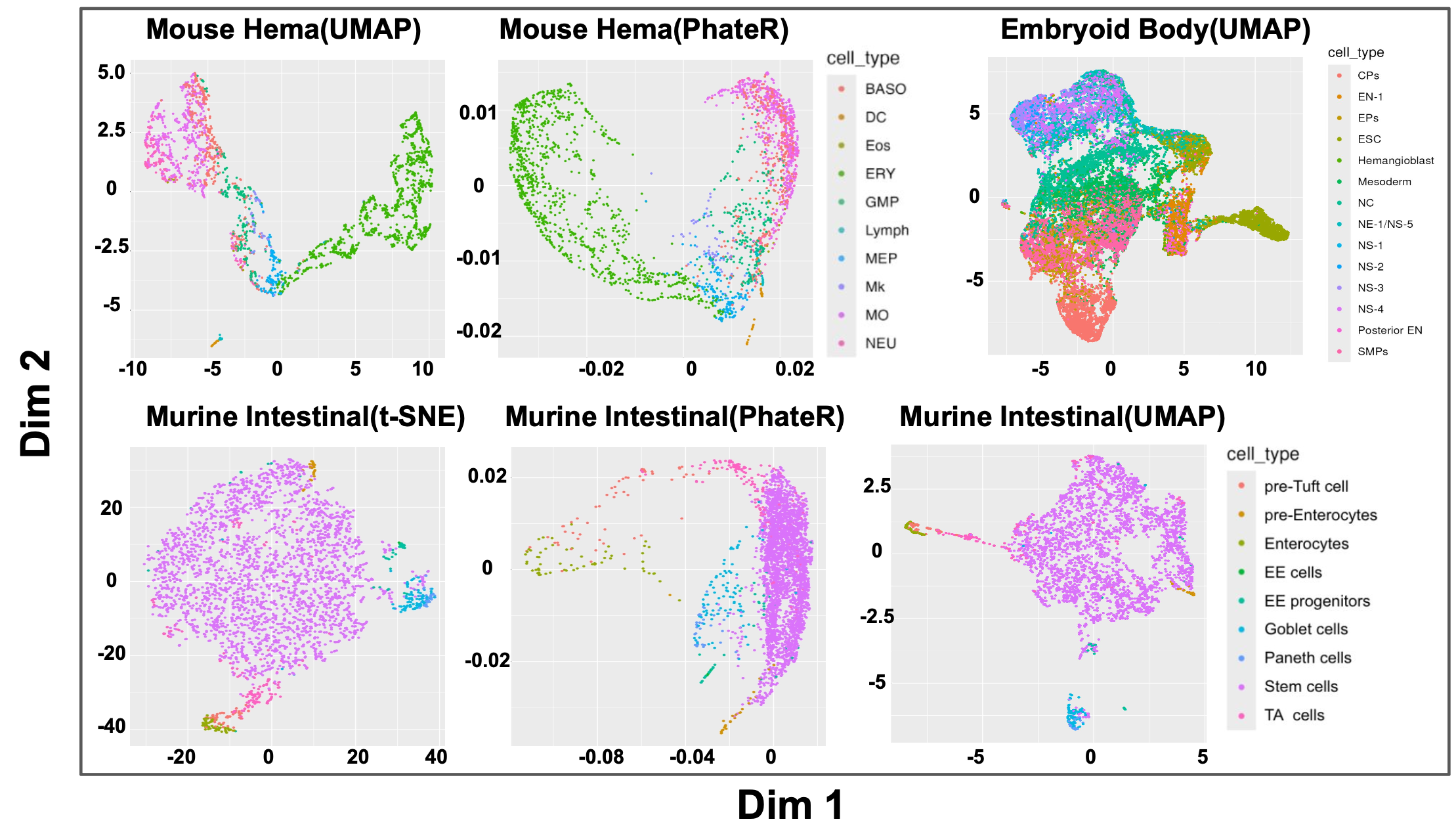} 
    \caption{Additional examples of low-dimensional embeddings of single-cell datasets, produced by  NE  algorithms under default GCPs. For UMAP, the default GCP is \texttt{n\_neighbor}=15; for t-SNE, the default GCP is \texttt{perplexity}=30; and for PhateR, the default GCP is \texttt{knn}=5. In some cases, the default application of NE algorithm introduces artifacts, such as UMAP's fragmentation of ERY population in the Mouse Hema dataset. In other cases, the default application of NE algorithms appear to successfully reveal the underlying lineage structure, as observed in the t-SNE and UMAP embeddings of Murine Intestinal dataset.}
    \label{supp.fig.9}
\end{figure}

\begin{figure}[h] 
    \centering
    \includegraphics[width=1\textwidth]{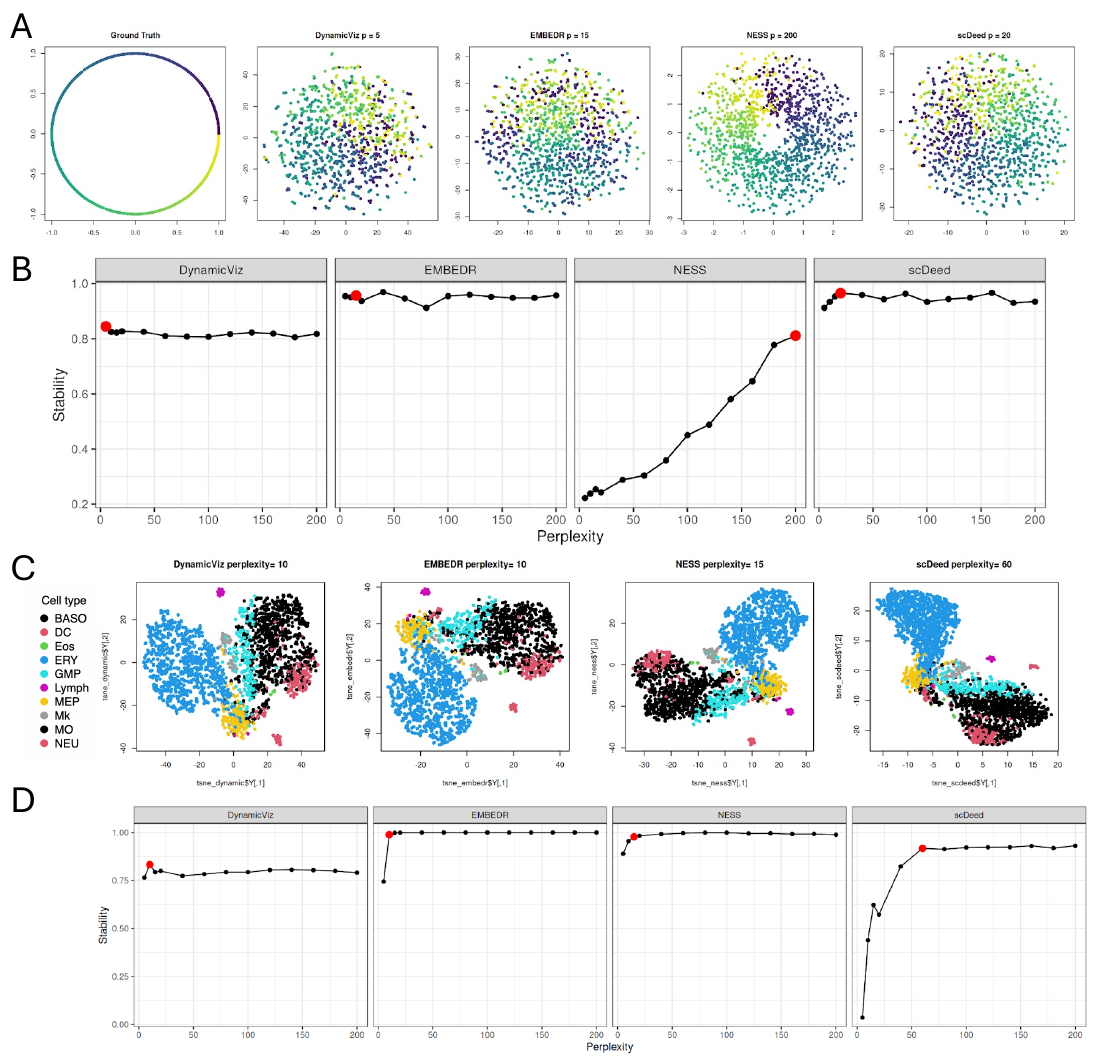} 
    \caption{{Comparison with existing hyperparameter selection methods. (a) ground truth ring structure and t-SNE embeddings of the simulated high-dimensional noisy ring dataset based on different GCP (perplexity) selection methods; (b) GCP versus global stability (or corresponding selection score) across methods on the noisy ring dataset; (c) t-SNE embeddings of the Mouse Hema dataset based on different GCP selection methods; (d) GCP versus global stability (or corresponding selection score) across methods on the Mouse Hema dataset. NESS is benchmarked against EMBEDR, scDEED, and DynamicViz, and performs on par with (Mouse Hema data) or outperforms (noisy ring data) existing methods in selecting appropriate GCP values.}}
    \label{supp.bench}
\end{figure}

\begin{figure}[h] 
    \centering
    \includegraphics[width=1\textwidth]{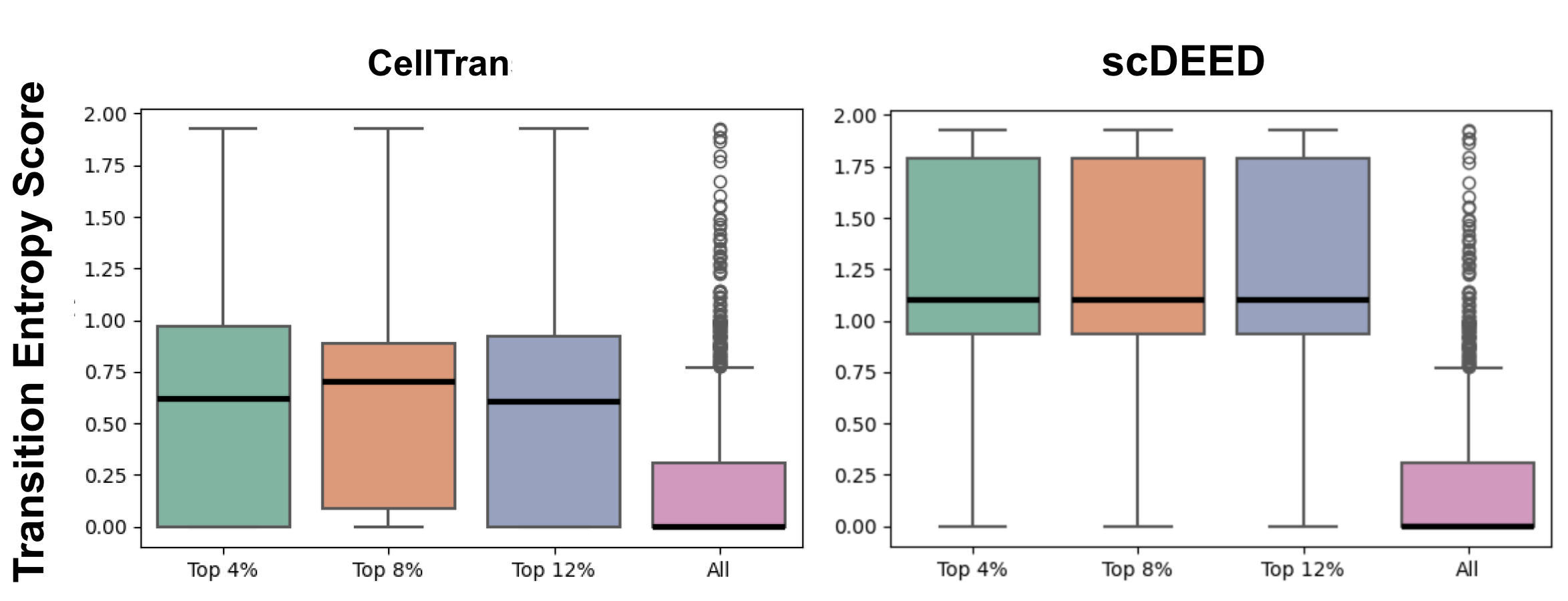} 
    \caption{{Comparison with existing cell-state transition methods. Boxplots of MuTrans transition entropy scores for cells whose pointwise stability measures from CellTran \cite{wang2024transition} (left) and scDEED (right, based on its reported p-values) fall within the bottom $4\%,8\%,12\%$, and $100\%$ percentiles. Higher entropy scores indicate a greater likelihood of being in a transitional cell state. These results suggest that while CellTran and scDEED can largely distinguish transitional from non-transitional cells, they are less sensitive in resolving finer variations in transition entropy among the most unstable cells.}}
    \label{supp.fig.entropy}
\end{figure}

\begin{figure}[h] 
    \centering
    \includegraphics[width=0.6\textwidth]{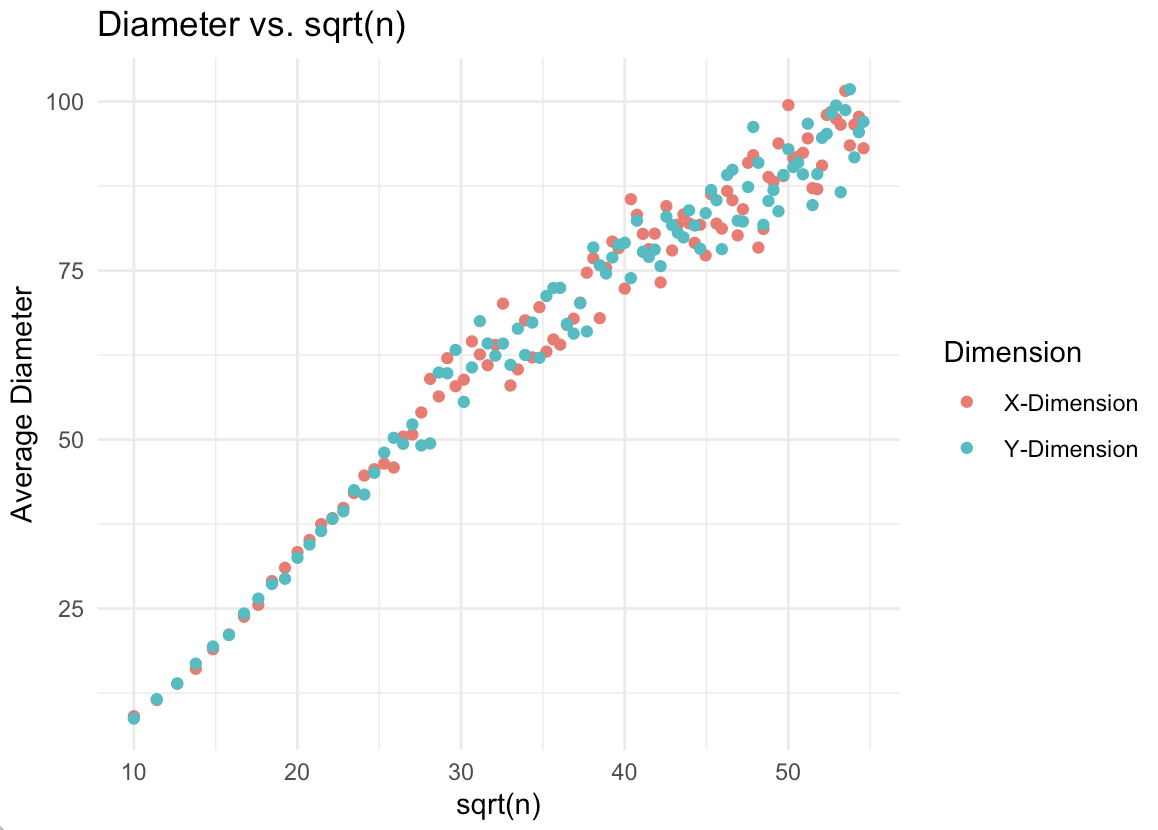} 
    \caption{{Scaling of embedding size with sample size. For t-SNE embeddings of a discrete circle with fixed perplexity (=30) across varying $n$, the average range of the embedding coordinates (in both x- and y-directions) exhibits an approximately linear relationship with $\sqrt{n}$.}}
    \label{supp.fig.diameter}
\end{figure}

\begin{figure}[h] 
    \centering
    \includegraphics[width=0.6\textwidth]{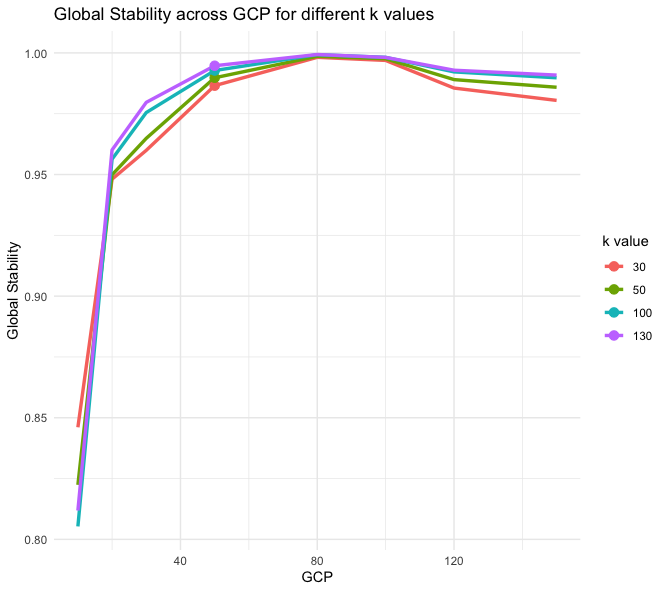} 
    \caption{{Robustness of stability curves to the choice of $k$. On the Mouse Hema dataset, using $k=30,50,100$, or 130 for the local stability score yields similar global stability curves across GCP values. The relative ordering of GCPs and the location of the stability plateau remain largely unchanged, indicating that stability assessment is robust to $k$ and distinct from the role of GCP in embedding construction.}}
    \label{supp.fig.krobust}
\end{figure}

\begin{figure}[h] 
    \centering
    \includegraphics[width=0.9\textwidth]{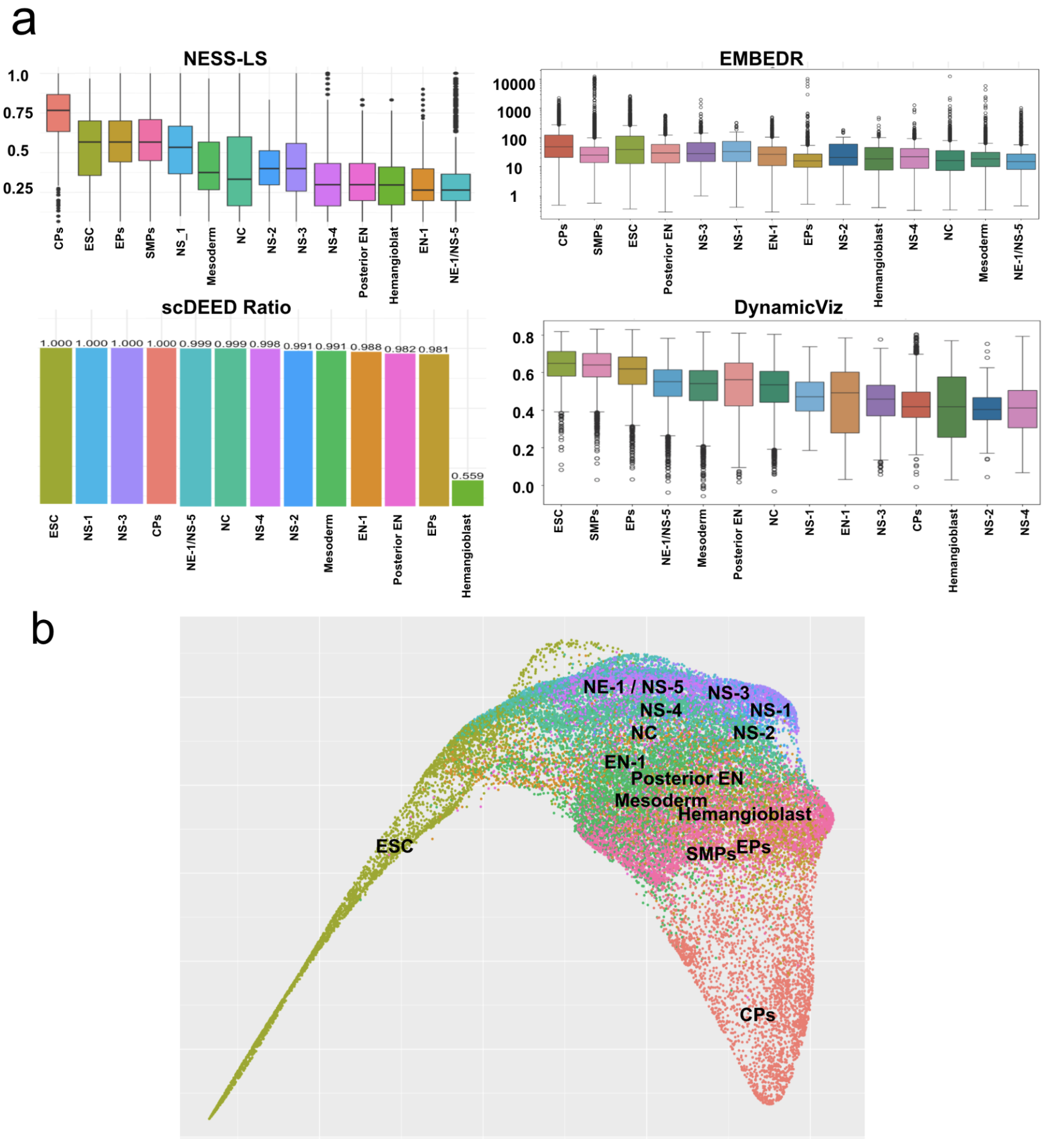} 
    \caption{\small Comparison of stability scores in the Embryoid Body dataset. (a) Comparison of cell-type-specific stability scores in the Embryoid Body dataset, ordered by median values. The top-left panel presents NESS local stability (LS) scores, representing local stability estimates for different cell types. The top-right panel shows EMBEDR stability scores, which quantify embedding reliability. The bottom-left panel displays scDEED stability ratios (computed as Trustworthy / (Trustworthy + Dubious)), where lower values indicate a higher fraction of dubious cells. The bottom-right panel illustrates DynamicViz stability scores, capturing variability in the embedding structure. Abbreviated cell type labels include: embryonic stem cells (ESCs), the primitive streak (PS), mesoderm (ME), endoderm (EN), neuroectoderm (NE), neural crest (NC), neural progenitors (NPs), epicardial precursors (EPs), smooth muscle precursors (SMPs), cardiac precursors (CPs), and neuronal subtypes (NS). The results indicates the advantages of NESS in identifying transitional and stable cell states, compared with alternative methods. (b) NESS(PhateR) visualization of the Embryoid Body dataset, with NESS recommended GCP=100. Specific cell types/states are differently colored and labeled on the low-dimensional embedding for better identification. }
    \label{supp.fig.8}
\end{figure}


\clearpage 

\section{Proof of Theorem \ref{tsne.thm}} \label{sec.proof}

We start by sketching the proof idea.
First, we consider a special mapping $\phi_0: P_n\to \R^2$. We show that for any $S_0>0$, there exists a specific mapping $Y_0=\phi_0(X)$ satisfying (\ref{bilip}) for some $S(\phi_0)=S_0$ and $L(\phi_0)=L_0$. In particular, we obtain a lower bound estimate
\beq\label{low}
L(S_0)\le C(Y_0),
\eeq
for some function $L(\cdot)$ that only depends on $X$. Next, 
let $Y^*=\phi^*(X)$ be the maximizer of $C(Y)$. We show that, as long as $\phi^*$ satisfies (\ref{bilip}) for some $S(\phi^*)\equiv S^*>0$ and $L(\phi^*)\equiv L^*\ge 1$, it holds that
\beq\label{up}
C(Y^*)\le U(S^*,L^*),
\eeq
for some function $U(\cdot,\cdot)$ that only depends on $X$.
Then it follows that, by choosing $S_0=S^*$,  we can construct $\phi_0$ so that, by (\ref{up}) and (\ref{low}), along with the definition of $Y^*$, it holds that
\beq\label{logic.ineq}
L(S^*)=L(S_0)\le C(Y_0)\le C(Y^*)\le  U(S^*,L^*).
\eeq
Analyzing the inequality $L(S^*)\le  U(S^*,L^*)$ implies a lower bound for $L^*$. As a result, we obtain that $L^*\to \infty$ as long as $S^*=S(\phi^*)=n^\tau$ for $0<\tau<1$.

\subsection{Technical preparations and additional notation}

Note that under the affinities (\ref{pij}), the equivalent t-SNE objective (\ref{obj}) can be further reduced to maximizing the following objective
$$ C(y_1,y_2,...,y_n) = \sum_{i \neq j\atop x_i, x_j~\mbox{\tiny neighbors}}  \log(q_{ij}).$$
Hereafter, we also denote $Y=\{y_i\}_{1\le i\le n}$ and $X=\{x_i\}_{1\le i\le n}$.

Next we introduce a few useful lemmas. The first lemma collect some elementary mathematical facts which can be found in standard textbooks on analytical geometry and algebra.

\begin{lem} \label{lem0}
For any two points $a$ and $b$ on the unit circle, if the arc distance between $a$ and $b$ is $\theta$, then the length of the chord between them is $2\sin(\theta/2)$.
\end{lem}

The next few lemmas concerns upper and/or lower bound estimates of the t-SNE related quantities.

\begin{lem} \label{lem1}
For $S\gtrsim n$ and $\{x_i\}_{1\le i\le n}=P_n$, it holds that
\[
\sum_{k\ne \ell}\frac{1}{1+S^2\|x_k-x_\ell\|^2}\asymp\frac{n^3}{S^{2}}.
\]
\end{lem}

\begin{proof}
On the one hand, by Lemma \ref{lem0}, for sufficiently large $n$, we have 
 \[
 S^2\min_{k\ne \ell}\|x_k-x_\ell\|^2\ge S^2\cdot 4\sin^2(\pi/n)\gtrsim n^2\sin^2(\pi/n)\gtrsim 1,
 \]
where in the last inequality we use the fact that 
\[
\frac{x^2}{4}\le \sin^2 x,\qquad \text{for all $0\le x\le \pi/8$}.
\]
As a result, it follows that
\beq
\sum_{k\ne \ell}\frac{1}{1+S^2\|x_k-x_\ell\|^2}\ge \frac{C}{S^2}\sum_{k\ne \ell}\frac{1}{\|x_k-x_\ell\|^2},
\eeq
for some absolute constant $C>0$.
On the other hand, we also have
\beq
\sum_{k\ne \ell}\frac{1}{1+S^2\|x_k-x_\ell\|^2}\le \frac{1}{S^2}\sum_{k\ne \ell}\frac{1}{\|x_k-x_\ell\|^2}.
\eeq
It then holds that
\beq
\sum_{k\ne \ell}\frac{1}{1+S^2\|x_k-x_\ell\|^2}\asymp \frac{1}{S^2}\sum_{k\ne \ell}\frac{1}{\|x_k-x_\ell\|^2}.
\eeq
Without loss of generality, we assume $n$ is an odd integer. Now we calculate that
\begin{align}
&\sum_{k\ne \ell}\frac{1}{\|x_k-x_\ell\|^2}= \sum_{i=1}^{(n-1)/2}\frac{n}{4\sin^2(\pi i/n)}=\sum_{i:\pi i/n\le \pi/8}\frac{n}{4\sin^2(\pi i/n)}+\sum_{i:\pi i/n> \pi/8,\pi i/n<\pi/2}\frac{n}{4\sin^2(\pi i/n)}\nonumber\\
&\le \sum_{i:\pi i/n\le \pi/8}\frac{n}{(\pi i/n)^2}+\sum_{i:\pi i/n> \pi/8, \pi i/n<\pi/2}\frac{n}{4\sin^2(\pi/8)}\lesssim n^3\sum_{i=1}^n\frac{1}{i^2}+\sum_{n/8<i<n/2}n\lesssim n^3,
\end{align}
where the last inequality follows from the elementary inequality
\[
\frac{\pi^2}{6}-\frac{1}{n}\le \sum_{i=1}^n\frac{1}{i^2}\le\frac{\pi^2}{6}-\frac{1}{n+1},
\]
and that
\begin{align}
&\sum_{k\ne \ell}\frac{1}{\|x_k-x_\ell\|^2}= \sum_{i=1}^{(n-1)/2}\frac{n}{4\sin^2(\pi i/n)}=\sum_{i:\pi i/n\le \pi/8}\frac{n}{4\sin^2(\pi i/n)}+\sum_{i:\pi i/n> \pi/8,\pi i/n<\pi/2}\frac{n}{4\sin^2(\pi i/n)}\nonumber\\
&\ge \sum_{i:\pi i/n\le \pi/8}\frac{n}{4(\pi i/n)^2}\gtrsim n^3\sum_{i=1}^n\frac{1}{i^2}\gtrsim n^3,
\end{align}
where the third last inequality follows from 
\[
\sin^2x<x^2,\qquad \text{for all $x>0$}.
\]
It then follows that
\beq \label{sum.eq0}
\sum_{k\ne \ell}\frac{1}{\|x_k-x_\ell\|^2}\asymp n^3.
\eeq
This completes the proof.
\end{proof}

\begin{lem} \label{lem3}
Suppose $x_i$ and $x_j$ are $k-$nearest neighbors on $P_n$. Then for any mapping $\phi$ satisfying (\ref{bilip}) with the scaling factor $S>0$ and the distortion factor $L\ge 1$, for sufficiently large $n$, it holds that
\beq
- \log\left(1+\frac{C_1L^2S^2k^2}{n^2} \right) \le - \log\left(1+\|\phi(x_i) - \phi(x_j)\|^2 \right) \leq - \log\left(1 + \frac{C_2S^2}{n^2}\right),
\eeq
where $C_1,C_2>0$ are some universal constant. In addition, it also holds that
\beq
    \sum_{k \neq \ell}\frac{1}{1+ S^2L^2\|x_k -x_{\ell}\|^2}\le \sum_{k \neq \ell} \frac{1}{1+ \|\phi^*(x_k) - \phi^*(x_{\ell})\|^2} \le     \sum_{k \neq \ell}\frac{1}{1+ S^2\|x_k -x_{\ell}\|^2}.   
\eeq
\end{lem}

\begin{proof}
Since $x_i$ and $x_j$ are $k-$nearest neighbors, we have 
$$ \frac{1}{n}\lesssim \|x_i - x_j\| \lesssim \frac{k}{n}.$$
Therefore,  by (\ref{bilip}), we have
\beq\label{dist.x}
 \frac{S^2}{n^2}\asymp S^2\|x_i-x_j\|^2\le \|\phi(x_i) - \phi(x_j)\|^2 \lesssim S^2L^2\|x_i-x_j\|^2\asymp \frac{L^2S^2k^2}{n^2}.
\eeq
This completes the proof.
\end{proof}

\paragraph{A special mapping $\phi_0$.}
An important ingredient of our proof is a carefully constructed mapping $Y_0=\phi_0(X_0)$, which plays an role in both Part I and Part II of our proof. Below we formally define such a mapping and collect some useful notation and estimates.

Consider the following map $\phi_0:P_n\to \R^2$. Without loss of generality, we assume $n$ is a multiple of an integer $M$, where 
{\beq \label{M.assum}
M\ll n/k.
\eeq}
Define a partition $\{P'_m\}_{1\le m\le M}$ of $P_n$ where $P'_m$ includes $\left( \cos\left( \frac{2\pi i}{n}\right),  \sin \left( \frac{2\pi i}{n}\right) \right)$ for $n/M$ consecutive integers $i$. Suppose on each $P'_m$, $1\le m\le M$, the map $\phi_0:P'_m\to \R^2$ is isometric up to a universal scaling factor $S_0$, that is,
	\beq \label{iso.prop}
	\|\phi_0(x_i)-\phi_0(x_j)\|_2=S_0\|x_i-x_j\|_2, \qquad \text{for any $x_i,x_j\in P_m$.}
	\eeq
	Moreover, we assume that for any $P_k\ne P_j$, there exists an universal constant $c>0$ such that 
	\beq \label{sep.prop}
	cS_0\le \|\phi_0(x_i)-\phi_0(x_j)\|_2\le S_0/c,
	\eeq
	for any $x_i\in P_k$ and $x_j\in P_j$.
	The existence of such a map can be established by the following construction: (i) cut the unit circle evenly into $M$ parts with equal length; (ii) find the middle point of each curve, connect them into a $M$-lateral shape $Q$; (iii) magnify $Q$ by a factor of $M$ in all directions; (iv) move the curves isometrically to the corresponding edges of $Q$, without any rotation; (v) scale every point by a factor of $S_0$. 

Before we state our lower bound result, we introduce some useful notation. We define an equivalence relationship such that we write $x_i\sim x_j$ if and only if $x_i,x_j\in P'_m$ for some $1\le m\le M$. 
Now for each $1\le m\le M$, we consider a subset $T_m$ of $\{(k,\ell): x_k\nsim x_\ell\}$, concerning all the points on $P'_m$ and $P'_{m+1}$ near the edge of the partition point. Note that here we identify $P'_{M+1}$ with $P'_{1}$.  More specifically, if there exists a partition point $x_{i_m}\in P_n$ in the sense that $P'_m=\{x_{i_m-n/M+1}, x_{i_m-n/M+2}, ..., x_{i_m}\}$ and $P'_{m+1}=\{x_{i_m+1}, x_{i_m+2}, ..., x_{i_m+n/M}\}$, then the subset $T_m$ is defined as
\beq \label{Tm}
T_m=\bigg\{(k,\ell): i_m-\frac{n^\delta}{2}+1\le k\le i_m, i_m+1\le \ell\le i_m+\frac{n^\delta}{2}\bigg\},
\eeq
where $\delta\in(0,1)$ so that $n^\delta$ is an even integer, and
\beq \label{delta}
n^{\delta}<\frac{n}{M},
\eeq
or {$M<n^{1-\delta}$}. In this way, we have
so that $T_m\subset P'_m\times P'_{m+1}$ and $T_\ell\cap T_k=\emptyset$ for any $1\le k\ne \ell \le M$. Apparently, we have 
\[
\bigcup_{1\le m\le M}T_m \subseteq T_0=\{(k,\ell): x_k\nsim x_\ell\}.
\]
Throughout, we assume
\[
c-2\delta\le m<c,\quad \frac{1-c}{2}\le \delta\ \le 1-c,\quad m<1-\delta,
\]
where we calibrate $M\asymp n^m$ for some $m\ge 0$.
The following proposition provides a lower bound estimate of $C(Y_0)$.

\begin{prop} \label{propphi0}
	Suppose $Y_0=\phi_0(X)$ is the mapping satisfying (\ref{iso.prop}) and (\ref{sep.prop}) where $S_0=n^c$ for some constant $0<c<1$. We define $T_m$ as in (\ref{Tm}) for $1\le m\le M$ such that   (\ref{M.assum}) and (\ref{delta}) hold. Then we have
	\beq
	C(Y_0)\geq -nk\log(E_1+E_2+E_3+E_4)+E_5,
	\eeq
	where
\begin{align*}
E_1&=\sum_{(k,\ell): x_k\sim x_\ell}\frac{1}{1+ S_0^2\|x_k-x_{\ell}\|^2},\qquad
E_2=\sum_{(k,\ell)\in T_0\setminus \{\cup_{1\le m\le M}T_m\}}\frac{1}{1+ c^2S_0^2},\\
E_3&=M\sum_{j=1}^{n^{\delta}/2} \frac{j}{1+ c^2S_0^2},\qquad E_4= M\sum_{j=1}^{n^{\delta}/2-1} \frac{j}{1+ c^2S_0^2},\\
E_5&=  -(nk-Mk^2)\log \left(1+\frac{4\pi^2 k^2S^2_0}{n^2}\right)-Mk^2\log \left(1+C^2S_0^2\right).
\end{align*}
	\end{prop}
	
	\begin{proof}
 Note that in this case we have
 \[
C(Y_0)=\sum_{i \neq j\atop x_i, x_j~\mbox{\tiny neighbors}}  \log(q_{ij}),
 \]
 where
 $$ q_{ij} = \frac{(1+\|\phi_0(x_i) - \phi_0(x_j)\|^2)^{-1}}{\sum_{k \neq \ell} (1+ \|\phi_0(x_k) - \phi_0(x_{\ell})\|^2)^{-1}}.$$
	We start from the equation 
	$$ \log(q_{ij}) = - \log\left(1+\|\phi_0(x_i) - \phi_0(x_j)\|^2 \right) - \log\left(\sum_{k \neq \ell} (1+ \|\phi_0(x_k) - \phi_0(x_{\ell})\|^2)^{-1} \right).$$
Suppose $x_i$ and $x_j$ are $k-$nearest neighbors. Then we have 
\[
 \|\phi(x_i) - \phi(x_j)\|^2= S_0^2\|x_i - x_j\|^2\le \frac{4\pi^2 k^2S^2_0}{n^2},
 \]
 if $x_i\sim x_j$; and, for $C=1/c$, we have
 \[
  c^2S_0^2\le \|\phi(x_i) - \phi(x_j)\|^2\le C^2S_0^2,
 \]
  if $x_i\nsim x_j$.
  As a result, we have
  \begin{align*}
  \sum_{i \neq j\atop x_i, x_j~\mbox{\tiny neighbors}} \log \left(1+\|\phi(x_i) - \phi(x_j)\|^2 \right) &\le \sum_{\substack{i \neq j\\ x_i, x_j~\mbox{\tiny neighbors} \\ x_i\sim x_j}} \log \left(1+\frac{4\pi^2 k^2S^2_0}{n^2}\right)+\sum_{\substack{i \neq j\\ x_i, x_j~\mbox{\tiny neighbors} \\ x_i\nsim x_j}} \log \left(1+C^2S_0^2\right)\\
  &= (nk-Mk^2)\log \left(1+\frac{4\pi^2 k^2S^2_0}{n^2}\right)+Mk^2\log \left(1+C^2S_0^2\right)
  \end{align*}
  for some constant $C>0$.
On the other hand, we have
\begin{align*}
    \sum_{k \neq \ell} \frac{1}{1+ \|\phi(x_k) - \phi(x_{\ell})\|^2} 
    &=\sum_{(k,\ell): x_k\sim x_\ell}\frac{1}{1+ \|\phi(x_k) - \phi(x_{\ell})\|^2}+\sum_{(k,\ell): x_k\nsim x_\ell}\frac{1}{1+ \|\phi(x_k) - \phi(x_{\ell})\|^2}\\
    &\le  \sum_{(k,\ell): x_k\sim x_\ell}\frac{1}{1+ S_0^2\|x_k -x_{\ell}\|^2}+\sum_{(k,\ell): x_k\nsim x_\ell}\frac{1}{1+ c^2S_0^2}\\
    &\equiv E_{1}+E_{0}.
\end{align*}
In particular, we have
\begin{align*}
E_0 &= \sum_{(k,\ell)\in T_0\setminus \{\cup_{1\le m\le M}T_m\}}\frac{1}{1+ c^2S_0^2}+\sum_{m=1}^{M}\sum_{(k,\ell)\in T_m} \frac{1}{1+ c^2S_0^2}\\
&=\sum_{(k,\ell)\in T_0\setminus \{\cup_{1\le m\le M}T_m\}}\frac{1}{1+ c^2S_0^2}+M\sum_{j=1}^{n^{\delta}/2} \frac{j}{1+ c^2S_0^2}+M\sum_{j=n^{\delta}/2 +1}^{n^{\delta}-1} \frac{n^\delta-j}{1+ c^2S_0^2}\\
&=\sum_{(k,\ell)\in T_0\setminus \{\cup_{1\le m\le M}T_m\}}\frac{1}{1+ c^2S_0^2}+M\sum_{j=1}^{n^{\delta}/2} \frac{j}{1+ c^2S_0^2}+\sum_{j=1}^{n^{\delta}/2-1} \frac{j}{1+ c^2S_0^2}
\end{align*}
Thus
\begin{align*}
\sum_{i \neq j\atop x_i, x_j~\mbox{\tiny neighbors}}\log(q_{ij})&\ge -(nk-Mk^2)\log \left(1+\frac{4\pi^2 k^2S^2_0}{n^2}\right)-Mk^2\log \left(1+C^2S_0^2\right)\\
&\quad-nk\log(E_1+E_2+E_3+E_4),
\end{align*}
where
\begin{align*}
E_1&=\sum_{(k,\ell): x_k\sim x_\ell}\frac{1}{1+ S_0^2\|x_k-x_{\ell}\|^2},\\
E_2&=\sum_{(k,\ell)\in T_0\setminus \{\cup_{1\le m\le M}T_m\}}\frac{1}{1+ c^2S_0^2},\\
E_3&=M\sum_{j=1}^{n^{\delta}/2} \frac{j}{1+ c^2S_0^2},\\
E_4&= M\sum_{j=1}^{n^{\delta}/2-1} \frac{j}{1+ c^2S_0^2}.
\end{align*}
As a result, we conclude that
\beq\label{low.ineq}
C(Y_0)= \sum_{i \neq j\atop x_i, x_j~\mbox{\tiny neighbors}}\log(q_{ij})\ge-nk\log(E_1+E_2+E_3+E_4)+ E_5.
\eeq
This completes the proof.
\end{proof}

\begin{prop} \label{prop.E}
Under the assumptions of Proposition \ref{propphi0}, if in addition $T_m$ is chosen such that 
$M\lesssim n^c$,
it holds that
\beq
\frac{n^2}{S_0}\lesssim E_1+E_2+E_3+E_4\lesssim \frac{n^2}{S_0}+\frac{Mn^{2\delta}}{S_0^2}.
\eeq
\end{prop}

\begin{proof}
Recall that
\begin{align*}
E_1&=\sum_{(k,\ell): x_k\sim x_\ell}\frac{1}{1+ S_0^2\|x_k-x_{\ell}\|^2},\\
E_2&=\sum_{(k,\ell)\in T_0\setminus \{\cup_{1\le m\le M}T_m\}}\frac{1}{1+ c^2S_0^2},\\
E_3&=M\sum_{j=1}^{n^{\delta}/2} \frac{j}{1+ c^2S_0^2},\\
E_4&= M\sum_{j=1}^{n^{\delta}/2-1} \frac{j}{1+ c^2S_0^2}.
\end{align*}
As a result, since $S_0\to\infty$ as $n\to\infty$, we have
\[
E_2\lesssim \frac{n^2}{S_0^2},
\]
and by the assumption of the proposition, we have
{\[
E_3+E_4\lesssim \frac{Mn^{2\delta}}{S_0^2}
\]}
Now for $E_1$, we consider the following estimates. On the one hand, we consider
\[
F = \{ (k,\ell): x_k\sim x_\ell, k\ne \ell, \|x_k-x_\ell\|S_0\le 1\},
\]
and it follows that
\begin{align*}
E_1&=\sum_{(k,\ell)\in F}\frac{1}{1+ S_0^2\|x_k-x_{\ell}\|^2}+\sum_{(k,\ell)\in F^c}\frac{1}{1+ S_0^2\|x_k-x_{\ell}\|^2}\\
&\asymp |F|+\frac{1}{S_0^2}\sum_{(k,\ell)\in F^c}\frac{1}{\|x_k-x_{\ell}\|^2}.
\end{align*}
By Lemma \ref{lem.F}, 
{since $\min_{k\ne \ell}\|x_k-x_\ell\|\asymp 1/n$ and since $1\lesssim  \frac{n}{S_0}\lesssim \frac{n}{M}$, we have
\[
|F| \asymp M\cdot \frac{n^2}{MS_0}=\frac{n^2}{S_0}.
\]}
On the other hand, we have
\begin{align*}
&\sum_{(k,\ell)\in F^c}\frac{1}{1+S_0^2\|x_k-x_{\ell}\|^2}\lesssim\sum_{(k,\ell)\in F^c}\frac{1}{ S_0^2\|x_k-x_{\ell}\|^2}\lesssim \frac{1}{S_0^2}\sum_{(k,\ell):\|x_k-x_\ell\|_2S_0>1}\frac{1}{\|x_k-x_\ell\|^2}\lesssim \frac{n}{S_0},
\end{align*}
where the last inequality follows from
\begin{align}
&\sum_{(k,\ell):\|x_k-x_\ell\|_2S_0>1}\frac{1}{\|x_k-x_\ell\|^2}\le \sum_{i: 4S_0\sin^2(\pi i/n)>1}^{n/M-1}\frac{n-i}{4\sin^2(\pi i/n)}\nonumber\\
&\le \sum_{i:\arcsin(\frac{1}{2\sqrt{S_0}})<\pi i/n\le \pi/8}\frac{n-i}{4\sin^2(\pi i/n)}\nonumber\\
&\lesssim \sum_{i:\arcsin(\frac{1}{2\sqrt{S_0}})<\pi i/n\le \pi/8}\frac{n-i}{(\pi i/n)^2}\nonumber\\
&\lesssim n^2\sum_{i=\lfloor n/\sqrt{S_0}\rfloor}^n\frac{n-i}{i^2}\nonumber\\
&\lesssim n^2\sum_{i=\lfloor n/\sqrt{S_0}\rfloor}^n\frac{n-n/\sqrt{S_0}}{i^2}\lesssim nS_0.
\end{align}
It then holds that
\beq
E_1\lesssim \frac{n}{S_0}\ll \frac{n^2}{S_0}.
\eeq
This along with the previous estimates leads to $E_1+E_2+E_3+E_4\lesssim \frac{n^2}{S_0}+\frac{Mn^{2\delta}}{S_0^2}$. On the other hand, we have
\[
E_1+E_2+E_3+E_4\ge E_1\gtrsim |F| \asymp \frac{n^2}{S_0}.
\]
This completes the proof of the proposition.
\end{proof}

\begin{lem} \label{lem.F}
Let $\{a_i\}_{1\le i\le n}$ be equi-spaced point on $\R$ so that $a_{i+1}-a_i=h$. Let $F=\{(k,\ell): a_k\sim a_\ell, |a_k-a_\ell|S\le 1\}$. If $1\ll (hS)^{-1}\ll n$, then we have $|F|\asymp \frac{n}{hS}$. If instead $(hS)^{-1}\gtrsim n$, then we have $|F|\asymp n^2$.
\end{lem}
\begin{proof}
Consider a $n\times n$ matrix, whose entries are pairwise distance between $\{a_i\}$. Let $q$ be the biggest integer  such that $|a_1-a_{1+q}|S\le 1$. Then we have 
\[
q=\lfloor\frac{1}{hS}\rfloor,
\]
and, if $1\le q\ll n$,
\[
|F| = (n-1)^2-(n-q)^2=  2nq-2n-q^2+1= 2n(q-1)-q^2+1\asymp \frac{n}{hS}.
\]
When $q\gtrsim n$, we have $|F|\asymp n^2.$
This completes the proof.
\end{proof}

The following proposition provides an upper bound estimate of $C(Y^*)$.

\begin{prop} \label{iso.prop2}
 Suppose $\phi:P_n \rightarrow \mathbb{R}^2$ is a bilipschitz embedding at some scale, meaning that there exists a scaling factor $S>0$ and a distortion factor $L \geq 1$ such that
\beq \label{dist.def}
 \forall~p_1, p_2 \in P_n \qquad S\cdot \| p_1 - p_2\|_{\mathbb{R}^2} \leq \| \phi(p_1) - \phi(p_2) \|_{\mathbb{R}^2} \leq S \cdot L \cdot \| p_1 - p_2\|_{\mathbb{R}^2}.
 \eeq
Moreover, let
$$ q_{ij} = \frac{(1+\|\phi(x_i) - \phi(x_j)\|^2)^{-1}}{\sum_{k \neq \ell} (1+ \|\phi(x_k) - \phi(x_{\ell})\|^2)^{-1}},$$
 let $\{P'_m\}_{1\le m\le M}$ be a partition of $P_n$ satisfying (\ref{iso.prop}) and (\ref{sep.prop}), and let $\{T_m\}_{1\le m\le M}$ be defined as above. 
Then
$$\sum_{i \neq j\atop x_i, x_j~\mbox{\tiny neighbors}}  \log(q_{ij}) \leq - n k \cdot \log(D_1+D_2+D_3+D_4)-nk\log D_5,$$
where
\begin{align*}
D_1&=\sum_{(k,\ell): x_k\sim x_\ell}\frac{1}{1+ S^2L^2\|x_k - x_{\ell}\|^2},\qquad D_2=\sum_{(k,\ell)\in T_0\setminus \{\cup_{1\le m\le M}T_m\}}\frac{1}{1+4S^2L^2},\\
D_3&=M\sum_{j=1}^{n^{\delta}/2} \frac{j}{1+S^2L^2j^2 4\pi^2 /n^2},\qquad D_4= M\sum_{j=1}^{n^{\delta}/2-1} \frac{j}{1+S^2L^2 (n^\delta-j)^2 4\pi^2 /n^2}\\
D_5&=1+\frac{4\pi^2k^2S^2}{n^2}.
\end{align*}
\end{prop}

\begin{proof}
We argue with a pointwise bound. Note that
$$ \log(q_{ij}) = - \log\left(1+\|\phi(x_i) - \phi(x_j)\|^2 \right) - \log\left(\sum_{k \neq \ell} (1+ \|\phi(x_k) - \phi(x_{\ell})\|^2)^{-1} \right).$$
Suppose $x_i$ and $x_j$ are $k-$nearest neighbors. Then we have that
$$ \|x_i - x_j\| \leq \frac{2\pi k}{n}.$$
Therefore,  by (\ref{dist.def}), we have
$$  \|\phi(x_i) - \phi(x_j)\|^2 \geq S^2\|x_i-x_j\|^2\geq  \frac{4\pi^2k^2S^2}{n^2}  $$
and thus
$$  - \log\left(1+\|\phi(x_i) - \phi(x_j)\|^2 \right) \leq - \log\left(1 + \frac{4\pi^2k^2S^2}{n^2}\right).$$
Now we define an equivalence relationship such that we write $x_i\sim x_j$ if and only if $x_i,x_j\in P'_m$ for some $1\le m\le M$. We also have
\begin{align*}
    \sum_{k \neq \ell} \frac{1}{1+ \|\phi(x_k) - \phi(x_{\ell})\|^2} 
    &=\sum_{(k,\ell): x_k\sim x_\ell}\frac{1}{1+ \|\phi(x_k) - \phi(x_{\ell})\|^2}+\sum_{(k,\ell): x_k\nsim x_\ell}\frac{1}{1+ \|\phi(x_k) - \phi(x_{\ell})\|^2}\\
    &\geq  \sum_{(k,\ell): x_k\sim x_\ell}\frac{1}{1+ S^2L^2\|x_k -x_{\ell}\|^2}+\sum_{(k,\ell): x_k\nsim x_\ell}\frac{1}{1+ S^2L^2\|x_k -x_{\ell}\|^2}\\
    &\equiv D_{1}+D_{0}.
\end{align*}
In the following, we obtain a lower bound for $D_{0}$. To begin with, for each $(k,\ell)$ such that $x_k\nsim x_\ell$, we have the trivial bound
\beq
\frac{1}{1+ S^2L^2\|x_k -x_{\ell}\|^2}\geq \frac{1}{1+ 4S^2L^2}.
\eeq
As a result, since
\[
\bigcup_{1\le m\le M}T_m \subseteq T_0=\{(k,\ell): x_k\nsim x_\ell\}.
\]
we have
\begin{align*}
D_0 &\geq \sum_{m=1}^{M}\sum_{(k,\ell)\in T_m} \frac{1}{1+S^2L^2\|x_k-x_\ell\|^2}.
\end{align*}
Now since for each $T_m$, if $(k,\ell)\in T_m$ and without loss of generality that $k< \ell,$ we have
\beq
\|x_k-x_\ell\|\le \frac{(\ell-k)2\pi}{n},
\eeq
then it follows that
\begin{align*}
D_0 &\ge \sum_{(k,\ell)\in T_0\setminus \{\cup_{1\le m\le M}T_m\}}\frac{1}{1+4S^2L^2}+\sum_{m=1}^{M}\sum_{(k,\ell)\in T_m} \frac{1}{1+S^2L^2(\ell-k)^24\pi^2/n^2}\\
&=\sum_{(k,\ell)\in T_0\setminus \{\cup_{1\le m\le M}T_m\}}\frac{1}{1+4S^2L^2}+M\sum_{j=1}^{n^{\delta}/2} \frac{j}{1+S^2L^2 j^2 4\pi^2 /n^2}+M\sum_{j=n^{\delta}/2 +1}^{n^{\delta}-1} \frac{n^\delta-j}{1+S^2L^2 j^2 4\pi^2 /n^2}\\
&=\sum_{(k,\ell)\in T_0\setminus \{\cup_{1\le m\le M}T_m\}}\frac{1}{1+4S^2L^2}+M\sum_{j=1}^{n^{\delta}/2} \frac{j}{1+S^2L^2j^2 4\pi^2 /n^2}+M\sum_{j=1}^{n^{\delta}/2-1} \frac{j}{1+S^2L^2 (n^\delta-j)^2 4\pi^2 /n^2}
\end{align*}
Thus, we have
\beq
\log(q_{ij})\le -\log(D_1+D_2+D_3+D_4)-\log D_5,
\eeq
where
\begin{align*}
D_1&=\sum_{(k,\ell): x_k\sim x_\ell}\frac{1}{1+ S^2L^2\|x_k - x_{\ell}\|^2},\\
D_2&=\sum_{(k,\ell)\in T_0\setminus \{\cup_{1\le m\le M}T_m\}}\frac{1}{1+4S^2L^2},\\
D_3&=M\sum_{j=1}^{n^{\delta}/2} \frac{j}{1+S^2L^2j^2 4\pi^2 /n^2},\\
D_4&= M\sum_{j=1}^{n^{\delta}/2-1} \frac{j}{1+S^2L^2 (n^\delta-j)^2 4\pi^2 /n^2}.
\end{align*}
This completes the proof of the theorem.
\end{proof}

\subsection{Main proof argument}

\paragraph{Proof of (\ref{low}).} See Proposition \ref{propphi0}. We have
\beq
	C(Y_0)\geq -nk\log(E_1+E_2+E_3+E_4)+ E_5,
	\eeq
	where
\begin{align*}
E_1&=\sum_{(k,\ell): x_k\sim x_\ell}\frac{1}{1+ S_0^2\|x_k-x_{\ell}\|^2},\qquad
E_2=\sum_{(k,\ell)\in T_0\setminus \{\cup_{1\le m\le M}T_m\}}\frac{1}{1+ c^2S_0^2},\\
E_3&=M\sum_{j=1}^{n^{\delta}/2} \frac{j}{1+ c^2S_0^2},\qquad E_4= M\sum_{j=1}^{n^{\delta}/2-1} \frac{j}{1+ c^2S_0^2},\\
E_5&=  -(nk-Mk^2)\log \left(1+\frac{4\pi^2 k^2S^2_0}{n^2}\right)-Mk^2\log \left(1+C^2S_0^2\right).
\end{align*}

\paragraph{Proof of (\ref{up}).} See Proposition \ref{iso.prop2}. We have
\beq
C(Y^*)\le - n k \cdot \log(D_1+D_2+D_3+D_4)-nk\log D_5,
\eeq
where
\begin{align*}
D_1&=\sum_{(k,\ell): x_k\sim x_\ell}\frac{1}{1+ S^2L^2\|x_k - x_{\ell}\|^2},\qquad D_2=\sum_{(k,\ell)\in T_0\setminus \{\cup_{1\le m\le M}T_m\}}\frac{1}{1+4S^2L^2},\\
D_3&=M\sum_{j=1}^{n^{\delta}/2} \frac{j}{1+S^2L^2j^2 4\pi^2 /n^2},\qquad D_4= M\sum_{j=1}^{n^{\delta}/2-1} \frac{j}{1+S^2L^2 (n^\delta-j)^2 4\pi^2 /n^2}\\
D_5&=1+\frac{4\pi^2k^2S^2}{n^2}.
\end{align*}

\paragraph{Completing the argument.}
	Now by the relation (\ref{logic.ineq}), we have
\[
-\frac{E_5}{nk}+\log(E_1+E_2+E_3+E_4)\ge \log D_5+\log(D_1+D_2+D_3+D_4).
\]
Note that $c+\delta\le 1$ and $m<c$, it then follows that  $m+2\delta+c<2c+2\delta\le 2$, we have
\begin{align*}
\frac{-\frac{E_5}{nk}-\log D_5}{\log(E_1+E_2+E_3+E_4)}&\lesssim \frac{\log \left(1+\frac{4\pi^2 k^2S^2_0}{n^2}\right)-\log D_5}{\log (E_1+E_2+E_3+E_4)}+\frac{Mk^2\log (1+CS_0^2)}{nk\log (E_1+E_2+E_3+E_4)}\\
&\lesssim \frac{k^2S_0^2}{n^2}+\frac{Mk\log n}{n}\\
&\lesssim \frac{Mn^{2\delta}S}{n^2\log n}\to 0,
\end{align*}
or 
\beq
-\frac{E_5}{nk}-\log D_5=o(\log(E_1+E_2+E_3+E_4)).
\eeq
Then it follows that
\[
(1+o(1))\log(E_1+E_2+E_3+E_4)\ge\log(D_1+D_2+D_3+D_4),
\]
that is, for sufficiently large $n$,
\beq \label{crite}
(E_1+E_2+E_3+E_4)^{1+\epsilon_n}\ge D_1+D_2+D_3+D_4,
\eeq
for some series $\epsilon_n=o(1)$. 
Note that we can find a sequence of $S_0$ such that $D_1+D_2=E_1+E_2$ for all $n$. Specifically, for each $n$, $D_1+D_2$ is monotonic decreasing in $S$ and $E_1+E_2$ is monotonic decreasing in $S_0$. For any fixed $S$, $T(S_0)=D_1+D_2-E_1-E_2$ as a function of $S_0=\theta S$ satisfies $T(S_0)>0$ if 
\[
\theta<\min\{L, 2L/c\},
\]
and $T(S_0)<0$  if
\[
\theta>\max\{L,2L/c\}.
\]
By continuity of $T(S_0)$, we know that there exists some $\theta=[\min\{L, 2L/c\},\max\{L, 2L/c\}]$ (possibly depending on $n$) such that $T(S_0)=0$. We choose such an $S_0$ so that $D_1+D_2=E_1+E_2$.
Next, we show that for the above selection of $S_0$ (so that $S_0\asymp S(\phi^*)$ or $c=\tau\in(0,1)$), if in addition $L\asymp 1$, then there exists some series $\eta_n$ such that $\eta_n\gtrsim \epsilon_n$ and
\beq \label{epsilon}
\eta_n<\frac{\log\frac{D_3+D_4}{3(E_3+E_4)}}{\log (E_1+E_2+E_3+E_4)},\qquad  \eta_n<\frac{\log(1+\frac{D_3+D_4}{3(D_1+D_2)})}{\log(E_1+E_2+E_3+E_4)}.
\eeq
It then follows that
\[
(E_1+E_2+E_3+E_4)^{\eta_n}<\frac{D_3+D_4}{3(E_3+E_4)},\qquad (E_1+E_2+E_3+E_4)^{\eta_n}<1+\frac{D_3+D_4}{3(D_1+D_2)},
\]
so that
\begin{align*}
(E_1+E_2+E_3+E_4)^{1+\eta_n}&\le(E_1+E_2)(E_1+E_2+E_3+E_4)^{\eta_n}+(E_3+E_4)(E_1+E_2+E_3+E_4)^{\eta_n}\\
&\le (D_1+D_2)+\frac{2}{3}(D_3+D_4).
\end{align*}
However, this is contradictory to (\ref{crite}). Therefore, we conclude that $L\asymp 1$ cannot be true.
As a result, we conclude that $L\gg 1$.

The rest of the proof is devoted to (\ref{epsilon}).
By Proposition \ref{prop.E}, we have
\[
1\lesssim \frac{n^2}{S_0}\lesssim E_1+E_2+E_3+E_4\lesssim \frac{n^2}{S_0}+\frac{Mn^{2\delta}}{S_0^2},
\]
\[
E_1+E_2\lesssim \frac{n^2}{S_0},\qquad E_3+E_4\lesssim \frac{Mn^{2\delta}}{S_0^2}.
\]
In particular, when {
\[
0\le \delta\le 1-c,\quad m\le c,
\]
we have
\[
m+2\delta-c\le 2\delta+2c\le 2,
\]
}
so that
\[
\frac{n^2}{S_0}+\frac{Mn^{2\delta}}{S_0^2}\asymp\frac{n^2}{S_0}. 
\]
Similarly, $D_1+D_2\lesssim \frac{n^2}{S}.$ However, for $D_3$, since {$\delta\le 1-c$}, we have
\begin{align*}
D_3&=M\sum_{j=1}^{n^{\delta}/2} \frac{j}{1+S^2L^2j^2 4\pi^2 /n^2}\\
&=M\sum_{j: S^2j^2<n^2, j\le n^\delta/2} \frac{j}{1+S^2L^2j^2 4\pi^2 /n^2}+M\sum_{j: S^2j^2\ge n^2,j\le n^\delta/2} \frac{j}{1+S^2L^2j^2 4\pi^2 /n^2}\\
&\asymp Mn^{2\delta}\\
&\gg E_3,
\end{align*}
and
\begin{align*}
D_4&= M\sum_{j=1}^{n^{\delta}/2-1} \frac{j}{1+S^2L^2 (n^\delta-j)^2 4\pi^2 /n^2}\\
&=M\sum_{j:S^2L^2 (n^\delta-j)^2<n^2 }\frac{j}{1+S^2L^2 (n^\delta-j)^2 4\pi^2 /n^2}+M\sum_{j:S^2L^2 (n^\delta-j)^2\ge n^2 }\frac{j}{1+S^2L^2 (n^\delta-j)^2 4\pi^2 /n^2}\\
&\asymp Mn^{2\delta}\\
&\asymp D_3\\
&\gg E_3.
\end{align*}
As a result, we have
\[
\frac{D_3+D_4}{3(E_3+E_4)}\ge CS_0^2.
\]
As long as there exists some constant $q>0$ such that
\beq \label{ep1}
\frac{D_3+D_4}{3(E_3+E_4)}\ge cS_0^2 \gg (n^2/S_0)^q\ge (E_1+E_2+E_3+E_4)^q,
\eeq
we can choose $\epsilon_n\lesssim \eta_n <q$ so that the first inequality in (\ref{epsilon}) holds. To see the existence of $q$ in (\ref{ep1}), we can just choose $q$ such that $q<\frac{2c}{2-c}$ holds.
Similarly, since
\[
\frac{D_3+D_4}{D_1+D_2}\ge \frac{c'Mn^{2\delta}S}{n^2},
\]
we have
\[
\frac{\log(1+\frac{D_3+D_4}{3(D_1+D_2)})}{\log(E_1+E_2+E_3+E_4)}\gtrsim \frac{\log(1+\frac{c'Mn^{2\delta}S}{n^2})}{\log n}.
\]
Since
\[
m+c+2\delta<2(c+\delta)<2,
\]
we have
\[
\frac{\log(1+\frac{c'Mn^{2\delta}S}{n^2})}{\log n}\asymp \frac{Mn^{2\delta}S}{n^2\log n}.
\]
we can choose 
\beq \label{ep2}
\epsilon_n\lesssim \eta_n \asymp \frac{Mn^{2\delta}S}{n^2\log n}
\eeq
so that the second inequality in (\ref{epsilon}) holds. To see the existence of $\eta_n$ in (\ref{ep2}), note that
\begin{align*}
\epsilon_n&\lesssim \frac{nk\log(E_5/D_5)}{nk\log (E_1+E_2+E_3+E_4)}+\frac{Mk^2\log (1+CS_0^2)}{nk\log (E_1+E_2+E_3+E_4)}\\
&\lesssim \frac{k^2S_0^2}{n^2}+\frac{Mk\log n}{n}\\
&\lesssim \frac{Mn^{2\delta}S}{n^2\log n},
\end{align*}
where in the last inequality we used
\[
k=O(1),\quad  m\ge c-2\delta, \quad \delta\ge \frac{1-c}{2}.
\]
This completes the proof of (\ref{epsilon}).

\end{document}